\documentclass[dvipsnames]{article}

\usepackage[final]{slai}
\usepackage[utf8]{inputenc} 
\usepackage[T1]{fontenc}    
\usepackage{url}            
\usepackage{booktabs}       
\usepackage{amsfonts}       
\usepackage{nicefrac}       
\usepackage{booktabs} 
\usepackage{multirow}
\usepackage{wrapfig}
\usepackage{enumitem}
\usepackage{amssymb}
\usepackage{epigraph}
\usepackage{makecell}
\usepackage{listings}
\usepackage{subcaption}
\usepackage[labelfont=bf]{caption} 

\usepackage{tabularx,array}

\usepackage[ruled,vlined]{algorithm2e}
\usepackage{color, soul}
\usepackage{microtype}      
\usepackage{xcolor}         
\definecolor{slai}{HTML}{B34A78}
\usepackage{geometry}
\usepackage{mathpazo}
\usepackage[xcharter,bigdelims,vvarbb,noamssymbols,nosymbolsc]{newtxmath}
\usepackage[scaled=1.1]{zlmtt}
\usepackage{ragged2e}
\usepackage{amsmath}
\usepackage{bm}
\usepackage{latexsym}
\usepackage{graphicx}
\usepackage{float}
\usepackage{longtable}
\usepackage{booktabs} 
\usepackage{multirow}
\usepackage{amssymb}
\usepackage{longtable}
\usepackage{tabu}
\usepackage{bbding}
\usepackage{awesomebox}
\usepackage{bbding}
\usepackage[most,breakable]{tcolorbox}
\usepackage{etoolbox}
\usepackage{fancyhdr}
\usepackage{lipsum}
\usepackage{CJKutf8}
\usepackage{pdfpages}
\usepackage{multicol}
\usepackage{pgffor} 
\usepackage{fontawesome5}
\usepackage{nicematrix} 
\usepackage[edges]{forest}
\usepackage{siunitx}
\usepackage{tikz-dependency}
\usepackage{tikz}
\usepackage{bbm}
\usepackage{amssymb}
\usepackage[english]{babel}  
\usepackage{tikz}
\usepackage{hyphenat}
\usepackage{siunitx}
\usepackage{algorithmic}
\usepackage{float}
\usepackage{placeins}
\newcommand*\circled[1]{\tikz[baseline=(char.base)]{
            \node[shape=circle,draw,inner sep=0.3pt] (char) {#1};}}

\newcommand{\Rmnum}[1]{\expandafter\@slowromancap\romannumeral #1@}

\usepackage{caption} 
\usepackage{chngcntr}

\usepackage[colorlinks, linkcolor=RoyalBlue, anchorcolor=BrickRed, citecolor=RoyalBlue, urlcolor=RoyalBlue]{hyperref}

\usepackage{cleveref}
\crefname{section}{§}{§§}
\Crefname{section}{§}{§§}

\newcommand{\modelname}{SLAI T-Rex}

\newcommand\refsec[1]{Section~\hyperref[sec:#1]{\ref{sec:#1}}}
\newcommand\refsecs[2]{\hyperref[sec:#1]{§\ref{sec:#1}:~\textsc{#1}}, \hyperref[sec:#2]{§\ref{sec:#2}:~\textsc{#2}}}

\usepackage[tikz]{bclogo}
\definecolor{msftBlue}{RGB}{0,164,239}
\definecolor{msftGreen}{RGB}{127,186,0}
\definecolor{msftYello}{RGB}{255,185,0}
\definecolor{mypurple}{RGB}{138,43,226} 
\definecolor{msftBlack}{RGB}{0,0,0}

\newtcolorbox{myboxnote}[1][]{
  breakable,
  title=#1,
  colback=cyan!0,
  colbacktitle=cyan!0,
  coltitle=black,
  fonttitle=\bfseries,
  bottomrule=0pt,
  toprule=0pt,
  leftrule=1.5pt,
  rightrule=1.5pt,
  titlerule=0pt,
  arc=0pt,
  outer arc=0pt,
  colframe=lightgray,
}
\usepackage{mdframed}

\definecolor{academicblue}{RGB}{54, 95, 145}

\tcbset{
  iclrtakeawaybox/.style={
    width=\linewidth,
    colback=gray!5,                 
    colframe=gray!60,              
    colbacktitle=gray!30,            
    coltitle=black,                
    fonttitle=\bfseries,     
    boxrule=0.5pt,                 
    arc=2pt,                       
    sharp corners=south,           
    attach boxed title to top left={yshift=-0.1in,xshift=0.15in},
    boxed title style={boxrule=0pt, colframe=white},
    enhanced,
    drop shadow southeast           
  }
}
\newtcolorbox{TakeawayBox}[2][]{iclrtakeawaybox,title=#2,#1}

\usepackage{fancyhdr} 
\usepackage{blindtext} 
\usepackage{ulem}

\usepackage{xparse}
\usepackage{colortbl}

\usepackage{svg}

\title{
\vspace{-2em}
\fontsize{15}{19}\selectfont 
\modelname: Full-Parameter Post-training of the DeepSeek-V4 Family on Ascend SuperPOD
}
\author{
 AI Training Platform Team, Shenzhen Loop Area Institute\\
\href{https://www.modelscope.cn/profile/SLAIAITP}{\includegraphics[height=1.2em]{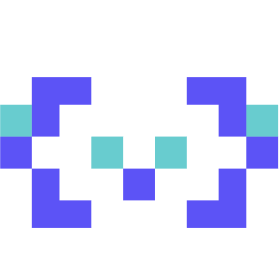} Models}   
\href{https://github.com/SLAI-AITP/Deepseek-OR}{\includegraphics[height=1em]{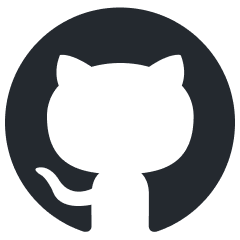} Codes}
}

\begin{document}
\vspace{-2em}
\maketitle
\vspace{-1em}
\vspace{-4mm}

\begin{abstract}
Full-parameter post-training of trillion-parameter-scale Mixture-of-Experts~(MoE) models poses substantial challenges in memory management, communication orchestration, training stability, and kernel efficiency. This report presents \textbf{\modelname}, an integrated framework that combines full-stack Ascend SuperPOD optimization with solver-grounded Continued Pre-Training~(CPT) and Supervised Fine-Tuning~(SFT) for Operations Research~(OR). Using the DeepSeek-V4 model family as the target workload, we develop a hierarchical optimization framework spanning model-level parallelism, computation--communication orchestration, and low-level kernel execution. On DeepSeek-V4-Pro, the optimized system achieves $34.22\%$ Model FLOPs Utilization~(MFU), a $2.93\times$ improvement over the open-source baseline recipe, while maintaining training stability. For OR specialization, we construct CPT and SFT data pipelines that combine collected domain resources with solver-verified synthetic optimization documents, producing 10K high-quality SFT samples across four task categories and three problem representations. Using DeepSeek-V4-Flash as the primary experimentation platform, the specialized model achieves the highest average score among the evaluated models, reaching $71.81\%$ and outperforming GPT-5.4-Mini and the original DeepSeek-V4-Flash by $3.98$ and $11.27$ percentage points, respectively. Applying the workflow to DeepSeek-V4-Pro improves its average OR score from $70.16\%$ to $77.33\%$, providing scale-up validation at the trillion-parameter level. Overall, \modelname~establishes a full-stack pathway from efficient trillion-parameter post-training on Ascend SuperPOD to solver-grounded OR specialization across the DeepSeek-V4 model family.
\end{abstract}

\vspace{-4mm}

\begin{figure*}[h]
    \centering
    \includegraphics[width=0.9\linewidth]{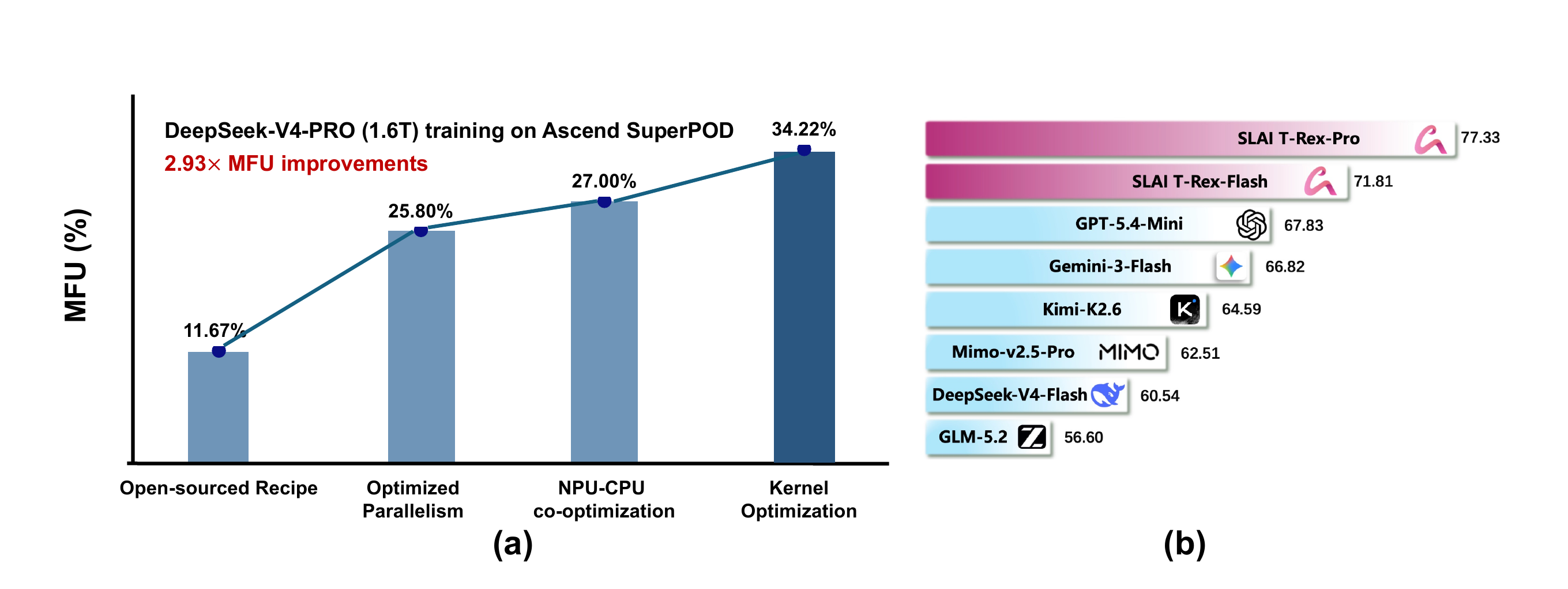}
    \caption{(a) MFU improvements during DeepSeek-V4-Pro (1.6T) training on Ascend SuperPOD. Starting from open-sourced recipes, optimized parallelism, NPU-CPU co-optimization, and kernel optimization increase MFU from 11.67\% to 34.22\%. (b) Overall accuracy, computed as the mean of NL4OPT~\citep{ramamonjison2023nl4opt}, OptiBench~\citep{yang2025optibench}, Bench4Opt (B4O) -Feasible~\citep{orgeval}, and Bench4Opt (B4O)-ORGEval~\citep{orgeval}. \modelname~achieves the highest overall score while using a smaller disclosed activated-parameter budget.}
    \label{fig:ascend-posttrain-dashboard}
\end{figure*}

\newpage
{
  \hypersetup{linkcolor=RoyalBlue, linktoc=page}
  \tableofcontents
}


\clearpage
\section{Introduction}
\label{sec:introduction}

Post-training has become a standard stage for enhancing LLMs after pretraining \citep{lai2025survey, cheng2025empowering}, improving instruction-following, reasoning power, alignment with human preferences, and the power of agentic tasks with long-horizon executions.
In particular, recent efforts on LLM post-training mainly focus on Continued Pre-Training (CPT), Supervised Fine-Tuning~(SFT)~\citep{mecklenburg2024injecting} and reinforcement learning~\citep{schulman2017proximal, shao2024deepseekmath,li2026rethinking}, together with dedicated agentic environments ~\citep{glm5team2026glm5vibecodingagentic,dong2026agent} and data pipelines~\citep{nvidia2026nemotron3superopen} for general or domain-specific tasks.

As LLMs evolve from simple chatbots into agentic workforces, the growing demand for long-context agentic reasoning continues to drive model scaling in pursuit of stronger intelligence. Recent frontier LLMs~\citep{deepseekai2025deepseekv3technicalreport, glm5team2026glm5vibecodingagentic} have pushed model sizes toward the trillion-parameter scale while adopting the hybrid attention mechanism~\citep{yang2025gated,dao2024transformers} or intricate residual connection design~\citep{mhc2025} to improve the efficiency of long-context computation and related optimizations~\citep{chen-etal-2026-dynamic-long, li-etal-2026-lycheecluster}. 
In the meantime, Mixture-of-Experts~(MoE)~\citep{fedus2022switch} architecture introduces a different route for scaling by routing each token to a selected subset of expert networks rather than activating all parameters, per-token computation grows sub-linearly with model size, enhancing reasoning power while preserving memory efficiency.

Most prior studies on large-scale MoE post-training remain tightly coupled with CUDA- or TPU-oriented training infrastructure~\citep{deepseekai2025deepseekv3technicalreport,kimiteam2026kimik2openagentic,glm5team2026glm5vibecodingagentic}, which are built around either Single-Instruction, Multiple-Thread~(SIMT) execution or systolic-array-based accelerator architectures. 
In contrast, post-training trillion-parameter-class LLMs on Single-Instruction, Multiple-Data~(SIMD) hardware remains comparatively underexplored. 
Motivated by this research gap, this report first presents a system-level infrastructure optimization for trillion-parameter-class LLM post-training based on Ascend CloudMatrix384 SuperPOD with Ascend 910C NPUs. 
As one of the recent state-of-the-art LLMs for agentic tasks and reasoning, DeepSeek-V4~\citep{deepseekai2026deepseekv4highlyefficientmilliontoken} reaches top-tier intelligence via intricate sparse attention and residual connection mechanisms. Full-parameter post-training DeepSeek-V4-Pro model on SuperPOD cluster introduces a host of challenges, spanning AscendC kernel design, communication orchestration, memory management, and multi-dimensional parallelism, as illustrated in Figure~\ref{fig:ascend-posttrain-dashboard}.  

Beyond the challenges of training infrastructure, recent LLM post-training efforts have increasingly concentrated on verifiable reasoning domains, particularly mathematical reasoning, coding, and agentic software-engineering tasks. However, many long-tailed industrial decision-making domains remain comparatively under-represented. Operations Research~(OR), for example, is central to real-world optimization problems such as scheduling, transportation, supply-chain management, and resource allocation, yet its integration with LLM training and evaluation is still at an early stage. 
Recent studies on LLM-based OR mainly focus on automatic model formulation and solver-oriented program synthesis, where LLMs translate natural-language problem descriptions into mathematical formulations or executable optimization code. Representative works~\citep{xiao2024chain,zhang2025or} further introduce domain-knowledge retrieval, multi-agent collaboration, and execution-feedback-based refinement into the OR modeling pipeline. However, these approaches remain limited by small-scale benchmarks, brittle semantic-to-structure mapping, and insufficient guarantees of formulation correctness in complex industrial OR scenarios.
Furthermore, existing LLM-based OR studies primarily emphasize agentic-system design, while domain-specific training for OR remains under-explored. This gap becomes more pronounced for trillion-parameter-class models~(e.g., DeepSeek-V4-Pro), where OR-oriented post-training requires not only domain knowledge acquisition via SFT or CPT, but also a dedicated data preparation pipeline, and verifiable feedback from optimization environments. 

In this work, we present \textbf{\modelname}, a full-stack post-training
framework for the DeepSeek-V4 model family on Ascend SuperPOD with Ascend
910C NPUs~\citep{zuo2025serving}. \modelname integrates Ascend-native
training-system optimization with solver-grounded CPT--SFT specialization for
Operations Research. DeepSeek-V4-Flash is used to develop and analyze the OR
post-training workflow, while DeepSeek-V4-Pro is used for trillion-parameter
system optimization and scale-up validation.

At the system level, we jointly optimize multi-dimensional parallelism,
communication orchestration, memory management, and kernel execution. We
further propose AuraKernel, an AscendC kernel optimization agent for the
dominant bottleneck operators of DeepSeek-V4. These optimizations increase the
MFU of DeepSeek-V4-Pro from 11.67\% to 34.22\%, corresponding to a
2.93$\times$ improvement over the open-source baseline recipe.

At the OR-specialization level, direct SFT on DeepSeek-V4-Flash improves
B4O-Feasible from 60.47\% to 65.93\% and B4O-ORGEval from 34.26\% to
48.73\%. Initializing SFT from the selected CPT checkpoint further raises the
scores to 71.22\% and 59.39\%, respectively, under the same SFT data and
optimization settings. SLAI T-Rex-Flash achieves an average score of 71.81\%
across the four OR benchmarks, exceeding GPT-5.4-Mini by 3.98 percentage
points and the original DeepSeek-V4-Flash by 11.27 percentage points. The
results indicate that SFT provides task-specific supervision, whereas CPT adds
OR-domain modeling knowledge that improves executable optimization-program
generation and formulation correctness.

Applying the workflow to DeepSeek-V4-Pro raises its average OR score from
70.16\% to 77.33\%. Together with the system results, this demonstrates that
SLAI T-Rex supports both efficient trillion-parameter post-training and
solver-grounded OR specialization on Ascend SuperPOD.

Overall, the contributions of this work are: 

\begin{itemize}
\item \textbf{Ascend-native system optimization for training complex trillion-parameter MoE LLMs:}
Based on Ascend CloudMatrix384 SuperPOD, we present a full-stack optimization framework for trillion-parameter-scale DeepSeek-V4-Pro training, achieving 34.22\% MFU with 2.93$\times$ improvement over the default baseline. 
In addition to the training acceleration, this work characterizes the key principles of \emph{Ascend-friendly system-level optimization} with jointly optimized parallelism, communication orchestration, memory movement, and kernel execution under the architectural constraints of Ascend SuperPOD. 

\item \textbf{Hardware-aware analysis of MoE, sparse attention, and memory hierarchy on Ascend:}
Through detailed profiling of DeepSeek-V4 training on Ascend, we provide an end-to-end analysis of how model architecture interacts with Ascend kernel execution, communication patterns, memory hierarchy, and distributed orchestration. 
The resulting insights bridge model-level design choices with hardware-level computation efficiency, and comprehensive profiling and bottleneck analysis further motivate the proposed optimization strategies and future research.

\item \textbf{Solver-guided AscendC kernel optimization beyond manual and heuristic optimization:}
We present one of the first attempts to apply Operations Research to Ascend kernel optimization. In contrast to prior kernel-agent approaches that mainly rely on harness construction and loop-level engineering, the proposed AuraKernel introduces OR-guided tiling optimization into the agentic kernel optimization workflow. The AscendC OR solver and dedicated harness design enable accurate tiling selection, long-horizon kernel optimization, and more systematic exploration of the Ascend kernel search space.

\item \textbf{SuperPOD-aware orchestration for complex MoE communication and parallelism:}
We present an end-to-end optimization recipe for Ascend SuperPOD with multi-dimensional parallelism and optimized training orchestration for large-scale MoE architectures.

\item \textbf{Reproducible post-training workflow for domain-specific enhancement:}
We construct an OR-oriented CPT--SFT workflow covering solver-grounded CPT
data construction, self-distilled SFT data generation, contract-aware cleaning,
and benchmark evaluation. The workflow is systematically analyzed on
DeepSeek-V4-Flash and further applied to DeepSeek-V4-Pro, where it improves
the average OR score from 70.16\% to
77.33\%.

\item \textbf{Empirical evidence for CPT--SFT synergy in structured OR reasoning:}
Under matched SFT settings on DeepSeek-V4-Flash, direct SFT provides effective
task-specific supervision, while CPT initialization contributes additional
OR-domain modeling knowledge. Their combination further improves executable
optimization-program generation and formulation correctness, with the largest
additional gain observed on B4O-ORGEval.

\end{itemize}


\section{Training Infrastructure on Ascend CloudMatrix384 SuperPOD}\label{sec:train_infra}

Mixture-of-Experts~(MoE) and hybrid attention architectures improve the efficiency--intelligence trade-off of large language models by decoupling total model capacity from per-token computation and reducing the cost of long-context processing~\citep{krajewski2024scaling,deepseekai2026deepseekv4highlyefficientmilliontoken}. These architectural benefits, however, introduce substantially more complex training workloads. Expert routing requires frequent token dispatch and All-to-All~(A2A) communication, while sparse attention produces irregular computation patterns and heterogeneous kernel behavior. Efficient training therefore requires coordinated optimization across parallelism, communication scheduling, memory management, runtime orchestration, and kernel execution.

Existing large-scale LLM training systems primarily optimize multi-dimensional parallelism and computation--communication overlap on GPU- or TPU-based clusters~\citep{yan2026scalable}. By comparison, full-parameter training of trillion-parameter-class MoE models on SIMD-oriented accelerator clusters remains less explored. This section presents an end-to-end study of such training on the Ascend CloudMatrix384 SuperPOD~\citep{zuo2025serving}, with Ascend-native optimizations spanning both the distributed training system and architecture-specific kernels.

We use DeepSeek-V4-Pro as the target workload, which has 1.6 trillion parameters and combines large-scale MoE layers with hybrid sparse attention~\citep{deepseekai2026deepseekv4highlyefficientmilliontoken}. Its architecture stresses three major dimensions of the training infrastructure: \circled{1} parameter and optimizer-state management at trillion-parameter scale, \circled{2} fine-grained expert routing with intensive A2A communication, and \circled{3} irregular sparse-attention computation requiring specialized kernel optimization. These characteristics expose the principal bottlenecks of large-scale MoE training, including memory pressure, expert-parallel scalability, token-dispatch efficiency, communication--computation overlap, and kernel performance. 

Section~\ref{sec:bottleneck} first characterizes the end-to-end training bottlenecks on Ascend SuperPOD. Section~\ref{sec:infra_framework} then presents the proposed parallelism, communication, memory, and runtime-orchestration optimizations. Finally, Section~\ref{sec:kernel_opt} introduces the kernel-level optimization workflow, including AuraKernel, our operations-research-guided agent for long-horizon AscendC performance tuning.

\begin{figure}[h!]
    \centering
    \includegraphics[width=0.9\linewidth]{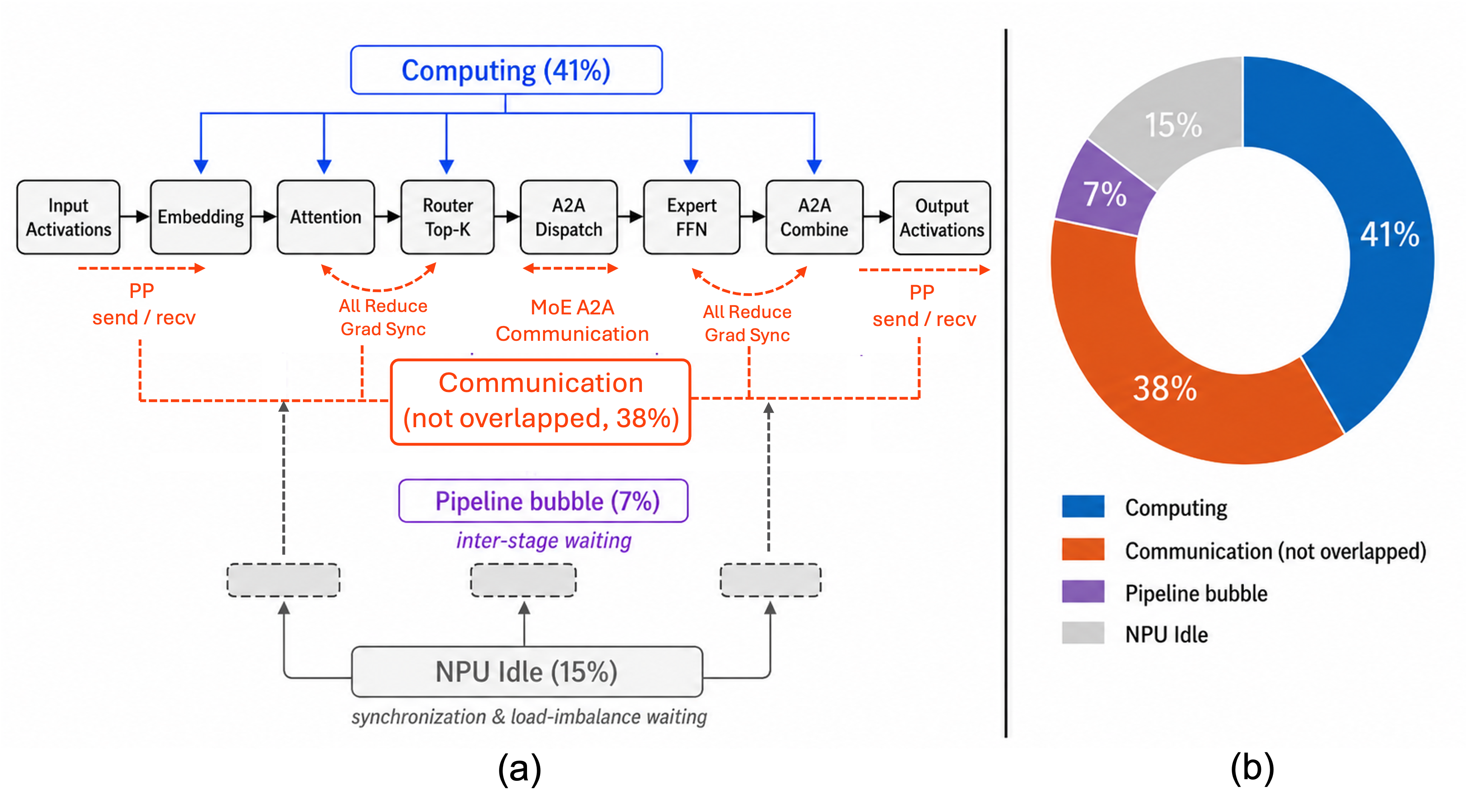}
    \caption{Latency breakdown of training trillion-parameter-class LLM model on Ascend SuperPOD NPU cluster. (a) Partition of latency in the end-to-end LLM training pipeline. (b) Averaged end-to-end latency breakdown, including computation, communication, and NPU idle bubble.}
    \label{fig:overview}
\end{figure}

\subsection{Bottlenecks of Training Trillion-parameter LLM on Ascend SuperPOD}\label{sec:bottleneck}

\begin{table}[htp]
\centering
\caption{Communication latency breakdown by parallelism dimension in DeepSeek-V4-Pro training on Ascend SuperPOD, aggregated over the profiled ranks. Communication accounts for $52.2\%$ ($40.76$\,s) of the total device kernel time. Each collective/point-to-point primitive is attributed to the parallelism dimension that induces it: expert parallelism (\texttt{AllToAllV} token dispatch/combine), pipeline parallelism (point-to-point \texttt{Send}/\texttt{Receive} of inter-stage activations and gradients), and tensor/data parallelism (\texttt{AllReduce}, \texttt{AllGather}, and \texttt{ReduceScatter} for intra-layer reductions and sharded parameter/gradient exchange).}
\label{tab:comm_breakdown}
\resizebox{\textwidth}{!}{\begin{tabular}{llccc}
\toprule
\makecell[l]{Parallelism\\Dimension} & Communication Primitives & \makecell{Latency\\(s)} & \makecell{Share of\\Comm. (\%)} & \makecell{Share of\\Total (\%)} \\
\midrule
Tensor / Data (TP+DP) & \texttt{AllReduce}, \texttt{AllGather}, \texttt{ReduceScatter} & $19.71$ & $48.4$ & $25.2$ \\
Pipeline (PP)         & \texttt{Send}, \texttt{Receive} (P2P)                    & $14.95$ & $36.7$ & $19.2$ \\
Expert (EP)           & \texttt{AllToAllV} (MoE dispatch/combine)                & $5.91$  & $14.5$ & $7.6$  \\
Misc.                 & \texttt{Broadcast}, others                               & $0.18$  & $0.4$  & $0.2$  \\
\bottomrule
\end{tabular}}
\end{table}

\paragraph{Latency overhead induced by the non-overlapped communication.}

Given the scale of the target 1.6T-parameter model, we first adopt the baseline multi-dimensional parallelism to ensure training stability while keeping model states, activations, and optimizer states within the device memory budget, as summarized in Figure~\ref{fig:overview}.
However, training MoE models under such a multi-dimensional parallelism strategy inevitably introduces latency overhead across different partitions, due to frequent synchronization, collective communication, and data redistribution among parallel groups, as summarized in Table~\ref{tab:comm_breakdown}. 

The profiling results clearly indicate that the primary latency bottleneck is collecting partial results from the TP and DP partitions. Meanwhile, the send and receive latency introduced by the traditional 1F1B pipeline parallelism~(PP)~\citep{narayanan2021efficient} contributes a large portion toward the overall latency per training step. 
Benefiting from the high-speed interconnects of the Ascend SuperPOD, the communication overhead introduced by MoE remains minimal. 


Based on this analysis, the latency of one training step can be decomposed into four major components:
\[
T_{\mathrm{step}} =
T_{\mathrm{comp}} +
T_{\mathrm{comm}}^{\mathrm{non-overlapped}} +
T_{\mathrm{bubble}} +
T_{\mathrm{idle}},
\]
where $T_{\mathrm{comp}}$ corresponds to useful dense and sparse model computation, $T_{\mathrm{comm}}^{\mathrm{exposed}}$ denotes communication that is not fully hidden by computation, $T_{\mathrm{bubble}}$ captures inter-stage waiting caused by pipeline parallelism, and $T_{\mathrm{idle}}$ represents device-side stalls caused by synchronization, load imbalance, and unresolved runtime dependencies.

In addition to the absolute latency introduced by the parallelism-induced communication~(\texttt{hcom\_} kernels), 
the imbalanced computation workload between different PP stages further introduces a latency bubble throughout the training process. 
As depicted in Figure~\ref{fig:overview}(b), the computation bubble exists uniformly across different stages of pipeline parallelism.

\paragraph{Dominant latency from model-specific kernels.}
Taking a closer look at the detailed kernel execution time, we observe that the latency of computation is dominated by a small set of architecture-specific kernels.
In particular, DeepSeek-V4 introduces the hierarchical sparse attention with heavy execution time on Ascend NPUs.
As depicted in Table~\ref{tab:kernel_bottleneck}, the backward kernel of sparse attention constitutes one of the most significant bottlenecks among all AscendC operators, where the total execution time is even comparable to the generic matrix multiplication operators. Throughout this analysis, an operator's Task-Duration share is its summed core-busy time---the per-kernel AI-Core or vector-core execution time---normalized over the $206{,}503$ compute kernels of one step after excluding communication and AI-CPU entries, with invocation count and average per-call duration listed alongside to expose the fragmented, long-tailed operators.

Concretely, the sparse-attention family contributes roughly a quarter of the non-communication kernel time, exceeding the combined share of the dense \texttt{MatMulV3} and MoE \texttt{GroupedMatmul} operators.
This concentration is intrinsic to the model architecture rather than an artifact of a particular layer configuration: the lightning indexer that selects the sparse key--value context and the shared-KV attention that consumes it are invoked in every attention block~\citep{deepseekai2026deepseekv4highlyefficientmilliontoken}, so their cost scales directly with the depth of the network.

The profiling results also make it clear \emph{why} these kernels are slow, and their bottlenecks differ fundamentally from those of matrix-multiplication operators.
The dense \texttt{MatMulV3} and MoE \texttt{GroupedMatmul} kernels are compute-bound and run on the CUBE core near saturation (over $90\%$ utilization), converting their latency into useful matrix throughput. Among these, \texttt{GroupedMatmul} is genuinely MAC-dominated, whereas \texttt{MatMulV3} is operand-load-bound, its MTE2 load share ($96.3\%$) far exceeding its MAC share ($52.5\%$).
The sparse-attention kernels do the opposite: their compute engines sit largely idle (vector-compute pipeline ratio of about $9.5\%$), and their runtime is instead spent moving key--value tiles through the local buffers and executing the scalar index and gather/scatter logic needed to assemble the sparse attention pattern.
In other words, these operators are memory- and control-bound rather than compute-bound---they occupy the top of the latency profile not because they perform heavy computation, but because their data-dependent, irregular access pattern keeps the hardware's matrix engine starved.

Motivated by these findings, this work proposes \textbf{AuraKernel}, an end-to-end AscendC kernel optimization agent that automatically restructures the large, memory- and control-bound AscendC kernels, as presented in Section~\ref{sec:kernel_agent}. 

\begin{table}[htp]
\centering
\caption{Representative top operators in DeepSeek-V4-Pro training on Ascend 910C at global batch size $1024$, ranked by Task-Duration share. Cube rows report AIC counters (MAC, MTE2, Fixpipe store, cube utilization) and vector rows AIV counters (vector compute, MTE2, MTE3, scalar utilization); the store column is therefore Fixpipe for cube rows and MTE3 for vector rows. The final row aggregates the remaining $92$ operator types. Share definition and profiling scope are given in the text.}
\label{tab:kernel_bottleneck}
\resizebox{\textwidth}{!}{%
\begin{tabular}{clcrrcccc}
	\toprule
	Rank & Operator & \makecell{Task Dur.\\Share} & Calls & \makecell{Avg\\($\mu$s)} & \makecell{Compute\\(MAC/Vec)} & \makecell{MTE2\\Load} & \makecell{Store\\(MTE3/Fix)} & \makecell{Cube/Scalar\\Util.} \\
	\midrule
	1 & \texttt{SparseAttnSharedkvGrad} (sparse-attn bwd) & $16.86\%$ & $224$    & $27{,}593$ & $9.5\%$  & $21.3\%$ & $34.2\%$ & $15.1\%$ \\
	2 & \texttt{MatMulV3}                                 & $12.77\%$ & $5{,}600$ & $836$      & $52.5\%$ & $96.3\%$ & $21.8\%$ & $85.4\%$ \\
	3 & \texttt{GroupedMatmul} (MoE experts)              & $9.50\%$  & $1{,}344$ & $2{,}592$  & $86.8\%$ & $96.2\%$ & $11.4\%$ & $95.4\%$ \\
	4 & \texttt{Cast}                                     & $6.02\%$  & $40{,}063$ & $55$      & $5.7\%$  & $89.7\%$ & $59.9\%$ & $1.7\%$ \\
	5 & \texttt{SparseLightningIndexerGradKLLoss}         & $5.52\%$  & $96$    & $21{,}082$ & $5.5\%$  & $7.9\%$  & $5.4\%$  & $96.0\%$ \\
	6 & \texttt{Add}                                      & $5.40\%$  & $16{,}579$ & $119$    & $14.0\%$ & $93.4\%$ & $23.0\%$ & $2.3\%$ \\
	7 & \texttt{GmmAddKernel} (MoE weight-grad)           & $4.72\%$  & $448$   & $3{,}859$  & $58.8\%$ & $89.4\%$ & $24.0\%$ & $95.8\%$ \\
	8 & \texttt{PpMatmulAccumAtomicKernel}                & $3.26\%$  & $896$   & $1{,}335$  & $63.5\%$ & $95.6\%$ & $9.2\%$  & $95.1\%$ \\
	9 & \texttt{BatchMatMulV2}                            & $3.06\%$  & $896$   & $1{,}253$  & $64.1\%$ & $98.7\%$ & $12.1\%$ & $94.3\%$ \\
	\midrule
	\multicolumn{2}{l}{Long tail ($92$ remaining operator types)} & $32.88\%$ & $140{,}357$ & $\approx 86$ & \multicolumn{4}{l}{predominantly small, memory-bound vector \& copy kernels} \\
	\bottomrule
	\end{tabular}}
\end{table}
\paragraph{The long tail of fragmented, memory-bound kernels.}
In addition to the top-ranked bottleneck kernels with insufficient utilization on Ascend NPUs, we observe a long tail of small kernel invocations that consistently appear in each training step, as shown in Table~\ref{tab:kernel_bottleneck}. While each individual kernel contributes only a short execution time, these fragmented kernel calls collectively account for a substantial fraction of the overall computation time.

A closer inspection shows that these long-tailed kernels are primarily element-wise and data-movement operators, such as \texttt{Cast} and \texttt{Add}, together with tensor copy and reshape operations. These operators are invoked frequently, yet each typically completes within only tens of microseconds. Unlike cube-bound matrix-multiplication kernels, these fragmented operators execute almost entirely on the vector core, where the load/store pipelines are significantly more active than the vector-compute pipeline. As shown in Table~\ref{tab:kernel_bottleneck}, the \texttt{Cast} operator alone contributes $5.7\%$ of the total computation time. However, the vector-compute utilization of \texttt{Cast} remains nearly idle~(computing core utilization $\approx 1.7\%$), whereas the memory-transfer pipelines remain heavily occupied, with AIV MTE2/MTE3 at $89.7\% / 59.9\%$. This imbalance indicates that such operators are primarily memory-bound and are natural candidates for fusion with neighboring kernels.

Overall, kernel fragmentation further degrades system-level efficiency due to the newly proposed model architecture~(e.g., mHC~\citep{mhc2025} hyper-connections) in the DeepSeek-V4 model. 
These kernels have very short execution times, where the fixed costs of kernel launch and scheduling can no longer be effectively amortized. 
Furthermore, repeated round trips to memory introduce redundant data movement and inflate the aggregate overhead. 
As a result, the collective footprint of these fragmented, memory-bound operators is far from negligible, motivating a consolidation strategy based on dedicated kernel fusion to suppress launch overhead and eliminate redundant memory traffic, as presented in Section~\ref{sec:kernel_fusion}.

\subsection{Optimizing Trillion-parameter LLM Training System on Ascend SuperPOD }\label{sec:infra_framework}

\begin{figure}[t]
  \centering
  \includegraphics[width=\textwidth]{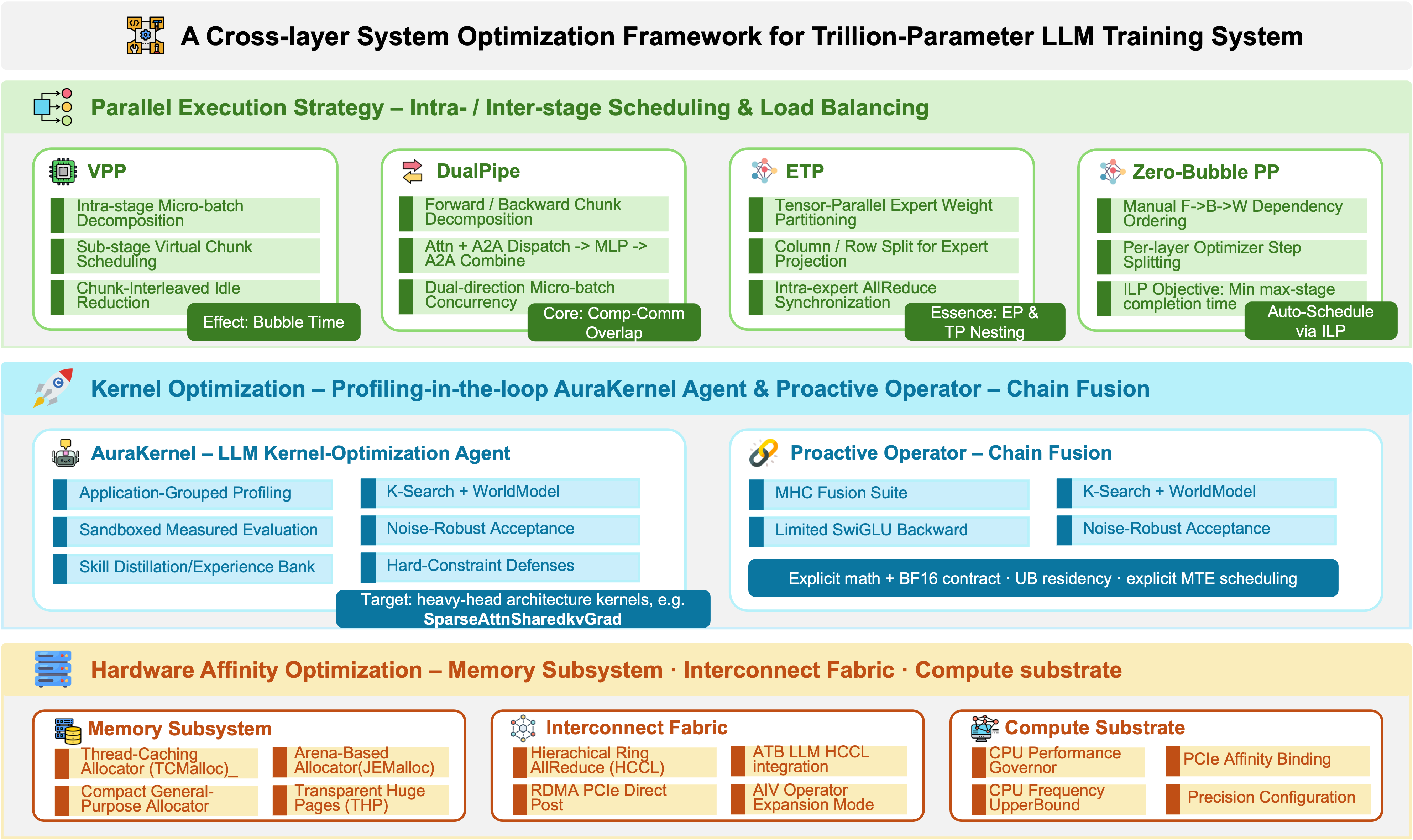}
  \caption{Overview of the proposed full-stack optimization framework, which includes three cooperating layers: parallel execution strategy, kernel optimization, and hardware-affinity tuning.}
  \label{fig:train_framework}
\end{figure}

Motivated by the bottleneck analysis in Section~\ref{sec:bottleneck}, we optimize end-to-end training performance through a hierarchical three-layer strategy: \circled{1} Optimal computation-communication orchestration via optimized training parallelism, \circled{2} Agent-driven optimization of top-ranked bottleneck kernels with AuraKernel, and \circled{3} Full-stack operator fusion to eliminate fragmented kernel overhead. The overall framework is illustrated in Figure~\ref{fig:train_framework}. 


\paragraph{Load-balanced pipeline scheduling and expert--tensor parallelism (ETP).}
\label{subsec:etp}
As discussed in Section~\ref{sec:bottleneck}, reducing the exposed communication overhead introduced by tensor parallelism~(TP) is a primary optimization target. Due to the limited memory capacity of each Ascend 910C NPU~(64GB), a naive parallelization strategy requires a relatively high TP degree, such as TP=4, to fit the model states and activations within the device memory budget. However, this also increases the frequency and cost of TP collectives.

\begin{figure}[t]
\centering
\includegraphics[width=0.9\linewidth]{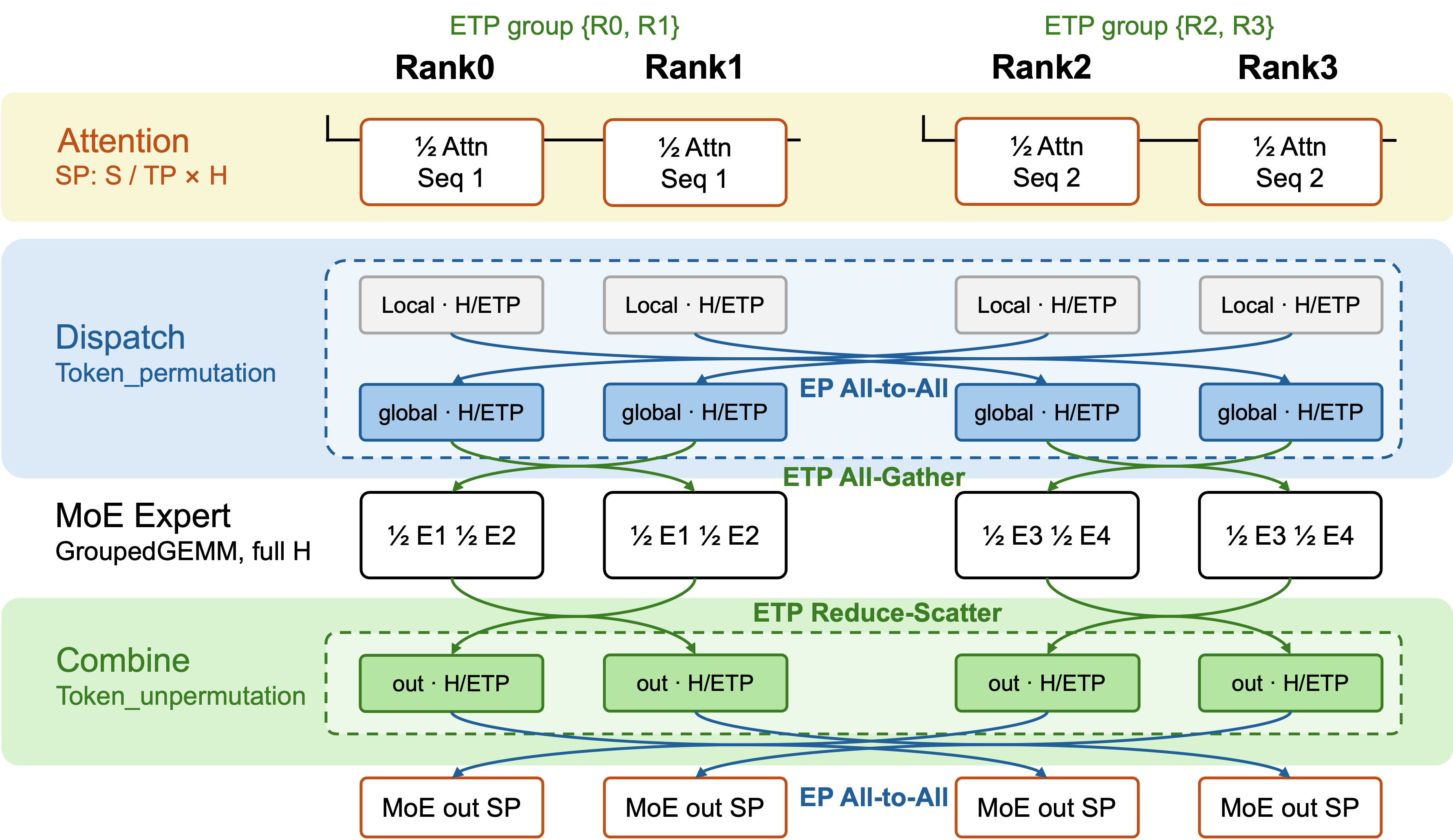}
\caption{Communication pattern of the ETP token dispatcher (ETP2‑EP2 example). Dispatch and combine wrap the expert grouped-GEMM with an EP all‑to‑all (token routing) and an ETP all‑gather/reduce‑scatter over the hidden dimension~\citep{gale2023megablocks}. With MoE Parallel Folding, ETP folds into EP as a single group, so one all‑to‑all performs dispatch and the separate ETP collectives are eliminated.}
  \label{fig:etp-dispatcher}
\end{figure}

Following the Parallel Folding strategy in Megatron-LM~\citep{shoeybi2019megatron}, we enable expert-tensor parallelism~(ETP) for MoE layers while keeping the pipeline parallelism~(PP) configuration fixed. This decouples the parallelism choices between attention and MoE modules: attention layers can adopt a lower TP degree~(TP=2) to reduce communication overhead and improve GEMM efficiency on large dense matrices, while MoE layers use ETP=1 to preserve the full expert width and avoid further fragmenting expert computation. The resulting dispatch and execution strategy is illustrated in Figure~\ref{fig:etp-dispatcher}.

\paragraph{Virtual Pipeline Parallelism~(VPP) vs. DualPipeV.}
For the pipeline dimension (with a pipeline degree of $8$), we adopt VPP ~\citep{narayanan2021efficient} rather than the bidirectional DualPipeV ~\citep{qi2025dual,deepseekai2025deepseekv3technicalreport} schedule, for two complementary reasons grounded in our profiling.
First, DualPipeV~\citep{qi2025dual,deepseekai2025deepseekv3technicalreport} is memory-prohibitive at this scale: to sustain its bidirectional overlap, each device must retain two model chunks, roughly doubling the resident parameters and gradients, together with a larger population of in-flight micro-batch activations.
Our traces show that, even under the more memory-frugal VPP schedule, the peak reserved device memory already reaches $51.7$--$54.6$\,GB per NPU across the profiled ranks, leaving only a narrow margin from the per-device HBM budget; the additional residency demanded by DualPipeV would readily overshoot this limit, forcing a smaller micro-batch size or heavier recomputation that erodes its purported efficiency gains.
Second, and more fundamentally, DualPipeV's headline capability, overlapping the expert-parallel all-to-all (MoE dispatch/combine) with computation, targets a cost that is simply not our bottleneck.
In our profiling, the all-to-all accounts for only $14.5\%$ of communication time, whereas the communication cost is dominated by pipeline point-to-point transfers ($36.7\%$ of communication, driven largely by receive-side stalls that reflect stage-level bubbles and load imbalance) and by tensor-/data-parallel \texttt{AllReduce} ($48.4\%$ of communication).

The pipeline-bubble is precisely what VPP mitigates by introducing fine-grained interleaving of virtual stages.
Consequently, DualPipeV would incur a substantial memory penalty in exchange for accelerating a component that contributes marginally to our end-to-end latency, whereas VPP suppresses the dominant pipeline bubbles while keeping the per-device footprint within the hardware budget and training stability. 
We will further investigate the optimal pipeline parallelism strategy in the near future.

\paragraph{Overlapped synchronization and computation for MoE token dispatch.}
\label{subsec:token_dispatcher}

\begin{figure}[t]
\centering
\includegraphics[width=\linewidth]{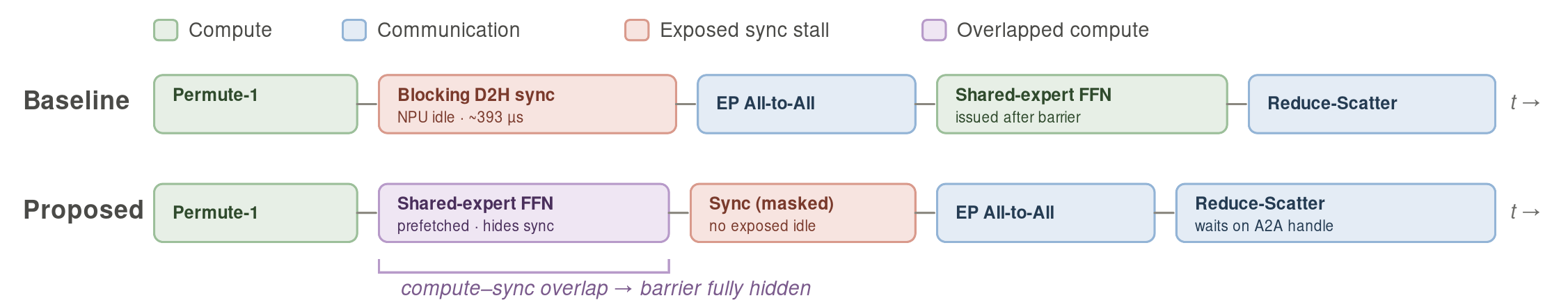}
\caption{Compute--communication schedule of the MoE token dispatcher. Issuing the shared-expert forward ahead of the \texttt{before\_ep\_alltoall} synchronization fills the exposed sync stall with useful computation while communication overlap is preserved.}
  \label{fig:cover_dispatch_sync}
\end{figure}

Expert parallelism turns routing into an all-to-all (A2A) communication in distributed LLM training. 
Our dispatcher structures one MoE layer as a pipeline (permutation, an A2A dispatch, grouped-GEMM expert computation, an A2A combine, and unpermutation), and applies three Ascend-centric choices to keep the collectives off the critical path.

For the MoE module, both token dispatch and combine are implemented using asynchronous all-to-all~(A2A) collectives. Synchronization is deferred until the communicated results are consumed, allowing independent computation to execute concurrently on separate streams and thereby reducing the exposed communication latency.
Specifically for Ascend-based MoE training, we further hide the synchronization stall preceding the dispatch operation, as illustrated in Figure~\ref{fig:cover_dispatch_sync}. Because the per-expert A2A split sizes cannot be determined until token routing is completed, resolving these dynamic split sizes introduces a visible \texttt{before\_ep\_alltoall} synchronization stall on the device timeline. The shared expert, however, depends only on the MoE-layer input and is independent of the routing-derived split sizes. We therefore advance its forward computation to overlap with this synchronization interval. This scheduling strategy fills an otherwise idle period with useful Cube computation while preserving the subsequent computation-communication overlap, effectively converting the barrier-induced latency identified in Section~\ref{sec:bottleneck} into productive execution.

\paragraph{Double-buffered Swap Optimizer.}
\label{subsec:swap_optimizer}
At the trillion-parameter scale, optimizer states become a dominant source of device-memory consumption. Distributed AdamW~\citep{loshchilov2017decoupled} maintains an FP32 master copy of the parameters together with the first- and second-moment states, amplifying the overall HBM usage compared to the model parameters alone. 
On Ascend NPUs, enabling swap optimizer alleviates the memory pressure.

A conventional swap optimizer mitigates this pressure by offloading optimizer states~\citep{ren2021zero}, together with the FP32 master parameters when configured, to pinned host memory and reloading them only when needed for parameter updates. Although this approach reduces the HBM footprint, it introduces serialized host-to-device and device-to-host transfers around each optimizer update. These transfers extend the optimizer step and can offset the performance benefits gained from the reclaimed memory.

To reduce this overhead, we implement a double-bufferedt swap optimizer that overlaps state transfers with optimizer computation. The swapped states are partitioned into chunks of
$\texttt{swap\_numel}
= N_{\mathrm{swap}}/\texttt{swap\_optimizer\_times}$ elements, where
$N_{\mathrm{swap}}$ denotes the total number of swapped parameters. While a fused AdamW kernel updates the current chunk, the next chunk is prefetched from host memory and the previously updated chunk is asynchronously written back through dedicated swap-in and swap-out streams. This pipelined schedule removes most transfer latency from the critical path. The chunk size controls the trade-off among the temporary HBM working set, transfer granularity, and achievable overlap. As a result, the optimizer substantially reduces HBM consumption with limited runtime overhead. The reclaimed memory can support more activations and larger micro-batches, thereby reducing activation recomputation and pipeline bubbles and increasing the useful-computation share $T_{\mathrm{comp}}$.

\subsection{Kernel Optimization for Accelerated LLM Training}
\label{sec:kernel_opt}

\subsubsection{Bottleneck Kernel Analysis}
\label{sec:kernel_bottleneck}

Following the training step-level decomposition in Section~\ref{sec:bottleneck}, we further investigate the device-compute bottlenecks at the kernel level. The profiling results reveal two distinct sources of inefficiency. \circled{1}~The  architecture-intrinsic kernels, including sparse attention, dense matrix multiplications, and grouped expert GEMMs, account for substantial device time within individual invocations. \circled{2}~Model-specific stages remain implemented as PyTorch-level compositions and are lowered by the \texttt{eager mode} into long sequences of casts, views, tensor-movement, and reductions, resulting in excessive kernel launches and fragmented execution. These observations motivate two complementary optimization directions: kernel-internal tuning for dominant compute-intensive kernels, and fusion or re-expression for launch-fragmented operator chains.

We organize the profiling results at the granularity at which optimization decisions can be made. Task Duration identifies individual kernels that dominate the device timeline and are therefore candidates for kernel-internal optimization. Kernel launch counts and submodule annotations reveal how high-level model stages are decomposed into low-level runtime operations. For AIV-dominated kernels, the utilization ratios of the vector-compute, scalar, MTE2 load, and MTE3 store pipelines further indicate whether execution is constrained by arithmetic, scalar control, or data movement. Taken together, these measurements distinguish compute-intensive standalone kernels from launch-fragmented operator chains and from memory-traffic bottlenecks introduced by the frontend implementation.

On the profiled NPUs, the trace contains $206{,}503$ compute-kernel launches and $39.84$\,s of aggregate device-side compute duration after excluding communication and AI-CPU execution. The resulting kernel profile exhibits four recurring patterns. The first is a small heavy head of architecture-intrinsic kernels whose latency is dominated by computation within the kernel body. The second is a launch-fragmented tail produced by PyTorch \texttt{eager computation}, where overhead accumulates across a large number of short-lived kernels. The third is a memory-bound vector tail, in which execution time is dominated by data movement rather than arithmetic. The fourth consists of generic Triton-generated kernels whose functional computation is correct, but whose lowering does not fully exploit on-chip data residency or pipeline scheduling. The first category primarily requires kernel-internal optimization, whereas the remaining categories motivate operator fusion or re-expression as AscendC kernels that retain intermediate tensors in Unified Buffer whenever possible. Table~\ref{tab:kernel_bottleneck} summarizes representative operators and their vector-pipeline characteristics. We first examine the dominant kernels and then turn to the fragmented and memory-bound long tail.

We begin with the dominant kernels, a small set of architecture-specific operations that account for a substantial fraction of device-compute time. 
In particular,  \texttt{SparseAttnSharedkvGrad} is the largest individual contributor, accounting for $16.86\%$ of the total kernel computation time due to the expensive gradient computation with respect to the DeepSeek sparse attention. 
Unlike ordinary operators~(e.g., Matrix Multiplication) which are thoroughly optimized, architecture-specific kernels exhibits high code-level complexity in hierarchical memory layout and computation pipelines.
This work addresses the challenge of kernel complexity via the proposed \textbf{AuraKernel} agent, as elaborated in Section~\ref{sec:kernel_agent}.


\begin{figure}[h]
	\centering
	\includegraphics[width=\textwidth]{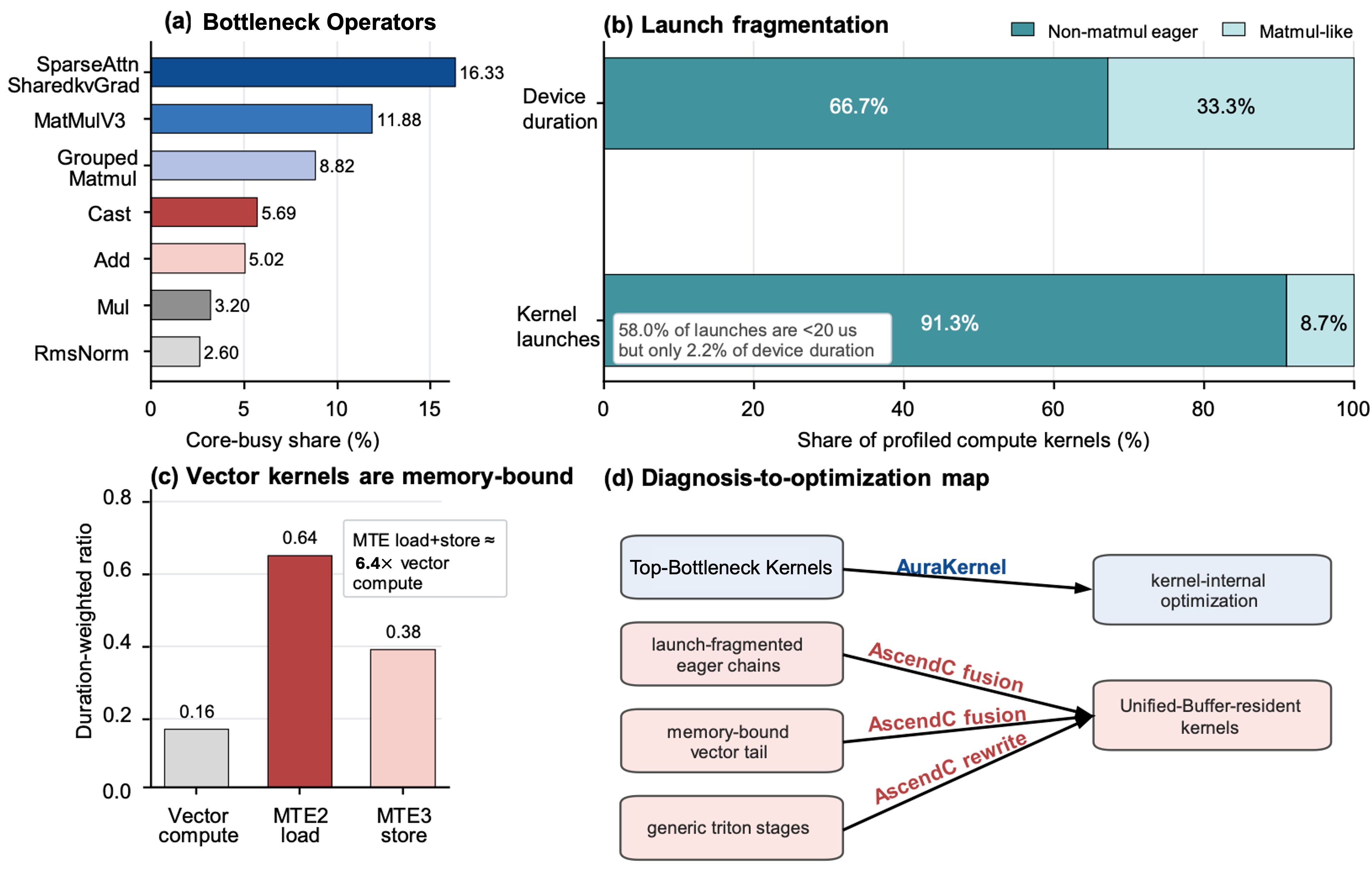}
	\caption{Bottleneck anatomy of one representative DeepSeek-V4-Pro training step on Ascend 910C ($206{,}503$ compute kernels, $39.84$\,s device-side duration). Panel~(a): the heavy head of architecture-intrinsic operators by core-busy share (the tool's ``Task Duration'' column)---sparse-attention backward, dense matrix multiplication, and grouped MoE multiplication. Panel~(b): the complementary launch-fragmentation view, contrasting the launch share of non-matmul eager operators with their much smaller share of device duration, and isolating the dense sub-$20\,\mu$s tail. Panel~(c): duration-weighted vector-pipeline ratios---vector compute well below the MTE2 load and MTE3 store ratios, whose combined memory movement is about $7.4\times$ the vector-compute share. Panel~(d): the resulting division of work---architecture-intrinsic heavy-head kernels routed to kernel-internal optimization by AuraKernel, and the launch-fragmented, memory-bound, and generic-Triton tails routed to re-expression as UB-resident AscendC kernels.}
	\label{fig:bottleneck_anatomy}
\end{figure}

The same trace also exposes a second execution regime. Rotary position embedding, weight-free RMS normalization, limited SwiGLU, and the mHC projection path are implemented as torch-level compositions, and autograd expands them into long sequences of generic operators. In the profiled step, non-matmul eager operators account for $91.3\%$ of launches but only $66.7\%$ of the Task Duration, against $33.3\%$ for the far fewer matmul-like operators (Figure~\ref{fig:bottleneck_anatomy}b). A dense small-kernel tail gives the regime its shape: $58.0\%$ of launches finish in under $20\,\mu$s yet together contribute only $2.2\%$ of Task Duration. The performance cost is therefore caused by repetition and by the boundaries between kernels: tensor materialization, dispatch, scheduling, and memory traffic are paid at every link in the chain.

The vector-pipeline counters show that the long tail is primarily constrained by data movement rather than arithmetic. Vector-core kernels account for $36.5\%$ of the total Task Duration, but their vector-compute utilization is only about $15\%$, substantially lower than the corresponding MTE load and store ratios of approximately $68\%$ and $40\%$, respectively (Figure~\ref{fig:bottleneck_anatomy}c). The imbalance indicates that most vector-kernel execution time is consumed by loading, storing, and rearranging data instead of actual computation.

The distribution of operators~(frequency and latency) further confirms that the long tail is dominated by fragmented data-movement operations. \texttt{Cast} is invoked $40{,}063$ times and contributes $6.02\%$ of the total Task Duration, while operators such as \texttt{ViewCopy}, \texttt{TensorMove}, \texttt{ZerosLike}, and \texttt{Slice} introduce additional overhead through frequent tensor materialization and buffer movement. These operations commonly arise when high-level backward computations, such as rotary embedding and limited SwiGLU, are decomposed into multiple small kernels under eager execution.

The bottleneck analysis motivates two separate kernel-optimization paths. Architecture-intrinsic kernels are optimized within the operator through improved tiling, data layout, scheduling, and pipeline overlap, which can be performed via agentic pipeline with long-context understanding~(Section~\ref{sec:kernel_agent}). On the other hand, the fragmented long tail is addressed by fusing and rewriting each target stage as a hardware-aware AscendC kernel. For eager autograd chains, this requires recovering the high-level computation and eliminating unnecessary data movement and type conversion introduced by framework execution. For existing Triton kernels, the computation is reimplemented in AscendC with explicit control over Unified Buffer~(UB) residency, tiling, and MTE scheduling. Both approaches reduce global-memory traffic by retaining intermediate results on-chip and consolidating the computation into fewer kernel passes. Section~\ref{sec:kernel_fusion} describes this fusion-and-rewrite path in detail.

\subsubsection{AuraKernel: Performance Optimization Agent for Complex Ascend Kernels}
\label{sec:kernel_agent}

\begin{figure}[t]
  \centering
  \includegraphics[width=1.0\textwidth]{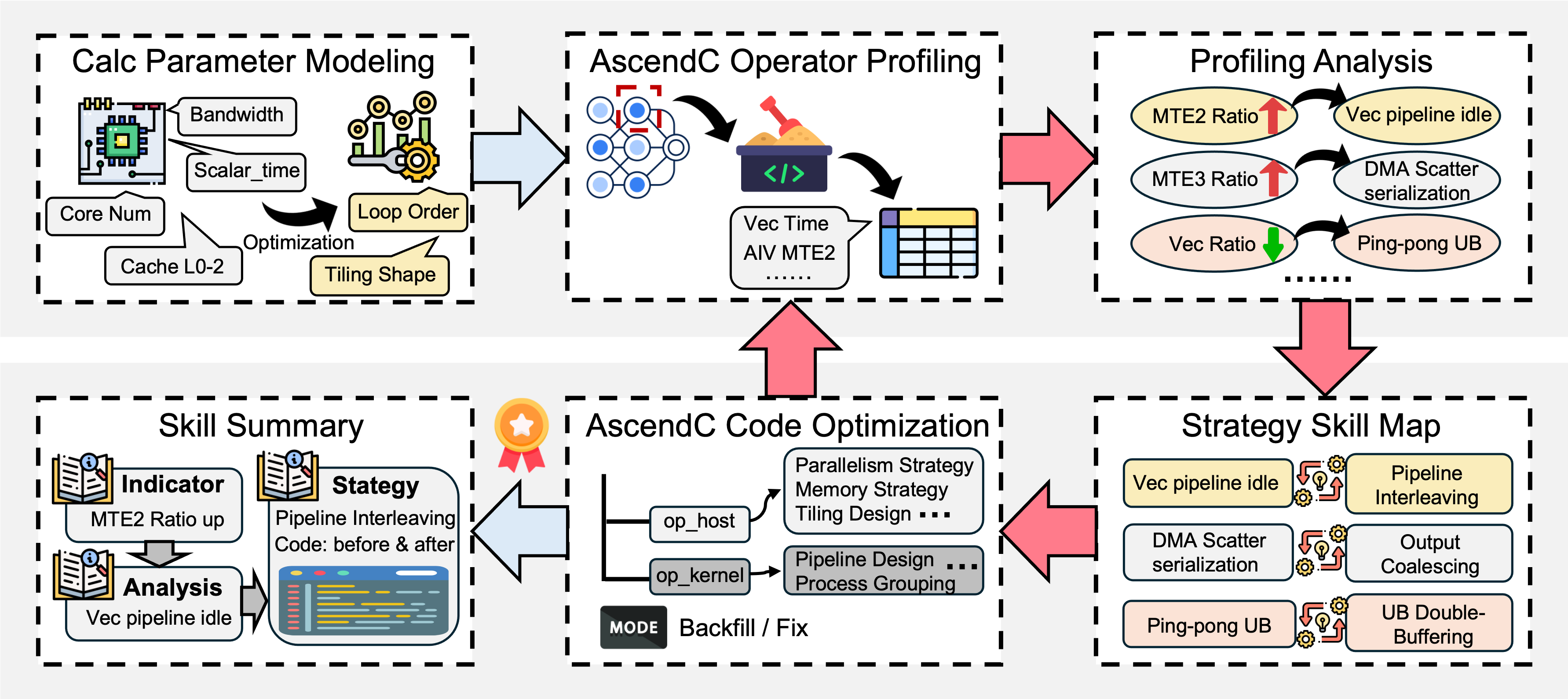}
  \caption{The architecture of AuraKernel. Starting from a functionally correct baseline kernel, AuraKernel generates an optimized operator in three stages. (1)~\textbf{OR-based tiling optimization} diagnoses the operator's dominant regime and instantiates a hardware-aware parameter space, yielding an initial configuration and a set of promising optimization directions. (2)~\textbf{Hardware-grounded iterative optimization} runs a closed modify--verify--profile loop: a harness compiles each candidate, gates it on correctness with the returnd profiling, while K-Search uses this feedback to explore a branchable, rollbackable tree of optimization directions and commits a checkpoint on every verified improvement. (3)~\textbf{Skill distillation} converts each verified gain into a reusable skill that is retrieved to guide hypothesis generation in later episodes.}
  \label{fig:aurakernel}
\end{figure}

Achieving high kernel efficiency on Ascend 910C is fundamentally different from optimizing CUDA kernels. 
Benefiting from the well-established open-source community with diverse data sources, CUDA-based kernel writing has been widely studied and used to train in recent LLM models~\citep{dao2022flashattention,dao2023flashattention2,shah2024flashattention3,tillet2019triton,hsu2024liger,ye2025flashinfer,wei2025astra,dong2025stark}. 
However, agent-based kernel optimization for non-CUDA hardware remains challenging due to uncommon hardware architecture and programming languages. 
With fine-grained memory control and pipeline, optimizing Ascend operators via agent system faces two critical challenges: 
\begin{itemize}
\item \textbf{Operator optimization with disaggregated host-kernel design:}
Unlike CUDA kernels, a complete AscendC operator package consists of \textbf{both} host-side scheduling logic and device-side kernel implementation, enabling flexible control over execution parameters such as tiling, memory movement, and launch configuration. Optimizing AscendC operators therefore requires the agent to reason beyond isolated kernel code and understand the explicit contract between the host and device sides.

\item \textbf{Long-context understanding with hierarchical agent memory:} 
Complex AscendC operators, such as sparse-attention kernels, often involve long-range code dependencies, multi-stage data movement, stream-level coordination, and pipeline scheduling. During iterative hardware-agent interaction, the agent must preserve and retrieve relevant design decisions, profiling observations, and correctness constraints across long optimization trajectories. Effective hierarchical memory is therefore essential for maintaining optimization consistency and avoiding repeated or conflicting modifications.

\item \textbf{Heuristic optimization leads to suboptimal efficiency:} 
Existing AscendC operator optimization still heavily relies on expert-crafted heuristics for tiling, buffering, and pipeline scheduling. However, these heuristics are often hardware-, shape-, and operator-specific, making them difficult to generalize across diverse workloads. 
As a result, purely heuristic optimization can easily converge to locally efficient but globally suboptimal configurations, motivating a more systematic optimization framework that combines Kernel agent with mathematical modeling~(for optimal initialization) followed by iterative hardware feedback.
\end{itemize}


Recent AscendC kernel agents focus on domain-specific training, DSL-guided generation, and profiling-in-the-loop search~\citep{ascendkernelgen2026,ascendcraft2026,wu2025ascendoptimizer}. However, production LLM training introduces a stricter setting: the target kernels must be optimized under real application shapes, memory pressure, layout constraints, and profiling signals from the end-to-end training workload rather than isolated microbenchmarks. 

In this work, we propose \textbf{AuraKernel}, a performance optimization agent for AscendC kernels.  Figure~\ref{fig:aurakernel} presents the system-level design of AuraKernel. 
The agent first builds a hardware-aware parameter model to constrain the optimization space; it then enters a profiling-driven loop that diagnoses bottleneck indicators, maps them to concrete optimization strategies, and applies code rewrites; finally, successful indicator--strategy--code patterns are distilled into reusable skills. 

\paragraph{Optimized Tiling Strategy via Operations Research.}
Given a pre-crafted AscendC kernel, AuraKernel initializes the optimization process with a solver-guided tiling strategy tailored to the target computation shape. Instead of relying solely on heuristic search, AuraKernel formulates operator tuning as a constrained performance-modeling problem. The optimizer determines how the workload is partitioned across AI Cores, how tiles are staged through GM, L2, L1, L0, and UB, how DMA transfers are overlapped with CUBE or vector execution, and how intermediate results are written back.

Given an operator shape, AuraKernel first identifies its dominant execution regime, such as compute-bound, memory-bound, scalar/control-bound, or communication-bound. It then instantiates a family-specific parameter space. For GEMM-like compute kernels, the parameters include L0 tile shapes, L1 blocking factors, core decomposition, loop ordering across memory levels, L0/L1 buffer counts, and ping-pong buffering settings. For vector- or reduction-style kernels, the same principle applies to row/block granularity, UB residency, reduction-axis placement, dtype-conversion points, and MTE2/MTE3 scheduling. In this way, the performance model serves as a structured interface between hardware constraints and agent actions.

Rather than optimizing a single opaque latency value, the model explicitly targets the bottleneck stage under a steady-state pipelined execution assumption:
\begin{equation}
\min_{\mathbf{p}\in\mathcal{P}(s)}
\max\Big\{T_{\mathrm{MTE1}}(\mathbf{p}),T_{\mathrm{MTE2}}(\mathbf{p}),
T_{\mathrm{MAC}}(\mathbf{p}),T_{\mathrm{FixPipe}}(\mathbf{p}),T_{\mathrm{scalar}}(\mathbf{p})\Big\},
\end{equation}
subject to data-coverage, L0/L1/UB capacity, legal core partitioning, alignment, and pipeline-validity constraints. Each stage cost is constructed as an explicit clock-cycle-level function of the tiling decision, defined as a transfer or instruction count multiplied by a cycles-per-unit constant, rather than as a single wall-clock latency sample. The per-cycle constants are fitted separately for each level of the on-chip storage hierarchy, with GM, L2, L1, and the L0A/L0B/L0C buffers each assigned a distinct, independently measured read and write bandwidth, so that a given transfer's cost is determined by the specific storage level it traverses rather than by a single averaged rate.

MTE1/MTE2 cycle counts are fitted under a systematic sweep of tensor shapes crossed with layout, transpose, and traversal-order configurations, with each configuration profiled over repeated trials and averaged to suppress measurement noise before cycles are regressed against transfer count and byte volume; an alignment constraint on transfer size is derived directly from this fit, following the observation that misaligned transfers cost disproportionately more cycles per byte moved. MAC cost is fitted as cycles per L0-tile element, and FixPipe cost as cycles per output-tile element under a given writeback layout, using the same regression procedure. Fitting error identified in specific shape regimes is corrected with explicit sub-cycle terms, adjusting fixed per-core launch overhead and per-instruction MAC cost so that the fitted cycle counts match measured behavior rather than being left unmodeled. The steady-state pipeline assumption behind the $\max\{\cdot\}$ formulation is itself derived from cycle-level traces of the double-buffered, ping-pong CUBE-core execution across the MTE1 $\to$ MTE2 $\to$ CUBE $\to$ FixPipe stage sequence, in which the four stages are observed to overlap rather than execute serially; their per-cycle occupancies are accordingly treated as concurrent, with the maximum taken rather than a serial sum. This cycle-level calibration is not uniformly exact, as cache hit rate at the L2 level is modeled more coarsely and is identified as a source of residual prediction error on some shapes, though the stage-cost functions themselves remain cycle-accurate constructions rather than opaque empirical latencies.

The solver output is therefore not merely a candidate tiling key, but also a diagnosis of the limiting factor: whether the operator is bottlenecked by movement from GM to L1, L2 reuse, CUBE utilization, writeback, or scalar overhead. This diagnosis is crucial for agentic optimization, because it prevents the agent from spending many iterations on plausible but irrelevant code edits.

The proposed Ascend solver is first validated on a common Matrix Multiplication (MatMul) kernel across frequent shapes in LLM models, with a $\mathbf{6.55\%}$ execution time reduction achieved by solely optimizing the tiling strategy. In other words, a solid starting point is created when the iterative kernel optimization process is initiated with optimized hyper-parameters, thereby avoiding the meaningless exploration and token waste that would otherwise occur.  


\paragraph{Hardware-Grounded Harness and Loop Engineering.}
The effectiveness of a kernel optimization agent depends on three critical foundations: \circled{1} a scalable and verifiable execution environment, \circled{2} a robust sampling mechanism with cross-iteration memory management, and \circled{3} a low-overhead harness that can orchestrate compilation, execution, verification, and profiling at scale. Without these components, the agent may either fail to validate candidate kernels reliably or spend most of its optimization budget on noisy measurements and redundant trials.

Considering these key factors, AuraKernel integrates an asynchronous sandbox system, cross-iteration optimization memory, and a dedicated orchestrator for parallel candidate sampling. The client side hosts the LLM decision loop, K-Search orchestrator, and persistent world model, while the backend performs source modification, compilation, execution, correctness verification, profiling, and version management on real Ascend NPUs. All hardware-related actions are exposed through stable JSON-over-HTTP tool calls, allowing the search policy or LLM provider to be changed without modifying the backend execution system.

\textit{\textbf{Compilation and execution harness with highly parallel sandbox services.}}
The backend provides a unified interface for both AscendC and Triton-Ascend operators. For AscendC, it supports the joint optimization of host-side tiling logic and device-side kernel code. For Triton-Ascend, it supports meta-parameter tuning such as grid decomposition, block size, pipeline stages, and other backend-specific configuration choices for Ascend NPUs. Kernel development and evaluation are conducted in a highly parallel sandbox system, which provides lightweight isolation for individual Ascend NPU executions while preserving scalable parallel evaluation across devices.

In particular, each sandbox maintains an isolated execution context, including runtime configuration, operator dependencies, temporary build artifacts, and profiling outputs, to reduce cross-candidate interference and improve measurement reproducibility. Meanwhile, independent sandbox instances can compile, execute, and profile multiple candidates concurrently. This design allows AuraKernel to turn expensive hardware feedback into a scalable optimization signal, enabling the agent to explore loop transformations, tiling strategies, buffering choices, and pipeline schedules under real execution constraints.
For each candidate kernel ($k\in\mathcal{K}$), the harness first compiles and executes the operator against a user-provided accuracy test that is read-only to the agent. Only implementations satisfying the functional-correctness predicate ($C(k)=1$) enter performance evaluation, which defines the feasible set
\begin{equation}
\mathcal{F} = \{k \in \mathcal{K} \mid C(k) = 1\}.
\end{equation}
Rather than using analytical estimates as the final selection criterion, AuraKernel optimizes the measured operator duration under representative application shapes. Specifically, the search process evaluates a sequence of candidate kernels ($(k_i)_{i=1}^{n}$) under a bounded profiling and LLM-call budget:
\begin{equation}
\sum_{i=1}^{n} b(k_i)\le B,
\end{equation}
and returns the best feasible candidate observed:
\begin{equation}
k^{\star}\in\operatorname*{arg\,min}_{k_i\in\mathcal{F}} T(k_i),
\end{equation}
where $T(k_i)$ denotes the measured end-to-end duration of one operator invocation and $b(k_i)$ denotes the evaluation cost of candidate ($k_i$). Each profiling pass additionally returns a metric vector:
\begin{equation}
\mathbf{m}(k)=
\big(m_{\mathrm{vec}},m_{\mathrm{cube}},m_{\mathrm{scalar}},
m_{\mathrm{mte2}},m_{\mathrm{mte3}},\dots\big)\in\mathbb{R}^{d},
\end{equation}
which characterizes vector or CUBE computation, scalar execution, and MTE2/MTE3 data movement. The measured duration determines whether a candidate improves the incumbent, whereas the metric vector explains why its performance changes and helps the agent distinguish compute-bound, memory-bound, and scalar-bound behavior.

Because hardware measurements inevitably contain noise, a candidate (k') replaces the incumbent ($k_b$) only when it passes both an improvement threshold and a deduplication check:
\begin{equation}
T(k')\le(1-\varepsilon)T(k_b)
\quad\text{and}\quad
h\big(\mathrm{canon}(k')\big)\notin\mathcal{H},
\qquad \varepsilon=0.02,
\end{equation}
where ($\mathrm{canon}(\cdot)$) removes comments and whitespace, (h) is a content hash, and ($\mathcal{H}$) stores hashes of previously evaluated candidates. The first condition rejects changes that are indistinguishable from profiling jitter, while the second avoids redundant evaluation of textually equivalent submissions. Accepted candidates are stored as versioned checkpoints together with their source code, correctness status, measured duration, and profiling metrics.

The harness also enforces the validity of the measured objective. A measurement-boundary rule forbids moving tensor computation from the measured kernel into the Python script or host launcher; an operator-kind lock prevents cross-paradigm replacement, such as rewriting a Triton-Ascend operator as AscendC when the task requires preserving the original implementation paradigm; These constraints prevent direct speedups obtained by changing the operator contract, bypassing unsupported hardware paths, or moving work outside the profiling boundary. Consequently, every optimization hypothesis is evaluated through the same reproducible sequence: modify the code, precision verification, candidate profiling, and behavior recording.

Unlike recent CUDA-oriented kernel agents~\citep{chen2026avo,mit_han_lab_kernel_design_agents_2026}, where the pre-defined knowledge bases play an important role in guiding optimization, our findings suggest that static knowledge bases and case-specific optimization skills can constrain the exploration space of LLM agents, especially on non-traditional computing platforms such as Ascend NPUs. 

In contrast, AuraKernel relies on self-evolved memory and a self-maintained Skills that are continuously updated through hardware-verified feedback. During iterative search, successful transformations, failed edits, profiling diagnoses, and versioned performance records are accumulated into persistent memory, allowing the agent to refine its optimization strategy across candidates rather than repeatedly relying on fixed, case-by-case rules. This feedback-driven memory mechanism enables AuraKernel to achieve direct performance improvements on both complex AscendC kernels and Triton-Ascend operators, as shown in Section~\ref{sec:experiment}.

\textit{\textbf{K-Search optimization-path exploration.}}
On top of this harness, we adopt K-Search~\citep{cao2026k} to organize candidate generation as a stateful, branchable, and rollbackable tree search, referred to hereafter as the world model, rather than a single linear ``view--edit--profile'' loop. Its root is the verified baseline, its internal nodes describe optimization directions such as tiling-granularity adjustment, MTE2 reduction, and pipeline interleaving, and its leaves are concrete code-change attempts. Each leaf is annotated with its predicted gain, confidence, implementation difficulty, visit count, correctness status, profiling result, and metric-vector transition.

At each iteration, K-Search selects a leaf from the current frontier $L(\mathcal{T})$ using an upper-confidence acquisition function that balances expected gain, exploration, and implementation difficulty:
\begin{equation}
\ell^{\star}=\operatorname*{arg\,max}_{\ell\in L(\mathcal{T})}
\left[
\underbrace{c_{\ell}\hat{g}_{\ell}}_{\text{expected gain}}
+\kappa\sqrt{\frac{\ln\!\big(1+N\big)}{1+n_{\ell}}}
-\gamma d_{\ell}
\right],
\label{eq:leaf-selection}
\end{equation}
where $\ell^{\star}$ denotes the leaf node from which further exploration is launched. $\hat{g}_{\ell}$ is the predicted speedup of leaf $\ell$, $c_{\ell}\in[0,1]$ is its confidence, $d_{\ell}$ is its estimated difficulty, $n_{\ell}$ is its visit count, $N=\sum_{\ell}n_{\ell}$, and $\kappa,\gamma>0$ control the trade-off between exploration and implementation cost. Each search cycle follows a continuous optimization loop: it retrieves relevant optimization experience, formulates a hypothesis, modifies the AscendC or Triton code, invokes the harness for compilation and correctness verification, profiles each valid candidate, and updates the best-known checkpoint when the measured improvement satisfies the acceptance criterion. Let $(T(k_i))$ denote the measured execution time of the $(i)$-th profiled candidate. The best observed execution time after profiling $(n)$ candidates is:
\begin{equation}
T_n^{\star}=\min_{i\le n}T(k_i),
\label{eq}
\end{equation}
which is non-increasing with $n$, even though candidates may regress or fail during exploration.

K-Search further calibrates its optimization priors through sibling-priority back-propagation. After a leaf $\ell$ is evaluated, its realized gain $g_{\ell}$ is compared with its predicted gain $\hat{g}_{\ell}$, and the priorities of related sibling directions are updated as
\begin{equation}
\mathrm{score}(\ell')\leftarrow\mathrm{score}(\ell')
+\eta\Big(
\mathbbm{1}[r_{\ell}>\tau_{\uparrow}]
-\mathbbm{1}[r_{\ell}<\tau_{\downarrow}]
\Big),
\qquad
r_{\ell}=\frac{g_{\ell}}{\hat{g}_{\ell}},
\label{eq:sibling-update}
\end{equation}
for every $\ell'\in\mathrm{sib}(\ell)$, with $\eta=0.1$, $\tau_{\uparrow}=1.5$, and $\tau_{\downarrow}=0.5$. Here, $\mathrm{score}(\ell')$ instantiates the confidence weight $c_{\ell'}$ used in the acquisition function of Eq.~\eqref{eq:leaf-selection}, so promoting or demoting a sibling directly raises or lowers its $c_{\ell'}\hat{g}_{\ell'}$ term in subsequent applications of Eq.~\eqref{eq:leaf-selection}. Directions related to an unexpectedly effective candidate are promoted, whereas those associated with an underperforming candidate are demoted, and this adjustment is what carries the evaluated leaf's outcome into the selection score of its still-unevaluated siblings. The WorldModel is thus continuously recalibrated by measurements from the actual NPU rather than by the LLM's initial confidence alone.

When a path fails correctness checks, regresses in latency, or reaches a performance plateau, K-Search rolls back to a previously verified checkpoint and opens a new branch. This enables non-monotonic exploration of the coupled tiling, buffering, memory-layout, and pipeline-scheduling space while protecting the global incumbent. Each cycle is governed by limits on profiling calls, tool calls, and consecutive non-improving attempts, together with a global cycle cap. For searches spanning dozens of cycles, a three-layer context-compression mechanism keeps the transcript tractable without discarding recoverable state: a failed profiling round, spanning every message since the last successful profile, is collapsed into a single summary recording the failure cause and the last verified checkpoint; repeated views of the same file are replaced by a compact marker referencing the most recent view, keyed by file and board; and repeated directory listings are deduplicated the same way, keyed by operator and board. The harness and K-Search therefore form complementary halves of the same loop: the former establishes whether a modification is correct and measurably faster, while the latter determines which optimization direction to attempt next and how hardware evidence should reshape the remaining search.

\paragraph{Skill distillation and continual improvement.}
For each checkpoint that achieves a verified gain, the agent distills a structured optimization \emph{skill} from the metric-vector transition observed before and after the code change, represented as a triple
\begin{equation}
\sigma=\big(\mathbf{m}_{\mathrm{before}},\,\Delta\mathbf{m},\,\pi\big),
\qquad \Delta\mathbf{m}=\mathbf{m}_{\mathrm{after}}-\mathbf{m}_{\mathrm{before}},
\label{eq:skill-tuple}
\end{equation}
where $\pi$ is the applied code transformation. Given the current metric vector $\mathbf{m}$, the agent retrieves stored skills whose pre-image best matches it and conditions hypothesis generation on a temperature-controlled softmax over similarities,
\begin{equation}
w_j\;\propto\;\exp\!\Big(\tfrac{1}{\beta}\,\mathrm{sim}\big(\mathbf{m},\,\mathbf{m}^{(j)}_{\mathrm{before}}\big)\Big),
\qquad
\mathrm{sim}(\mathbf{a},\mathbf{b})=\frac{\langle\mathbf{a},\mathbf{b}\rangle}{\lVert\mathbf{a}\rVert\,\lVert\mathbf{b}\rVert},
\label{eq:skill-weight}
\end{equation}
with temperature $\beta>0$. These skills are stored in an experience bank and retrieved to condition LLM hypothesis generation in subsequent episodes, so the agent progressively accumulates transferable AscendC optimization knowledge as the bank grows. All runs emit complete, replayable artifacts---including the serialized world model---so an optimization can be resumed from an intermediate cycle on another machine. In effect, AuraKernel recasts kernel optimization as an automatable search problem in which the LLM's hypothesis generation and the NPU's objective measurement continually calibrate each other, converging toward near-optimal AscendC implementations within a strict budget. 

Applied to DeepSeek-V4-Pro training (Section~\ref{subsec:eval_aurakernel}), AuraKernel accelerates the shared-key-value sparse-attention backward step, the single largest device-time contributor, by up to $1.24\times$ in its end-to-end training context, and in a single campaign optimizes $14$ reduction-heavy framework operators to full correctness with a $2.06\times$ average and an $11.90\times$ best speedup. Consistent with our bottleneck analysis, these gains come from cutting redundant global-memory traffic rather than from raising raw compute utilization.

\subsubsection{Ascend NPU-aware Kernel Fusion and Rewrite}
\label{sec:kernel_fusion}
The second optimization path addresses the three non-head patterns of the bottleneck analysis---the launch-fragmented eager tail, the memory-bound vector tail, and the generic Triton kernels. Whereas AuraKernel improves the body of already-large operators, this path re-expresses each target stage as one or a small number of hardware-affine AscendC kernels. The first two patterns reach the runtime as eager Python and torch chains and the third as generic Triton kernels, so all three collapse onto the two entries introduced above. For the eager chains, fusion recovers the high-level operation from the trace, classifies the emitted kernels into model arithmetic, shape bookkeeping, data movement, precision conversion, and autograd writeback, and collapses the chain. For the Triton kernels, the arithmetic is unchanged but the operator is rebuilt in AscendC with the on-chip residency and scheduling the generic lowering cannot express. The two entries overlap: the weight-free RMS normalization and the mHC pre-projection contraction are simultaneously fused and rewritten, because their Triton kernels also carried avoidable stage boundaries. The system-level communication and pipeline bottlenecks remain outside this scope. In both cases the arithmetic is unchanged, and only the substrate on which it runs changes.

On a high level, the mathematical contract records the formula that the fused implementation must preserve, including clamp subgradients, complex RoPE rotation, weighted expansion, reduction, and dtype-sensitive accumulation. The frontend-artifact record identifies operators that appear in the runtime trace because of the Python expression, view semantics, dtype promotion, or autograd expansion, such as casts, views, fills, tensor moves, concatenations, and scatter-style writebacks. The hardware execution contract specifies the Ascend plan: which tensors are resident in UB, how GM--UB movement is scheduled through MTE2/MTE3, where reductions or broadcasts occur, and which dtype is carried at each boundary. This structure lets each fusion be reviewed as a mathematics-preserving rewrite rather than as a collection of local peephole optimizations.

\begin{figure}[h]
\centering
\includegraphics[width=\textwidth]{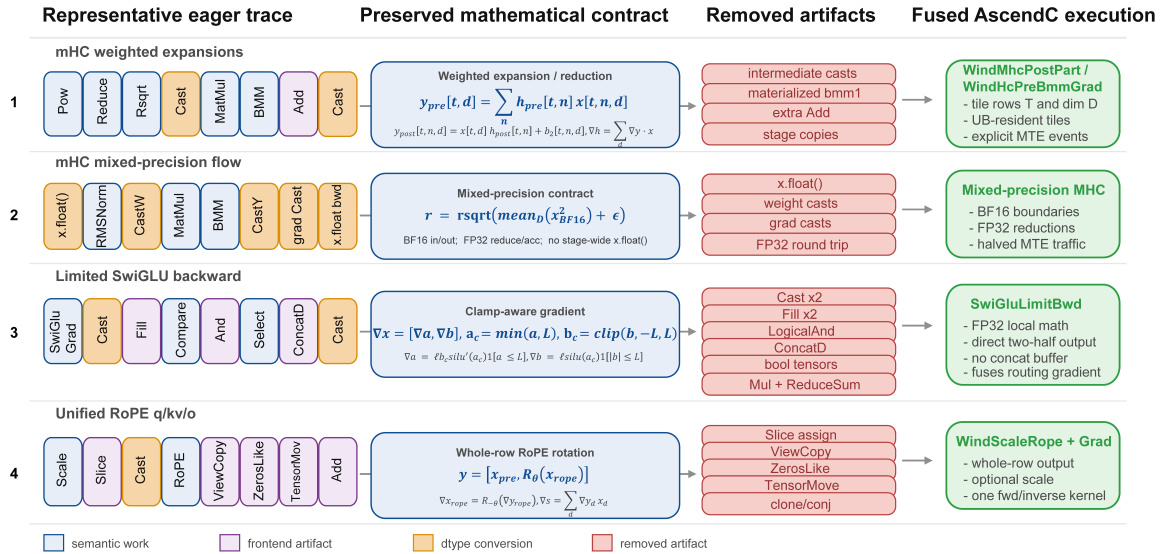}
\caption{Operator-chain fusion atlas. Rows show four DeepSeek-V4-Pro rewrites: mHC weighted expansion, mHC mixed-precision dataflow, limited SwiGLU backward (with routing-weight gradient) and unified RoPE. Columns give the emitted trace, semantic invariant, removed frontend artifacts and fused AscendC boundary.}
\label{fig:operator_chain_fusion_atlas}
\end{figure}

The resulting kernels follow three hardware-facing design rules. First, intermediate tensors remain inside the Unified Buffer whenever their only role is to connect adjacent eager operators. Second, the kernel interface returns whole rows or whole stage outputs, so that slice-assignment and scatter-style autograd expansions are not generated in the first place. Third, dtype transitions follow the numerical contract of the stage rather than the conservative dtype inherited from a frontend expression. These rules do not remove model arithmetic. Instead, they change where the arithmetic is executed and how many times the surrounding memory traffic is paid. We apply this method to three representative targets, presented in the order used throughout the evaluation: the mHC compression suite, limited SwiGLU backward, and unified rotary embedding. A mixed-precision dataflow rewrite of the mHC pre-projection path is developed alongside the first, and appears as its own row in the atlas below. Figure~\ref{fig:operator_chain_fusion_atlas} serves as the method overview for these rewrites. It separates each case into the observed trace, the semantic contract used to audit correctness, the frontend artifacts removed by fusion, and the AscendC boundary that implements the revised dataflow.

\begin{figure}[h]
\centering
\includegraphics[width=\textwidth]{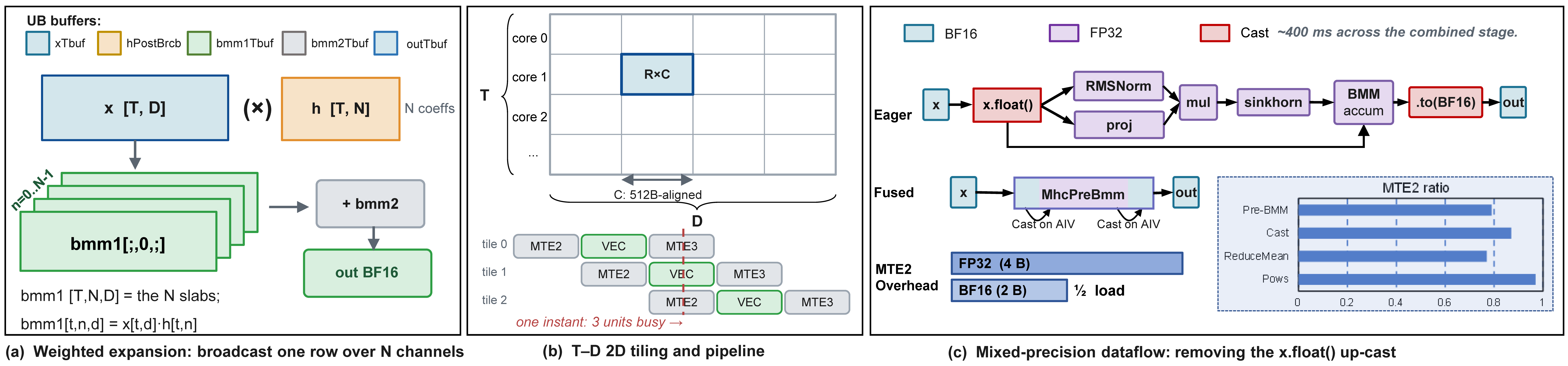}
\caption{Three mechanisms of the mHC weighted-expansion fusion, detailed in the text. \textbf{(a)}~The weighted expansion $\mathrm{bmm1}[t,n,d]=x[t,d]\,h[t,n]$ over the $N$ mixture channels, accumulated and emitted as BF16 in one UB-resident kernel. \textbf{(b)}~The two-dimensional $[T,N,D]$ tiling and the three-stage MTE2/VEC/MTE3 pipeline, where the dashed line marks the instant at which all three engines are busy. \textbf{(c)}~The pre-projection dataflow, in which \texttt{x.float()} forks into RMS normalization and the projection, whose product feeds the Sinkhorn step and then the BMM. The inset gives the per-operator MTE2 ratios that make the chain memory-bound.}
\label{fig:mhc_mechanisms}
\end{figure}

\paragraph{mHC fusion suite.}
The Manifold-Constrained Hyper-Connections (mHC) path~\citep{mhc2025} is a larger instance of the same problem, and it is not represented by a single fused operator. It also exercises both entries of this path at once: the weight-free RMS normalization and the pre-projection contraction originated as generic Triton kernels, so their rewrite into AscendC is a substrate change, while the surrounding casts and elementwise accumulation are eager fragments removed by a stage-boundary fusion. Its pre-projection stage combines weight-free RMS normalization, a small projection whose output is normalized by a Sinkhorn step into per-channel mixture weights, and a batched weighted expansion driven by those weights. Its post-projection stage combines another weighted expansion with an accumulated tensor and a final BF16 output. In the original trace, these semantics appear as \texttt{Pows}/\texttt{ReduceMean}/\texttt{Add}/\texttt{Rsqrt}, \texttt{Cast}, \texttt{MatMul}, custom BMM kernels, and elementwise accumulation. The cost is spread across normalization, conversion, and movement around the projection rather than concentrated in one matmul. Figure~\ref{fig:mhc_mechanisms} summarizes the three mechanisms that make this path efficient on the NPU: the weighted expansion (a), the two-dimensional tiling and pipeline (b), and the mixed-precision dataflow (c).

The purpose of the mHC suite is to expose the compression path as a small number of stable stage boundaries rather than as several unrelated peephole fusions. Table~\ref{tab:mhc_fusion_suite} lists the five kernels and the contract each one preserves.

\begin{table}[h]
\centering
\caption{mHC fusion suite. Each row identifies a concrete AscendC kernel, the frontend-emitted chain it replaces, the mathematical contract preserved by the fused implementation, and the hardware-facing boundary used by the kernel.}
\label{tab:mhc_fusion_suite}
\resizebox{\textwidth}{!}{%
\begin{tabular}{llll}
\toprule
Kernel & Replaced chain & Preserved mathematical contract & Hardware boundary \\
\midrule
\texttt{WindMhcPostPart} &
\makecell[l]{\texttt{Cast} + post-BMM1\\+ \texttt{Add} + \texttt{Cast}} &
\makecell[l]{$y_{\mathrm{post}}[t,n,d] =$\\$x[t,d]h_{\mathrm{post}}[t,n]+b_2[t,n,d]$} &
\makecell[l]{Tile over rows $T$ and feature dim $D$;\\UB-resident expansion and direct BF16 output} \\
\texttt{wind\_rms\_norm\_without\_weight} &
\makecell[l]{\texttt{Pows} + \texttt{ReduceMean}\\+ \texttt{Add} + \texttt{Rsqrt}} &
\makecell[l]{$r[t]=\mathrm{rsqrt}(\frac{1}{D}\sum_d x[t,d]^2+\epsilon)$} &
\makecell[l]{Row-wise reduction in one local dataflow;\\FP32 reciprocal-scale output} \\
\texttt{WindHcPreBmmForward} &
\makecell[l]{pre-BMM forward casts +\\weighted-expansion contraction} &
\makecell[l]{$y[t,d]=\sum_{n=0}^{N-1} h_{\mathrm{pre}}[t,n]\,x[t,n,d]$\\$(N{=}4)$} &
\makecell[l]{Shared $T$--$D$ tiling; \texttt{Brcb}-broadcast weights,\\FP32 $h_{\mathrm{pre}}$ / BF16 $x$, BF16 output} \\
\texttt{wind\_rms\_norm\_without\_weight\_backward} &
\makecell[l]{\texttt{Pows}/\texttt{Mul}/\texttt{Div}\\RMS backward fragments} &
\makecell[l]{$\nabla x[t,d]=-\nabla r[t]r[t]^3x[t,d]/D$} &
\makecell[l]{Compute row scalar once;\\expand across $D$ inside the tile} \\
\texttt{WindHcPreBmmBackward} &
\makecell[l]{pre-BMM backward\\casts, zero/init, reduce fragments} &
\makecell[l]{$\nabla x[t,n,d]=\nabla y[t,d]h_{\mathrm{pre}}[t,n]$\\$\nabla h_{\mathrm{pre}}[t,n]=\sum_d\nabla y[t,d]x[t,n,d]$} &
\makecell[l]{Shared $T$--$D$ tiling for weighted expansion\\and reduction to $\nabla h_{\mathrm{pre}}$} \\
\bottomrule
\end{tabular}}
\end{table}

\texttt{WindMhcPostPart} handles the post-projection stage. It flattens the batch and sequence dimensions into $T=BS$, computes the weighted expansion $x[t,d]h_{\mathrm{post}}[t,n]$, accumulates the existing $b_2$ tensor, and emits BF16 output directly. The kernel tiles over rows $T$ and hidden dimension $D$: rows are split across the cores to provide independent token-level work, while $D$ is tiled into $512$\,B-aligned blocks that bound UB residency and MTE transfer granularity. This two-dimensional tiling is the common hardware contract across the mHC kernels. It lets the implementation reuse local buffers across expansion, accumulation, and output conversion, with MTE2/VEC/MTE3 movement scheduled inside the fused operator rather than by framework-level kernel boundaries.

The pre-projection side applies the same boundary principle to reductions and weighted expansion, and it is also where the substrate rewrite is most explicit: the weight-free RMSNorm and the pre-BMM contraction were previously generic Triton kernels, and rebuilding them in AscendC is what exposes the on-chip control the affinity argument relies on. The RMS-normalization forward and backward kernels fuse the square, reduction, epsilon addition, reciprocal square root, and gradient expansion so that the row scalar is computed once and reused locally. \texttt{WindHcPreBmmForward} carries the forward weighted expansion $y[t,d]=\sum_{n} h_{\mathrm{pre}}[t,n]\,x[t,n,d]$ under a $T$--$D$ tiling, broadcasting each FP32 $h_{\mathrm{pre}}$ coefficient with a \texttt{Brcb} instruction and consuming BF16 activations, so that the four-channel contraction is completed in one UB-resident pass instead of a matmul-plus-cast chain. A single vector multiply then applies one broadcast coefficient across a full register of feature elements, and a three-stage MTE2/VEC/MTE3 pipeline with ping-pong buffering overlaps the load of the next tile with the compute and store of the current one---the explicit residency, broadcast, and pipeline scheduling that a generic Triton lowering leaves to global-memory round trips. This substrate change moves the pre-BMM forward from movement-bound to compute-bound, and Section~\ref{subsec:eval_operator_chain_fusion} reports the measured speedup and the corresponding rise in vector-compute occupancy. \texttt{WindHcPreBmmBackward} then computes both outputs of the weighted expansion: $\nabla_x$ is formed by broadcasting $h_{\mathrm{pre}}$ across the feature dimension and multiplying by $\nabla_y$, while $\nabla h_{\mathrm{pre}}$ is obtained by reducing $\nabla y\cdot x$ across $D$. These kernels convert the mHC path from a sequence of frontend-generated fragments into a stage-level dataflow whose memory movement is scheduled once.

\paragraph{Backward of Limited SwiGLU.}
The limited SwiGLU activation adds a pre-SiLU clamp to the standard gated product: the gate half is clipped by an upper bound and the up half is clipped symmetrically before multiplication. In eager form, the backward pass expands this compact mathematical rule into a chain containing \texttt{SwiGluGrad}, two \texttt{Cast} kernels, scalar \texttt{Fill} kernels, three comparisons, \texttt{LogicalAnd}, two \texttt{SelectV2} kernels, and \texttt{ConcatD}. The trace shows a roughly $401\,\mu$s memory-bound chain per invocation. Most of that chain does not introduce new model arithmetic, but instead materializes clamp constants, boolean masks, dtype round trips, and the concatenation needed to rebuild the gradient tensor.

The fused kernel implements the closed-form backward directly. Let $x=[a,b]$, where $a$ is the gate half and $b$ is the up half, and let $\ell$ be the incoming gradient. With $a_c=\min(a,\mathrm{limit})$, $b_c=\mathrm{clip}(b,-\mathrm{limit},\mathrm{limit})$, and $\sigma=\sigma(a_c)$, the kernel computes
\begin{align}
  \nabla_a &=\ell \cdot b_c \cdot \left(\sigma+a_c\sigma(1-\sigma)\right)\cdot \mathbf{1}[a\le \mathrm{limit}], \\
  \nabla_b &=\ell \cdot a_c\sigma \cdot \mathbf{1}[-\mathrm{limit}\le b\le \mathrm{limit}].
\end{align}
The fused boundary writes the two gradient halves directly to their output offsets, removing the concatenation and the frontend-visible boolean-composition chain. BF16/FP16 inputs are promoted once for the local derivative computation, while the sigmoid, clamp masks, and intermediate products remain in UB and the result is converted only at the output boundary. We verified the formula symbolically and against a PyTorch autograd golden reference, including clamp-boundary cases. Unsupported dtypes and non-positive limits fall back to the original path.

\begin{figure}[h]
\centering
\includegraphics[width=0.96\textwidth]{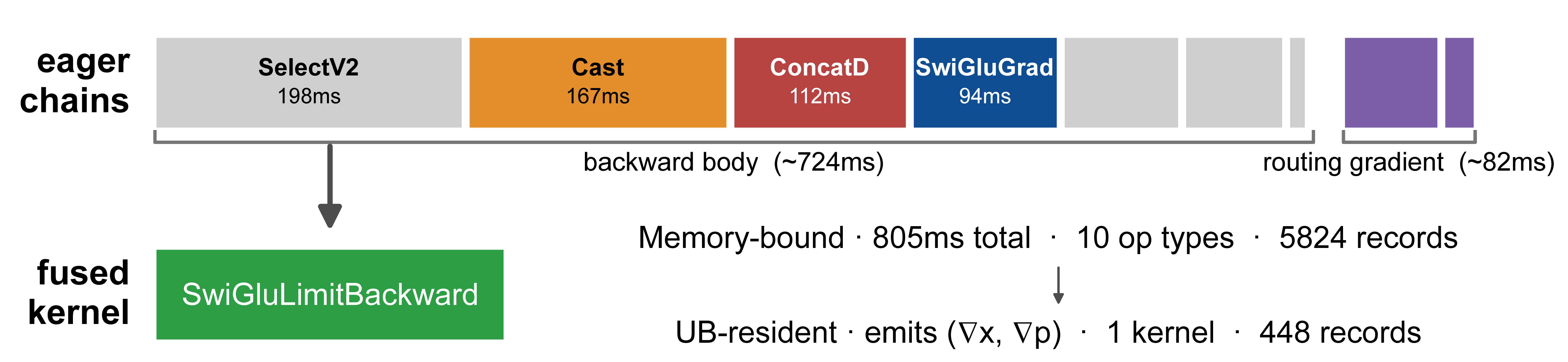}
\caption{Schematic of the limited-SwiGLU backward chain collapsing into one fused kernel. The eager row shows the operator types emitted by autograd: \texttt{SelectV2}, \texttt{Cast}, \texttt{ConcatD}, \texttt{SwiGluGrad}, a trailing group of comparison, mask, and fill kernels (\texttt{LessEqual}, \texttt{GreaterEqual}, \texttt{LogicalAnd}, \texttt{Fill}), and, in the MoE path, the routing-weight-gradient reduction (\texttt{Mul} and \texttt{ReduceSum}), with tile widths proportional to their per-type trace time. Fusion replaces the entire chain with a single UB-resident \texttt{SwiGluLimitBackwardV2} kernel that keeps the sigmoid, clamp masks, intermediate products, and routing-gradient reduction on chip and never materializes them to global memory.}
\label{fig:fusion_trace_schematic}
\end{figure}

In the MoE setting the same kernel also fuses the routing-gradient reduction, which the eager path emits as its own chain. Each expert activation is scaled by a per-token routing weight $p$, so the layer computes $\mathrm{out}=h\cdot p$ with $h$ the limited-SwiGLU output, and the routing network requires $\nabla p=\sum_d \ell[t,d]\,h[t,d]$, a reduction over the hidden dimension. In eager form this reduction is three further kernels: a reload of $h$, a full-size elementwise product $\ell\cdot h$, and a \texttt{ReduceSum}, each paying its own global-memory round trip. The fused kernel instead forms $\nabla p$ inside the same pass that computes the activation gradient. The values it needs, $\sigma(a_c)$, $a_c$, and $\mathrm{clip}(b)$, are already resident in UB, so the kernel reconstructs $h$ locally and reduces it row by row into $\nabla p$ without materializing $\ell\cdot h$ or issuing any extra load. Because $\nabla p$ depends on the unweighted $\ell$ while $\nabla x$ depends on $\ell\cdot p$, the reduction is taken before the incoming gradient is scaled by the routing weight. The reconstructed $h$ is kept in FP32 within the kernel, which matches the reference training path and avoids the BF16 reload that the eager chain reads back from global memory. When the hidden row is split across column tiles, each tile contributes a partial sum that is combined into the row total by a global-memory atomic accumulation. The routing gradient is validated against a FP32 autograd reference by relative $L_2$ error, appropriate for a reduction output whose per-element accumulation order differs from the reference. In the profiled step the eager routing-gradient chain is a full-size $\ell\cdot h$ product and a reduction that together cost about $82$\,ms. Both vanish once the reduction moves inside the fused kernel, and Section~\ref{subsec:eval_operator_chain_fusion} reports the combined effect on the backward stage. Figure~\ref{fig:fusion_trace_schematic} shows the whole backward chain collapsing into this single kernel.

\begin{figure}[h]
\centering
\includegraphics[width=\textwidth]{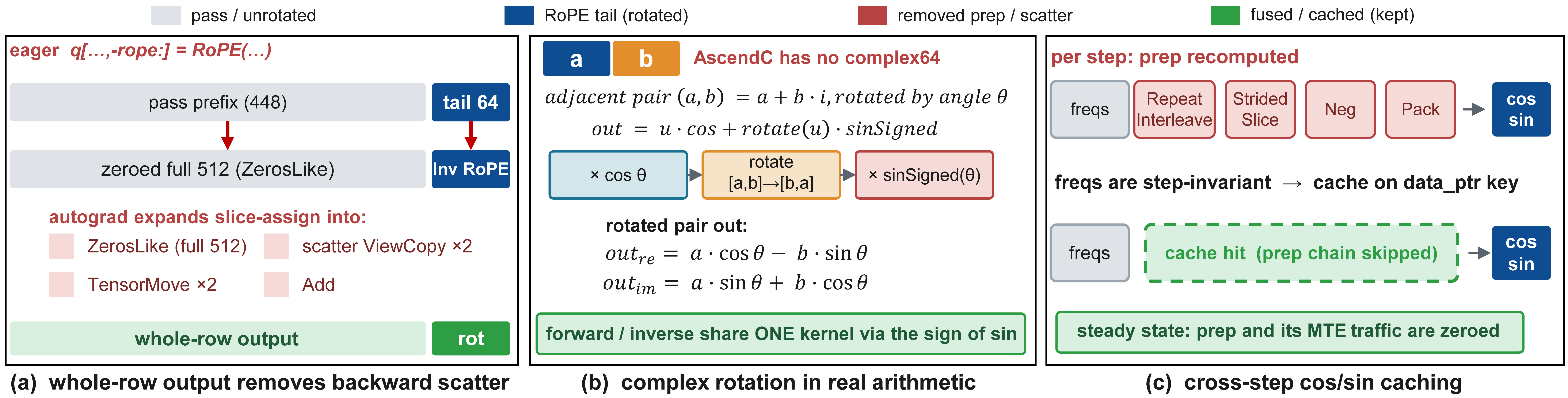}
\caption{Three mechanisms of the unified rotary-embedding operator. (a)~The eager in-place slice assignment and the full-width backward scatter it forces, contrasted with the whole-row output that avoids it. (b)~The real-arithmetic realization of the complex rotation, showing the cosine multiply, adjacent-pair swap, and sign-flipped sine that produce the rotated pair. (c)~The per-step cosine/sine preparation chain against its cached counterpart, where a step-invariant frequency table lets the chain be skipped after the first step.}
\label{fig:rope_mechanisms}
\end{figure}

\paragraph{Unified rotary embedding.}
Rotary position embedding appears in three attention sites with different surrounding context. The query path first multiplies by the exposed RMS-normalization scale and then rotates the RoPE tail. The compressed key-value path applies RoPE without an external scale and uses local frequencies. The output-projection path applies the inverse rotation and previously relied on a clone plus conjugate construction. These sites differ in frontend expression, but they share the same stage boundary: preserve a full-row prefix, rotate the RoPE tail, and return the row without exposing slice assignment to autograd. In the backward pass, exposing that assignment produces \texttt{ZerosLike}, \texttt{TensorMove}, \texttt{ViewCopy}, \texttt{Cast}, and \texttt{Add} chains around the actual rotation gradient. Figure~\ref{fig:rope_mechanisms} summarizes the three mechanisms the unified operator applies.

We implement the three sites with one forward operator and one backward operator. The operator returns the pass-through prefix and the rotated tail as a whole-row output, so no slice assignment is exposed to autograd and the backward scatter chain is never generated (Figure~\ref{fig:rope_mechanisms}a). Because the target hardware has no native complex type, the operator reproduces the complex rotation in real arithmetic~\cite{roformer}: it treats each adjacent element pair as one complex number and rotates it with a cosine multiply, an adjacent-pair swap, and a sign-flipped sine, so that a single vector code path realizes the complex product without a complex-construction chain (Figure~\ref{fig:rope_mechanisms}b). Forward and inverse rotation share one kernel through the sign of the sine term, and the same kernel serves q, kv, and o because site-specific differences reduce to scale, frequency, and direction arguments rather than separate operator implementations.

The backward kernel follows the same boundary choice, keeping the rotation derivative and the scale-gradient reduction inside the fused operator. The query path still feeds the mathematically required RMS-normalization backward and preserves the final accumulation of the two gradient branches. In isolated per-site validation, the unified operator covers q, kv, and o with relative error within the accepted training tolerance.

The operator consumes real cosine and sine tensors, which are produced from the complex frequency table by a preparation chain of \texttt{RepeatInterleave}, \texttt{StridedSlice}, \texttt{Neg}, and \texttt{Pack}. Left in the graph, this chain re-executes every step, even though the frequencies depend only on position and do not change across steps. We therefore cache the converted tensors under a key derived from the frequency table, so that after the first step the cache hits and the entire preparation chain, together with its memory-movement traffic, is skipped (Figure~\ref{fig:rope_mechanisms}c). The cache is keyed so that a stable frequency table across steps keeps the downstream lookup valid, which is what lets the preparation cost fall to zero in steady state.

\paragraph{Mixed-precision dataflow.}
The mHC pre-projection path also contained a stage-wide precision artifact. The Python implementation performed an early \texttt{x.float()} after flattening the input, which promoted the activation tensor from BF16 to FP32 before RMS normalization, the projection, and the BMM. Because the surrounding training path is BF16 mixed precision, this promotion forced a full-stage round trip: activations, gradients, and intermediate accumulations were repeatedly converted or moved at FP32 width, and the final outputs were converted back. In the profile, the mHC pre-projection stage is invoked hundreds of times per step (896 forward and 448 backward calls); the \texttt{x.float()} choice therefore amplifies into visible \texttt{Cast} kernels and doubled transfer volume across the forward and backward chains. Crucially, this stage is memory-bound rather than compute-bound: its dominant operators run at an AIV vector-compute ratio of only $\approx 0.06$ against an MTE2 load ratio of $\approx 0.87$ (\texttt{Pows} $0.97$, \texttt{ReduceMean} $0.77$, \texttt{Cast} $0.87$; Figure~\ref{fig:mhc_mechanisms}c), so the wall time is set by how many bytes cross global memory, and promoting the whole chain to FP32 doubles exactly the quantity that bounds it. The single largest mover is the entry \texttt{Cast} from BF16 to FP32, at $222$\,ms per step.

We remove the up-cast by making the mixed-precision contract explicit, and the central move is to shift the dtype at which operands cross global memory. Because the chain is MTE2-bound, the kernels load activations from global memory in BF16 and convert them to FP32 inside the vector core, rather than loading an already-promoted FP32 tensor. This halves the load volume of exactly the pipe that bounds the stage while preserving accuracy, since the FP32 conversion still happens before any reduction. The RMS-normalization kernels accept BF16 activations while retaining FP32 reciprocal-square-root values and FP32 internal arithmetic where the reduction requires it. The projection computes BF16$\times$BF16 products with FP32 accumulation and emits the dtype required by the downstream stage. The pre-BMM forward consumes $h_{\mathrm{pre}}$ in FP32 and $x$ in BF16, and returns BF16 output; its backward returns BF16 $\nabla_x$ and FP32 $\nabla h_{\mathrm{pre}}$. The weight-gradient path similarly returns  gradient with the matched data type as the original BF16 weight, rather than retaining the FP32 tensors because the corresponding activation was upcast for computation.

This change is small in code but broad in effect because it changes the dtype carried by the whole mHC subgraph. The analysis identifies each removed or narrowed movement: the initial activation cast, weight casts inserted before matmul, forward and backward BMM input/output casts, gradient-add casts, and the final \texttt{x.float()} backward conversion. The scale of the opportunity is set by the entry cast alone, which the FP32 promotion inflates to $241$\,ms in the forward pass, so shifting the dtype at which operands cross global memory acts on the dominant term rather than a marginal one. The concrete before-and-after reduction across the combined stage, together with the end-to-end module speedups measured after fusion, is reported in Section~\ref{subsec:eval_operator_chain_fusion}. Correctness is argued operator by operator: reductions and accumulations that require FP32 retain FP32 internally, while tensor-wide BF16 values remain BF16 at the stage boundaries, matching the mixed-precision contract used by the surrounding training system.

Taken together, this fusion and rewrite turns the long-tail diagnosis into an implementation discipline. Across the four targets the recurring move is the same: recover the model-level operation from its frontend expansion, hold its mathematical contract fixed, and choose the granularity and substrate at which it meets the NPU, whether the starting point was an eager autograd chain or a generic Triton kernel. What changes is not the arithmetic but where it runs and how often its operands cross global memory. The module-level effect of these fusions is reported in Section~\ref{subsec:eval_operator_chain_fusion}, and their aggregate contribution to end-to-end training throughput is reported with the other operator-level optimizations in Section~\ref{sec:train_infra}.

\section{OR-Oriented Post-Training Workflow}
\label{sec:or_post_training_workflow}

Building on the system infrastructure established in Section \ref{sec:train_infra}, we further
explore how to adapt the DeepSeek-V4~\citep{deepseekai2026deepseekv4highlyefficientmilliontoken} family to the complex domain of
Operations Research (OR). We use DeepSeek-V4-Flash as a
validation model for the OR-oriented post-training pipeline. DeepSeek-V4-Flash retains the core structured reasoning
capabilities of the V4 family, and the verified pipelines can
directly serve as a blueprint for larger models such as DeepSeek-V4-Pro.

Operations Research (OR) modeling provides a demanding setting for evaluating
domain-oriented post-training. A model must translate natural-language
requirements into variables, objectives, constraints, data interfaces, and
solver-executable programs, while preserving the mathematical structure of the
underlying optimization problem. Recent studies on LLMs for optimization have
shown that this process involves more than answer generation: it requires
semantic understanding, symbolic formulation, solver-aware implementation, and
structure-sensitive verification~\citep{xiao2025optimization_llm_survey,
wang2025llm_or_survey,orqa,orgeval}. These properties make OR a suitable
domain for examining whether post-training can improve both executable behavior
and structural modeling competence.

As a lighter-weight member of the DeepSeek-V4 family~\citep{deepseekai2026deepseekv4highlyefficientmilliontoken}, DeepSeek-V4-Flash 
enables rapid iteration over data construction, continued pre-training,
supervised fine-tuning, checkpoint selection, and benchmark evaluation,
while retaining the modeling and reasoning characteristics needed for a
realistic vertical-domain study.
The original DeepSeek-V4-Flash checkpoint already exhibits non-trivial OR
modeling ability, but its behavior varies across Pass@1, Pass@16, and 5-shot.
This variation provides a
useful starting point for analyzing which capabilities are already present,
which capabilities are only activated by context, and which failures require
parameter-level adaptation.

This section presents an OR-oriented post-training validation framework that
connects original-checkpoint diagnosis, CPT--SFT design, and stage-wise
evaluation. We first analyze the original checkpoint on OR benchmarks and
summarize representative failure cases, focusing on protocol compliance,
solver-facing implementation, mathematical formulation, and structural
equivalence. We then connect these observations to the training design: CPT is
used to broaden OR-domain knowledge and modeling priors, while SFT is used to
align the checkpoint with verified Gurobi code, task-specific output contracts,
and canonical formulation styles. Finally, we validate the workflow through
stage-wise experiments, including CPT training stability, SFT data scaling and
cleaning, and matched CPT-to-SFT transfer.



\subsection{Capability Boundary Analysis of the Original Checkpoint}
\label{subsec:original_checkpoint_diagnosis}

Before introducing the post-training recipe, we diagnose the original
DeepSeek-V4-Flash checkpoint on the same OR evaluation suite used throughout
this work. The base checkpoint is not a blank slate: it already writes many
valid Gurobi programs and solves a substantial fraction of natural-language
optimization problems. The practical question is more specific: which parts of
the OR modeling pipeline are already latent in the model, which parts can be
activated by prompting or sampling, and which parts require parameter updates
through CPT or SFT.

\subsubsection{Benchmark-Level Signal}

Table~\ref{tab:original_checkpoint_diagnosis} summarizes the original
checkpoint under Pass@1, Pass@16, and 5-shot. For NL4OPT, OptiBench, and
B4O-Feasible, a sample is
counted as correct when the generated solver program returns the reference objective
value. For B4O-ORGEval, correctness requires structural equivalence to the
reference optimization model, measured by the ORGEval WL reward. This
distinction is central to our diagnosis: executable code and even plausible
objective values do not guarantee the correct variable family, constraint
family, or constraint--variable graph.

\begin{table}[t]
\centering
\small
\caption{Diagnostic performance of the original DeepSeek-V4-Flash checkpoint
before our CPT and SFT stages. All benchmark entries are accuracies under the
corresponding evaluation setting; the final column reports the total number of
failed cases scanned under each setting.}
\label{tab:original_checkpoint_diagnosis}
\begin{tabular}{lccccc}
\toprule
Setting & NL4OPT & OptiBench & B4O-Feasible & B4O-ORGEval & Failures \\
\midrule
Pass@1  & 84.08 & 63.33 & 60.47 & 34.26 & 661 \\
Pass@16 & 93.77 & 78.33 & 80.52 & 60.66 & 370 \\
5-shot  & 88.24 & 66.00 & 78.20 & 71.57 & 425 \\
\bottomrule
\end{tabular}
\end{table}

The results reveal three diagnostic signals. First, B4O exposes a clear
structural bottleneck. Under Pass@1, B4O-Feasible reaches 60.47\%, while B4O
ORGEval reaches only 34.26\%. Meanwhile, the code/build pass rate for B4O
ORGEval is 79.19\%, indicating that many generated programs are executable but
do not preserve the intended mathematical structure. This gap shows that
solver executability and structural equivalence are only weakly aligned in the
original checkpoint.

Second, Pass@16 improves NL4OPT, OptiBench, and B4O-Feasible compared with
Pass@1, showing that additional sampling can recover more valid solver-value
solutions. However, Pass@16 still remains below 5-shot prompting on B4O
ORGEval, suggesting that sampling alone does not reliably select canonical
optimization formulations.

Third, the 5-shot performance of B4O-ORGEval increases from 34.26\% to 71.57\%, which demonstrates that the contextual examples can activate solver-facing
protocols, Gurobi implementation patterns, and formulation templates that are
already partially present in the checkpoint. These observations motivate the
following CPT--SFT design: CPT strengthens OR-domain modeling priors and
structural formulation knowledge, while SFT consolidates prompt-activated
behaviors into stable one-pass generation.

\subsubsection{Error Spectrum Across Settings}

\begin{figure}[t]
\centering
\includegraphics[width=0.92\linewidth]{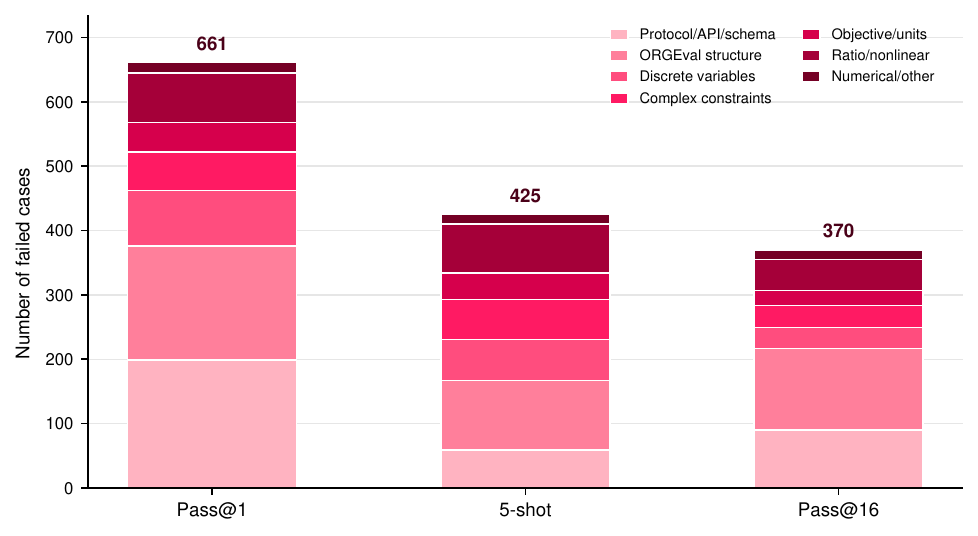}
\caption{Stacked error-family counts of the original checkpoint under Pass@1,
5-shot, and Pass@16. The bar heights show total failed cases, while the colored
segments show how the failure composition changes. 5-shot sharply reduces
protocol/API/schema failures; Pass@16 reduces total failures but leaves a large
structural ORGEval segment.}
\label{fig:base_error_family_bars}
\end{figure}

To make the diagnosis actionable, we conducted a comprehensive error analysis on the
original checkpoint across the three inference settings and four benchmark
cells, covering 1456 failed cases in total. Table~\ref{tab:original_checkpoint_error_families}
and Figure~\ref{fig:base_error_family_bars} group these failures into broad
error families to show how the distribution changes from Pass@1 to Pass@16 and
5-shot. Overall, structural equivalence errors form the largest family: WL graph
mismatch, variable-count mismatch, and constraint-count mismatch together
account for a large fraction of B4O-ORGEval failures. Under Pass@1, the most
visible failure surface is protocol, API, and schema compliance, including
markdown fences in code-only tasks, model extraction failures, and malformed
Python. The remaining high-frequency errors reflect missing OR modeling priors:
discrete entities modeled as continuous variables, incomplete constraint
systems, objective-scale mistakes, ratio or efficiency modeling errors, and
nonlinear forms that are not compatible with the solver.

\begin{table}[t]
\centering
\small
\caption{Error-family distribution of the original DeepSeek-V4-Flash
checkpoint. Entries are shares among failed cases in each setting
(Overall: 1,456; Pass@1: 661; Pass@16: 370; 5-shot: 425).}
\label{tab:original_checkpoint_error_families}
\begin{tabular}{lcccc}
\toprule
Error family & Overall & Pass@1 & Pass@16 & 5-shot \\
\midrule
Structural equivalence & 28.3 & 26.8 & 34.3 & 25.4 \\
Protocol / API / schema & 23.9 & 30.1 & 24.3 & 13.9 \\
Ratio / nonlinear modeling & 13.8 & 11.6 & 13.0 & 17.9 \\
Discrete-variable semantics & 12.5 & 13.0 & 8.6 & 15.1 \\
Complex constraints / objectives & 10.8 & 9.1 & 9.5 & 14.6 \\
Objective / units / extraction & 7.6 & 7.0 & 6.2 & 9.6 \\
Numerical tolerance / rounding & 3.2 & 2.4 & 4.1 & 3.5 \\
\bottomrule
\end{tabular}
\end{table}

\subsubsection{Inference-Setting Shift}

The distribution shift across inference settings separates promptable
behavioral failures from deeper modeling failures. Under Pass@1, the checkpoint
produces 661 failures, with protocol/API/schema errors forming the largest
family: 199 cases, or 30.1\%. This is the raw evaluator-facing failure
surface of the original checkpoint. Moving to 5-shot reduces failures to 425
and cuts protocol/API/schema errors to 59 cases, showing that many code-only
contracts, Gurobi idioms, and schema-handling habits are already latent and can
be activated by demonstrations. Moving to Pass@16 further reduces total
failures to 370 and improves value-based benchmarks, but structural ORGEval
errors become the largest remaining family, 127 cases or 34.3\%. Thus,
additional sampling can retrieve more numerically correct solutions, but it
does not reliably choose the canonical variable--constraint graph.

The same observations can be read as a training-oriented failure taxonomy.
Below, the failure types are ordered by their overall share among the 1,456
scanned failed cases. This paragraph-level view separates failures that mainly
block the evaluation interface from failures that expose missing OR modeling
priors, while keeping the representative cases and training implications tied
to concrete samples.

\paragraph{Canonical structure mismatch under ORGEval (28.3\%).}
This is the largest failure type in the scan and the most important structural
problem. In the pharmaceutical blending cases \texttt{BENCH4OPT\_0/1}, the
generated LP can be executable, but it fails variable-count or
constraint-count checks because the model uses a non-reference variable or
constraint family. \texttt{BENCH4OPT\_12} further shows that even when variable
and constraint counts match, the WL graph can still be wrong because the
constraint--variable connections differ. These failures are important because
solver success or a plausible objective value is not enough for ORGEval: the
model must produce the canonical LP graph. CPT is needed for domain templates
such as blending, scheduling, and flow; SFT is needed to align these templates
with benchmark-facing variable and constraint families.

\paragraph{Output protocol, syntax, API, and schema failures (23.9\%).}
This is the most fatal evaluation-interface problem. In \texttt{BENCH4OPT\_4},
a capital-budgeting problem, the Pass@1 response adds a markdown code fence in
a code-only target, causing a syntax failure before model construction. In
\texttt{BENCH4OPT\_176}, a nurse-scheduling problem, a string shift label is
used to index a list of requirements, triggering a schema/indexing runtime
error. These cases show that the checkpoint may have usable OR reasoning
internally but fail to expose it through the required program interface.
Non-executable code is scored as zero regardless of whether the mathematical
idea is correct, so this family can severely mask the checkpoint's real OR
capability. It primarily motivates SFT: code-only contracts, verified Gurobi
APIs, and schema-safe templates must become stable Pass@1 behavior.

\paragraph{Discrete variable semantics (12.5\%).}
NL4OPT 15 is a staffing problem that asks for numbers of senior and young
workers. The model encodes the worker counts as \texttt{GRB.CONTINUOUS} and
returns the LP relaxation value 28125.0 instead of the integer optimum
28250.0. The same relaxation persists under 5-shot and Pass@16, showing that
prompting and sampling do not reliably recover the missing integrality prior.
This is a core OR semantics problem: the model understands many constraints,
but fails to map countable business entities to integer or binary variables.
CPT should strengthen countable-entity priors, while SFT should provide
explicit integer/binary Gurobi demonstrations for workers, trips, batches,
vehicles, assignments, and selections.

\paragraph{Loss/yield-driven conservation constraints (10.8\%).}
\texttt{BENCH4OPT\_2} is a water-distribution network-flow problem with
capacity, cost, evaporation rate, and node demand. The Pass@1 model solves an
LP but mixes evaporation into both cost and conservation in a way that returns
265.58 instead of 531.15. 5-shot and Pass@16 recover the correct value,
showing that the template is promptable but not stable. This represents
complex constraint semantics rather than mere code generation. Loss, yield,
shrinkage, and evaporation determine where conservation equations change and
whether costs or delivered quantities change. CPT should teach these domain
patterns, while SFT should distill promptable templates into stable
data-separated code.

\paragraph{Ratio, unit, and total-quantity coupling (7.4\%).}
OptiBench 6 is a logistics investment problem where route-optimization
investment reduces per-truck delivery cost and total investment must not exceed
total savings. Pass@1 and Pass@16 recognize the bilinear relation but mishandle
investment units and total savings; 5-shot removes the truck-count multiplier
from the savings constraint. This failure corrupts the economic meaning of an
optimization model while still producing plausible algebra. Ratio, ROI,
average-cost, percentage, and efficiency expressions require the model to
distinguish per-unit quantities, totals, denominators, and scaling units. CPT
should broaden this semantic prior, while SFT can teach verified formulations.

\paragraph{Solver-compatible nonlinear reformulation (6.4\%).}
OptiBench 13 asks for the closest feasible rescue point on a parabolic
boundary. The Pass@1 solution writes a fourth-order distance objective directly
into Gurobi and triggers an ``objective must be linear or quadratic'' error.
5-shot and Pass@16 avoid the build error but still return the wrong distance,
which means the missing piece is not just API syntax but nonlinear
reformulation knowledge. This blocks a whole class of mathematically valid OR
tasks. The model must know when to introduce auxiliary variables, when to use
quadratic/nonconvex/general constraints, and when analytic simplification is
required. CPT is the main lever; SFT hardens the verified Gurobi patterns.

This taxonomy explains why the setting-wise distribution motivates both
training stages. The Pass@1 distribution is dominated by protocol/API/schema
and structural failures, so SFT is needed to make code-only generation, Gurobi
API usage, objective extraction, and JSON-style data access stable without
in-context demonstrations. The 5-shot distribution shows which failures are
distillable: markdown fences, several API mistakes, and some common
formulation templates improve sharply once examples are present. This directly
motivates verified SFT and few-shot distillation. However, the residual
5-shot errors concentrate in integer semantics, complex constraints,
ratio/nonlinear modeling, and ORGEval structure, which are not just output
habits. They require CPT to broaden the model's OR formulation priors before
SFT can reliably elicit them.

The Pass@16 distribution adds a different lesson. Because total failures fall
from 425 to 370 and value-based accuracies improve, the original sampling
distribution already contains many recoverable correct programs; this supports
verifier-guided SFT and rejection-sampled distillation from Pass@\(N\) outputs.
But Pass@16 leaves more structural ORGEval failures than 5-shot, both in
absolute count (127 vs. 108 grouped structural errors) and in share
(34.3\% vs. 25.4\%). Sampling therefore improves search over solutions without
enforcing the canonical LP graph. This is the key bridge to CPT plus SFT: CPT
shifts the model toward OR-native structural priors, while SFT turns those
priors into benchmark-facing, executable, and aligned outputs.

\subsubsection{Two-Stage Training Adaptation: Why Both CPT and SFT}

In large language model pipelines, continual pre-training serves as a knowledge-infusion stage---adapting the model to a domain through self-supervised learning on unlabeled corpora~\citep{gururangan2020dontstop, ke2025demystifying, siriwardhana2024domain}---while instruction tuning acts as a behavior-shaping stage~\citep{ouyang2022training, wei2022finetuned, sanh2021multitask} that aligns the model with human intent via supervised learning on instruction-response pairs. The core distinction is that continual pre-training determines what the model knows, and instruction tuning controls how it responds. Their relationship is one of foundation and activation: continual pre-training first builds a domain-informed knowledge base, and subsequent instruction tuning elicits it as coherent, instruction-following behavior~\citep{yang2025main, jindal2024balancing, jiang2025mix, zhang2025adept}. Only this cascaded synergy yields models that are both domain-proficient and reliably usable.

The setting-wise diagnosis gives a clean division of labor. CPT targets
knowledge acquisition: OR terminology, standard formulation patterns,
solver-aware nonlinear reasoning, fractional and efficiency modeling,
data-structure grounding, and structural modeling priors. These are the
failure modes that persist even when examples or multiple samples are added:
integer-domain semantics, nonlinear reformulation, ratio modeling, missing
constraint families, and canonical LP graph structure.

SFT targets solver-oriented behavioral alignment. It should teach the model
to obey code-only and LP-writing contracts, avoid markdown fences, use
verified Gurobi APIs, map JSON/list/dict data safely into index sets, extract
objective values with the right units, and reproduce benchmark-facing
canonical formulations. The 5-shot improvements show that many of these
habits are already promptable; SFT converts that prompt dependence into
stable Pass@1 behavior. The Pass@16 improvements show that correct solutions
often exist in the sampling distribution; verifier-guided distillation can
convert those retrieved samples into parameterized competence. The remaining
structural and modeling failures explain why this SFT stage should be built
on top of OR-oriented CPT rather than replacing it.

\subsection{OR-Oriented Continued Pre-Training}
\label{subsec:or_cpt}

The original-checkpoint analysis shows that DeepSeek-V4-Flash already contains
a useful OR modeling core, especially on standard LP and MILP templates, but
its remaining errors concentrate on variable-domain semantics,
solver-compatible nonlinear reformulation, ratio- and efficiency-based
modeling, complex constraint construction, and structural equivalence. These
failure modes indicate that the next stage should not only teach the model to
follow benchmark protocols, but also strengthen its prior over OR-domain
concepts, mathematical-programming structures, and solver-aware modeling
patterns. We therefore use CPT as the knowledge-acquisition stage in the
OR-oriented post-training workflow, following the domain-adaptive
pre-training paradigm for adapting language models to specialized
distributions~\citep{gururangan2020dontstop,chen2026continual_learning_llm}.

In this stage, the model is further trained on a domain-dominant mixture built
around solver-verified OR documents. The design is guided by the capability
gaps observed in the original checkpoint: integer and binary variable
semantics motivate broader exposure to countable-entity formulations;
nonlinear and ratio-based failures motivate solver-aware reformulation data;
and ORGEval structural mismatches motivate data that preserves canonical
variable families, constraint families, and constraint--variable connections.
Following this principle, our CPT workflow centers on an OR-CPT data engine
that converts parameterized optimization generators into traceable,
business-oriented, and executable modeling documents. The resulting CPT stage
serves as a bridge between data construction and SFT: it expands the model's
OR-domain modeling distribution before SFT aligns the checkpoint with verified
Gurobi code, strict output contracts, and benchmark-specific formulation
styles~\citep{jiang2025mix}.

\subsubsection{OR-CPT Data Construction and Domain-Dominant Mixture}
\label{subsubsec:cpt_data_construction}

\begin{figure}[t]
  \centering
  \includegraphics[width=1.0\textwidth]{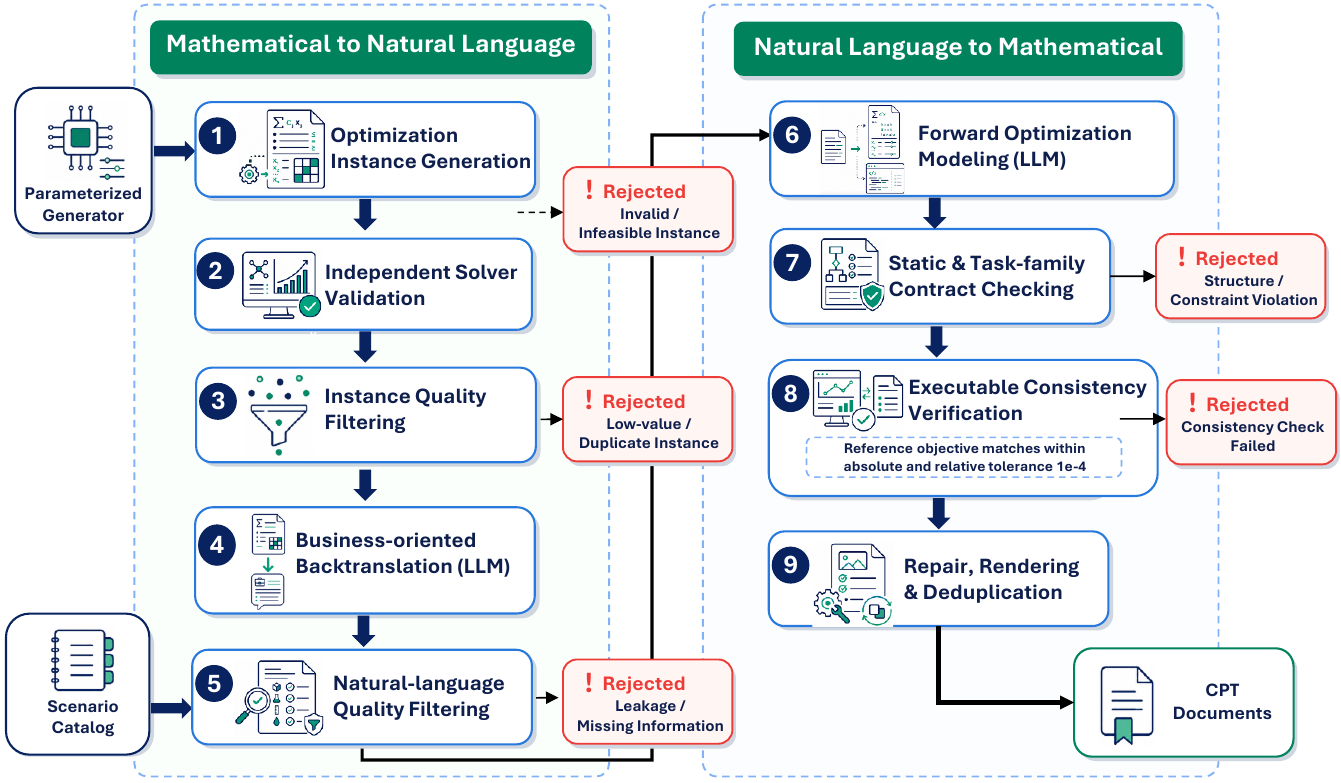}
  \caption{Pipeline of OR-CPT data construction. Parameterized optimization generators produce structured instances, which are verified with Gurobi, rendered into business-oriented problem statements, reconstructed into executable formulations, and filtered by contract checks, solver execution, and objective-value agreement before export.}
  \label{fig:or_cpt_pipeline}
\end{figure}

The core of our CPT data construction is the OR-CPT Data Engine, a
solver-verified bidirectional synthesis workflow that converts structured
optimization instances into natural-language modeling documents, as illustrated
in Figure~\ref{fig:or_cpt_pipeline}. The workflow
targets the capability gaps observed in the original-checkpoint analysis,
including variable-domain ambiguity, weak parameter grounding, incomplete
constraint construction, nonlinear or ratio-based modeling failures, and
structural mismatch under ORGEval-style evaluation. Each accepted CPT document
is traceable to a generated optimization instance, an independently verified
solver reference, and a checked forward modeling result.

The workflow starts from parameterized optimization instance generators. Each
generator specifies the decision variables, objective, constraints, numerical
parameters, and task-specific structure of an optimization family. The current
generator registry covers assignment, scheduling, facility location, network
flow, production planning, lot sizing, routing, covering, transportation,
portfolio optimization, and revenue management. For each instance, the engine records structured metadata, including task family, generator identity, concept tags, difficulty label, generation parameters, and canonical mathematical signature. This metadata is used for corpus sampling, auditing, and concept-level analysis, but is not exposed as model-facing CPT text. We also maintain separation between the data-generation assets and the ORGEval evaluation assets: no ORGEval reference formulation, canonical LP graph, or evaluator-side template is reused when rendering CPT documents. The intended overlap is at the level of broad OR problem families and modeling concepts, rather than at the level of shared canonical templates or instance provenance.

Each generated instance is independently executed with Gurobi to obtain a
reference solver status, objective value, variable assignment, and constraint
diagnostics. The engine then applies instance-level quality filters to remove
unstable or low-value samples, including infeasible instances, degenerate
all-zero solutions, duplicated parameter configurations, duplicated solution
signatures, and weak optimization trade-offs. This solver-grounded filtering
step provides the mathematical anchor for the synthetic corpus and improves the
quality of the subsequent natural-language modeling documents.

After solver validation, the structured instance is converted into a
business-oriented natural-language problem. A scenario catalog controls the
industry background, organization type, planning horizon, entity naming style,
units, and narrative perspective. Deterministic leakage checks prevent the
problem statement from exposing the optimal objective value, solver status,
implementation code, or other answer-related artifacts. This step turns a
mathematical instance into a realistic modeling prompt while preserving the
underlying optimization structure.

The second direction of the workflow performs forward modeling from the
generated natural-language problem. Given only the problem statement, a model
produces decision variables, objective functions, constraints, modeling
explanations, and executable Gurobi code. The generated answer is checked by
static rules and task-family contracts. For example, facility-location samples
are expected to connect facility-opening variables with assignment variables,
network-flow samples should contain flow-conservation constraints, and
scheduling samples should preserve assignment, coverage, and capacity
structures. The reconstructed Gurobi program is then executed independently
and compared with the solver reference from the original structured instance.
A sample is accepted only when the program reaches a valid solver status and
its objective value matches the reference solution within the verification
tolerance. Recoverable failures are repaired with structured solver feedback,
and accepted documents are rendered and deduplicated before export.

This bidirectional design gives the synthetic CPT corpus three useful
properties. It is scalable because new data can be produced from parameterized
OR generators across many task families. It is verifiable because each
accepted document is tied to solver execution and reference-objective
agreement. It is structurally informative because the workflow preserves the
connection among natural-language requirements, variable families, constraint
families, solver code, and objective values. These properties make the OR-CPT
Data Engine the main source of solver-grounded modeling signal in our CPT
stage.

The complete CPT corpus combines these solver-verified synthetic documents
with collected OR-domain resources. The collected portion covers optimization
modeling, mathematical reasoning, and solver-oriented problem solving,
including OptMATH~\citep{optmath2025}, ReSocratic and
OptiBench~\citep{yang2025optibench}, Text2Opt-Bench~\citep{gao2026text2opt},
OR-Instruct~\citep{orlm}, NLP4LP~\citep{optimus2024}, ORQA-derived
data~\citep{orqa,synthetic_orqa_2025}, and MAMO~\citep{huang2024mamo}. We
also include OR research papers, official Gurobi modeling examples, and
textbook-style materials to increase coverage of formal modeling conventions,
solver usage, and domain terminology
~\citep{frontieror2026,gurobi_modeling_examples,lownes2024or}. All collected
resources are converted into a unified CPT document format and processed
through source-aware cleaning, formatting, and deduplication. At the current
stage, the OR data pool contains approximately 100K samples, which serve as
the corpus inventory for CPT sampling.

We adopt a domain-dominant mixture strategy in which OR documents provide the
main adaptation signal. Mathematical, code, and general English corpora are
included as supporting components. The mathematical component is drawn from
selected AutoMathText-V2 subsets~\citep{automathtext_v2}, following the
AutoMathText data-selection methodology~\citep{automathtext}, and supports
symbolic formulation, derivation, and multi-step reasoning. The code component
is drawn from NVIDIA Nemotron pretraining data releases, including
scientific-coding data associated with the Nemotron Nano series and
code-concept data associated with Nemotron 3 Super
~\citep{nvidia2025nvidianemotronnano2,nvidia2026nemotron3superopen}, and
supports procedural reasoning, solver-oriented implementation, and robust code
generation. The general English component uses Ultra-FineWeb-L3, derived from
Ultra-FineWeb under the UltraData tiered data management framework
~\citep{ultra_fineweb,ultra_data_tiered}, to maintain broad language
comprehension and generation quality. In this mixture, OR data strengthens
formulation semantics and structural modeling priors, mathematical data
supports reformulation and symbolic reasoning, code data supports executable
solver implementation, and general English data stabilizes broad language
capability. The exact sampling allocation is treated as a recipe-level
parameter and reported with the training configuration.

\subsubsection{Training Execution in the Ascend Ecosystem}
\label{subsubsec:cpt_training_execution}

We execute OR-oriented CPT within the Ascend software ecosystem, using
MindSpeed and TorchTitan-NPU as two implementation routes. Both routes follow
the same autoregressive training objective, OR-oriented data interface,
monitoring protocol, and checkpoint-evaluation workflow. This design keeps the
CPT recipe consistent across different Ascend training abstractions, so that
data construction, optimization behavior, and downstream validation remain
comparable.

The execution setup is configured around the main requirements of
DeepSeek-V4-Flash CPT. Long-context training is supported through context
parallelism, which is important for OR documents that contain business
descriptions, mathematical formulations, solver traces, and multi-step
modeling explanations. Sparse MoE execution is supported through expert
parallelism, token dispatch, and grouped expert computation. The training stack
also supports model-specific attention and MoE operator paths, MTP training,
activation recomputation, and memory-aware optimizer execution. These
capabilities provide the execution basis for full-parameter CPT while keeping
the model architecture and language-modeling objective unchanged.

The broader Ascend system design, including cluster organization, collective
communication, operator optimization, runtime monitoring, and fault recovery,
is reported in Section~\ref{sec:train_infra}. Here, the training execution
serves as the substrate for the OR-oriented CPT recipe. During CPT, we monitor
loss dynamics, learning-rate progression, gradient norms, numerical stability,
throughput, memory behavior, and iteration stability. These signals are used to
identify stable candidate checkpoints for intermediate evaluation and
subsequent SFT. Recipe-level settings, including sequence length, batch size,
learning-rate schedule, parallelism degrees, optimizer configuration, and
memory-saving mechanisms, are summarized in
Appendix~\ref{app:post_training_recipes}.

\subsubsection{Checkpoint Selection and CPT-to-SFT Transfer Validation}
\label{subsubsec:cpt_checkpoint_selection}

CPT checkpoints are selected according to their value as SFT initializations.
We evaluate intermediate checkpoints with three complementary signals:
OR-domain likelihood, general-capability retention, and downstream OR transfer.
The OR likelihood signal is measured on a held-out OR validation set that is
disjoint from the CPT training mixture. It provides an early indication of
whether the checkpoint better models OR terminology, formulation patterns, and
solver-oriented documents. General validation loss and representative
general-capability benchmarks are tracked in parallel to monitor whether the
checkpoint preserves the mathematical, coding, and language abilities needed
for later instruction tuning.

We further evaluate selected checkpoints on OR downstream benchmarks. These
benchmarks cover different aspects of optimization modeling: NL4OPT emphasizes
natural-language-to-formulation ability, OptiBench evaluates broader
optimization problem understanding, B4O-Feasible measures solver-facing
feasibility behavior, and B4O-ORGEval tests structural modeling quality through
a stricter formulation-level evaluation. This benchmark set allows us to check
whether improvements in OR-domain likelihood translate into measurable gains in
modeling, implementation, and structural-equivalence behavior.

Since CPT is designed as a preparation stage for SFT, its most important
evaluation is matched transfer validation. We compare two controlled training
routes:
\begin{align}
\text{Route A:}\quad & \text{Original Checkpoint} \rightarrow \text{SFT}, \\
\text{Route B:}\quad & \text{Original Checkpoint} \rightarrow \text{CPT} \rightarrow \text{SFT}.
\end{align}
The SFT dataset, optimizer configuration, update steps, and evaluation protocol
are kept the same between the two routes. As a result, the difference
between the trained models reflects the contribution of CPT as an initialization
stage. For a downstream benchmark \(b\), we define the transfer gain as:
\begin{equation}
\Delta_{\mathrm{transfer}}^{(b)}
=
S_{\mathrm{CPT+SFT}}^{(b)}
-
S_{\mathrm{SFT}}^{(b)} ,
\end{equation}
where \(S_{\mathrm{CPT+SFT}}^{(b)}\) is the score of the model trained through
Route B and \(S_{\mathrm{SFT}}^{(b)}\) is the score of the matched SFT-only
route.

The final checkpoint is selected with a multi-objective criterion. A preferred
checkpoint should reduce OR validation loss, maintain stable optimization
signals during training, preserve general capabilities within the expected
range, and produce positive CPT-to-SFT transfer on the primary OR benchmarks.
This selection procedure connects the CPT stage directly to the later SFT and
experimental analysis: CPT is not only evaluated as a language-modeling
continuation, but as a domain-adaptive initialization whose value is determined
by downstream OR modeling gains under matched training conditions.

\subsection{Data-Efficient Supervised Fine-Tuning for Operations Research}
\label{sec:sft}

We conduct supervised fine-tuning for DeepSeek-V4-Flash on operations research modeling tasks. The goal is to adapt the model from general instruction following to solver-oriented mathematical programming, with emphasis on benchmarks such as NL4OPT, OptiBench, and Bench4Opt. Unlike general code-generation tasks, OR modeling evaluation does not only check executability. It also evaluates natural-language semantic understanding, variable-type selection, objective and constraint construction, Gurobi-style API usage, external data loading, answer-format contracts, and structural equivalence of mathematical formulations.

Since large-scale SFT data in the OR domain is scarce, heterogeneous, and uncertain in quality, we adopt a DeepSeek-V4-Flash cyclic self-distillation method inspired by iterative data-construction frameworks such as ReSocratic-style synthesis~\citep{yang2025optibench,zhang2025scoder}. More broadly, self-distillation has been shown to be effective for bootstrapping high-quality synthetic data~\citep{lee2026self}, especially when combined with structured validation for optimization modeling tasks~\citep{wu2025training}. We construct three training-data scales, namely 3K, 10K, and 50K. Based on the comparative experiments reported in Section~\ref{subsec:sft_training_cleaning}, we choose the 10K-scale data as the basis for further cleaning and enhancement. The final SFT design therefore focuses not only on scaling synthetic data, but also on improving the correctness, contract awareness, and learnability of the distilled samples.

\subsubsection{Solver-Oriented SFT Objective}
\label{subsec:sft_objective}

The SFT stage is designed to train the model to form stable optimization-modeling habits. The model should identify sets, parameters, variables, objectives, and constraints; correctly read structured data files; use Gurobi-style Python APIs to build and solve models; and, when required, write LP files without solving them. The target response is therefore not a simple numerical answer, but a structured modeling trajectory that connects natural-language understanding, mathematical formulation, and solver-facing implementation. This follows recent findings that supervised fine-tuning is an important foundation for equipping LLMs with optimization reasoning and modeling capabilities~\citep{orr1,xiao2026deepor}.

Given an instruction prompt $x$ and a target response $y=(y_1,\ldots,y_T)$, the SFT stage optimizes the standard autoregressive cross-entropy loss over the response tokens:
\begin{equation}
\mathcal{L}_{\mathrm{SFT}}
=
-\frac{1}{T}
\sum_{t=1}^{T}
\log p_{\theta}(y_t \mid x, y_{<t}).
\end{equation}
The loss on prompt, system, and formatting tokens is masked, so that gradient updates are applied only to the target response. This encourages the model to learn the desired reasoning trajectory, mathematical formulation, and solver implementation, while avoiding unnecessary optimization on fixed prompt templates.

\subsubsection{Self-Distilled OR Modeling Data Flywheel}
\label{subsec:sft_dataset_construction}
\label{subsec:sft_ir_flywheel}

We construct a self-distillation data-production pipeline for operations research and mathematical programming tasks. The input is a set of seed samples, including original problems, reference answers, reference code, data files, or LP files from high-quality modeling QA data. These seed samples are not directly exported as final training data. Instead, they serve as references for problem types, modeling structures, and target response specifications.


The seed samples are organized into three input modes, as summarized in Table~\ref{tab:sft_problem_modes}.

\begin{table}[t]
\centering
\small
\caption{Problem modes used in self-distilled OR modeling data.}
\label{tab:sft_problem_modes}
\begin{tabular}{llp{0.58\linewidth}}
\toprule
Mode & Name & Description \\
\midrule
DP & Data in Problem & Data are directly written in the problem statement. \\
DT & Data in Table & Data are provided in a structured table, such as a Markdown table. \\
\multirow{2}{*}{DPS} & \multirow{2}{*}{Data-Problem Separate} & The problem statement and data files are separated, such as \texttt{data.json}, \texttt{instance.json}, or CSV files. \\
\bottomrule
\end{tabular}
\end{table}

To make the self-distillation process controllable and auditable, we introduce three levels of intermediate representations. This layered design follows progressive synthesis strategies in optimization data generation, where intermediate representations help control problem complexity and structural diversity~\citep{yang2025optibench,wu2025training}.
\begin{itemize}
    \item \textbf{L1 Canonical IR} normalizes heterogeneous samples and records the source path, target contract, data interface, reference-file status, static code features, risk labels, and objective-function hints.
    \item \textbf{L2 Semantic IR} extracts the mathematical-programming semantics of the seed sample, including the objective function, sets, parameters, variables, constraints, data interface, and potential modeling risks.
    \item \textbf{L3 Synthetic IR} generates a new modeling-task abstraction based on the L2 seed and the historical accepted synthetic pool. It must not copy the seed problem statement, answer, code, LP file, file name, path, exact numbers, or formulation. Instead, it varies the domain, entities, objective semantics, variable families, constraint families, parameter schema, numerical range, and data shape.
\end{itemize}

To further reduce the risk of data leakage, we apply leakage-control checks throughout the distillation process. Candidate synthetic samples are compared against held-out evaluation samples at multiple levels, including surface-level problem wording and abstract modeling structure. Semantic-similarity filtering is used to reject near-duplicate problem statements, while IR-level constraints prevent the reuse of the same objective semantics, constraint patterns, parameter schema, or numerical values. These controls ensure that the accepted SFT samples differ from the test set not only in textual expression, but also in data form and modeling structure, thereby reducing the possibility that the model learns leaked benchmark instances rather than generalizable OR modeling behaviors.

After L3 generation, the pipeline renders the synthetic IR into DP, DT, or DPS problems. DP samples embed data directly into the statement, DT samples contain structured Markdown tables, and DPS samples generate external files such as \texttt{data.json}, \texttt{instance.json}, or CSV files. The answer rendering stage then produces target responses according to the task-specific output contract. For Markdown-style answers, the response contains \texttt{<think>}, \texttt{<model>}, and \texttt{<python>} tags, while the Python code uses Gurobi to build and solve the model and output the objective value. In this way, the model learns not a single code template, but a family of modeling-output protocols under different task requirements.

\begin{figure}[t]
\centering
\includegraphics[width=\linewidth]{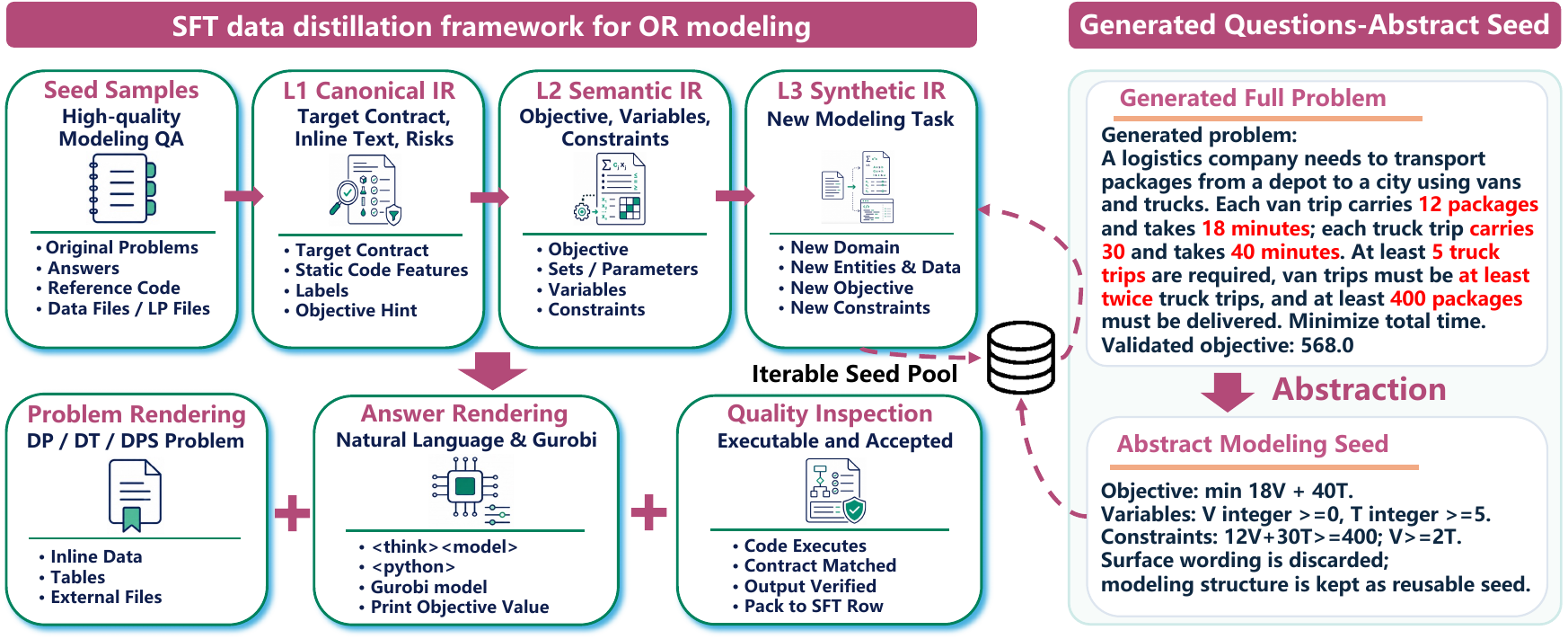}
\caption{Overview of the SFT data distillation framework for OR modeling. The pipeline converts high-quality seed samples into canonical and semantic intermediate representations, generates novel synthetic IRs, renders them into multiple problem formats, validates the corresponding answers, and feeds accepted synthetic IRs back into the next self-distillation round.}
\label{fig:sft_distillation_framework}
\end{figure}

The pipeline further adopts a flywheel-style iteration. Samples accepted into SFT in each round have their corresponding Synthetic IRs accumulated into the parent pool for the next generation round. The model therefore refers not only to the original Semantic IR, but also to previously accepted synthetic structures. This gradually reduces direct dependence on the original seeds and expands the modeling-structure space, while still using the seed data as anchors to prevent distributional drift. As illustrated in Figure~\ref{fig:sft_distillation_framework}, the overall framework connects seed samples, intermediate representations, problem rendering, answer rendering, automatic validation, and accepted-pool feedback.

\subsubsection{Contract-Aware Cleaning and CoT Enhancement}
\label{subsec:sft_cot_cleaning}

The observations from preliminary SFT runs motivate a shift from merely enlarging the self-distilled dataset to improving the learnability of already distilled samples. In the comparison among 3K, 10K, and 50K reported in Section~\ref{subsec:sft_training_cleaning}, the 10K dataset already significantly improves Bench4Opt protocol-oriented tasks, but it remains below the baseline on NL4OPT and OptiBench. Further scaling to 50K does not recover this ability. Therefore, instead of continuing to expand the same pipeline to 100K, we perform fine-grained cleaning of the existing 10K data, focusing on modeling errors and chain-of-thought expression.

The cleaning process targets concrete error types exposed in the previous stage. In optimization modeling tasks, structural errors that directly damage the training signal include incorrect variable types, wrong objective direction, wrong constraint direction, missing capacity or demand constraints, improper handling of ratios or averages, and incorrect use of the Gurobi API. For example, objects such as counts, vehicles, routes, assignments, and selections are usually modeled as integer or binary variables, whereas continuous quantities such as tons, liters, flow rates, and working hours should not be over-integerized.

We design two cleaning branches. The first branch is \textbf{clean-only}, which repairs real modeling or implementation errors while preserving the original reasoning style as much as possible. It does not actively add new explanatory content. The second branch is \textbf{Clean-CoT}, which repairs errors and adds a concise modeling checklist to the target response. This checklist makes the \texttt{<think>} block explicitly cover variable types, objective direction, key constraints, ratio or nonlinear forms, and Gurobi execution actions. The goal is not to generate long free-form reasoning, but to expose the model repeatedly to a stable modeling pattern: determine variable domains, determine the objective, check constraints, and implement the formulation in Gurobi. This design is consistent with recent findings that CoT-style SFT can transfer reasoning ability more effectively when the reasoning traces are structured and task-relevant~\citep{yu2025long,lin2025facilitating,yeo2025demystifying}.

The cleaning workflow uses four levels of gating: cleaner, reviewer, local validation, and export checking. The cleaner generates repaired candidate answers. The reviewer independently determines whether a candidate is suitable for SFT and blocks over-repair, such as incorrectly changing continuous variables into integers, deleting necessary constraints, or mishandling ratio and big-M formulations. Local validation then checks code executability, Gurobi calls, output contracts, and LP writing paths. In the full 10K run, clean-only passes reviewer filtering for 9,043 out of 10,000 samples, and 8,731 samples finally pass validation and are exported. Clean-CoT passes reviewer filtering for 9,174 out of 10,000 samples, and 8,881 samples are finally exported.

Overall, this cleaning stage makes self-distilled SFT data more contract-aware and more learnable. For tagged Markdown targets, short structured \texttt{<think>} blocks inject the most error-prone optimization-modeling checkpoints into the training distribution. For code-only and LP-structure-equivalence tasks, the output contract remains strict.

\subsubsection{Progressive SFT with General Capability Anchoring}

Our scaling experiments reveal a conflict between two capabilities that SFT must instill simultaneously: syntactic precision in solver-API usage and semantic reasoning in natural-language modeling. In a single-stage setup, gradient updates from pure code targets such as Bench4Opt and reasoning-rich targets such as NL4OPT may interfere destructively, leading to the non-monotonic performance observed when scaling from 10K to 50K samples. To reduce this conflict, we decouple these capability dimensions into a two-stage progressive training strategy. This follows the general idea of progressive domain adaptation, where broad task competence and specialized tuning are separated into different training stages~\citep{zhang2025adept}.

Stage I, \textbf{Solver-API Alignment}, trains exclusively on code-only OR targets, including LP-writing and feasibility tasks. It uses a conservative learning rate of \(5\times 10^{-7}\) to establish a stable syntactic backbone for Gurobi invocations, data-loading patterns, and output contracts. Stage II, \textbf{Full OR Modeling SFT}, continues from this checkpoint on the complete Clean-CoT 10K dataset with a learning rate of \(1\times 10^{-6}\). This allows the model to learn formulation and reasoning while preserving low-level solver syntax. By separating solver-API alignment from full modeling adaptation, the training process reduces the destructive interference observed in monolithic SFT~\citep{yildiz2024investigating}.

However, OR-specific SFT can still cause degradation in general language understanding and code syntax health, as observed in curriculum monitoring and the NL4OPT decline in Table~\ref{tab:main_results}. This drift toward domain specialization may erode the semantic representations needed for natural-language modeling and agentic diversity. To counteract this forgetting, we augment the Stage II mixture with \(10\%-15\%\) general-purpose instruction data, including open-domain QA, multi-turn dialogue, and creative writing. This general-capability anchoring maintains gradient updates from diverse linguistic tasks, helps preserve NL4OPT performance, and mitigates the parametric homogenization observed in agentic rollout experiments. As a result, the SFT checkpoint provides a more stable and explorable initialization for the downstream agentic pipeline.

\subsection{AI-Assisted Data Construction and Refinement}
\label{sec:ai}

\begin{figure}[t]
\centering
\includegraphics[width=\linewidth]{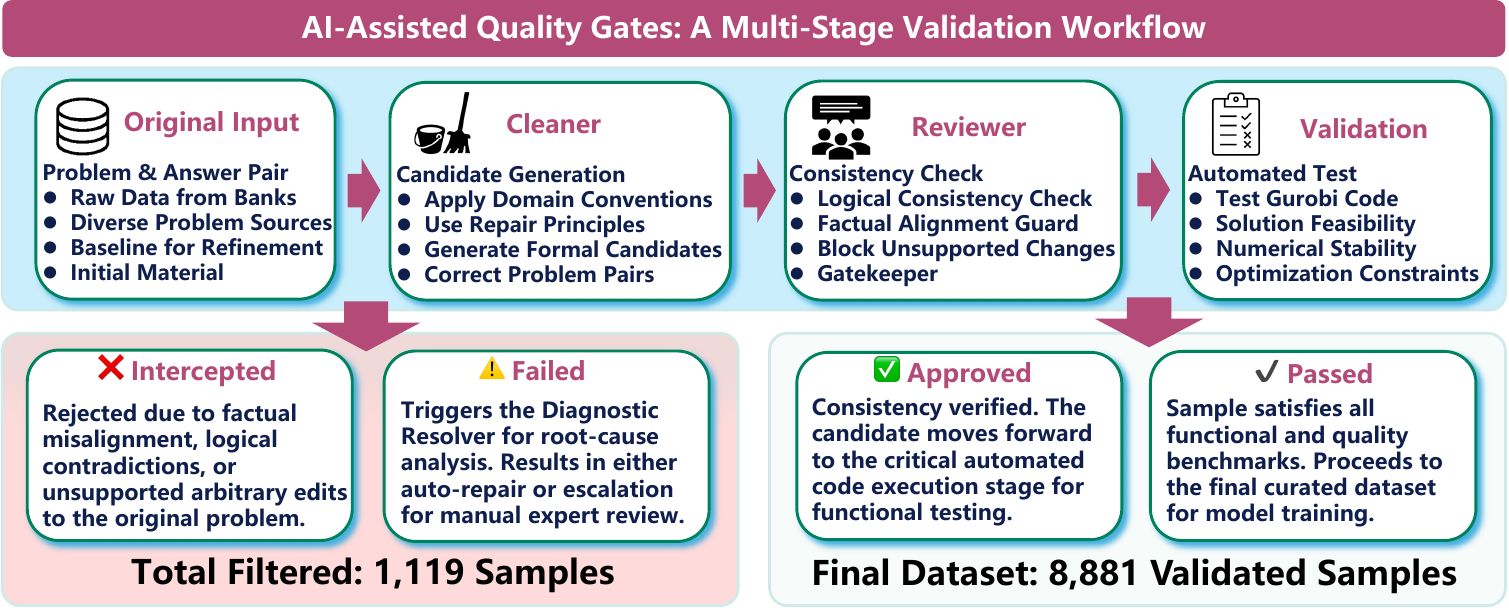}
\caption{The AI-Assisted Quality Gates workflow. This multi-stage validation process uses the Cleaner, Reviewer, and code execution Validation to ensure the quality and auditability of the final dataset.}
\label{fig:ai_workflow}
\end{figure}

To scale up the data construction process described above, we employ AI-assisted tools to automate candidate generation, repair, and validation. As illustrated in Figure~\ref{fig:ai_workflow}, the workflow combines an AI-based Cleaner and independent Reviewer with executable validation gates. Their tasks are inspired by multi-agent validation architectures~\citep{vallabhaneni2026ai}. In our workflow they operate as \textbf{prompt-and-rule-driven functions} under human-defined policies; no automatic rule invention is performed by the agents to ensure reproducibility. This section describes what these tools do and how they are orchestrated in our pipeline.

\begin{description}
    \item[Cleaner.] Given an original problem and answer, the Cleaner generates a repaired candidate. Its system prompt encodes the domain conventions, target contracts, and evidence-based repair principles described in the case studies. It is proactive in fixing clear errors (e.g., variable domain mismatches, missing constraints, obvious Gurobi API misuse) but can sometimes introduce new inconsistencies.

    \item[Reviewer.] The Reviewer evaluates each cleaned candidate independently, without access to the Cleaner's reasoning. It sees only the original problem, the original answer, and the cleaned answer. Its task is to check internal consistency and factual alignment with the prompt, and to block changes that are unsupported or introduce new errors. This adversarial check is essential to prevent over-repair.

    \item[Diagnostic Resolver.] When local validation (code execution against Gurobi) fails, the Resolver ingests the execution report, the problem statement, the current formulation, and the generated code. It produces a structured diagnosis attributing the failure to one or more categories: semantic modeling error, API usage error, data schema mismatch, or solver-incompatible expression. This diagnosis can trigger a second-stage repair or flag the sample for manual review.
\end{description}

The importance of the independent Reviewer is illustrated by an incident during development. For a transportation problem, the Cleaner accepted the sample with high confidence (\texttt{confidence = 0.95}) and reported only formatting changes. However, inspection of the cleaned output revealed that the cost from Warehouse~2 to Store~3 had been inadvertently changed from~2 to~3 in the \texttt{<model>} description, while the Python code correctly used~2. The Reviewer detected this inconsistency (\texttt{decision = needs\_repair}, \texttt{confidence = 0.95}) with the diagnosis: \texttt{issues = ["data\_inconsistency"]}. Without this gate, the conflicting sample would have entered SFT.

Across the full Clean-CoT run of 10,000 samples, the Reviewer intercepted 826 samples that the Cleaner had accepted; subsequent local validation caught an additional 293 samples. The final exported 8,881 samples are thus the intersection of three independent quality checks: Cleaner generation, Reviewer approval, and Validator execution. This multi-stage quality architecture is purely rule-driven and audit-trailed: every sample in the final SFT set carries verifiable history.

\section{Experimental Results}
\label{sec:experiment}

\label{subsec:stagewise_experimental_validation}

Our evaluation covers both halves of the full-stack effort in Section~\ref{sec:or_post_training_workflow}: the operator-level optimizations that raise device efficiency, and the post-training pipeline that turns the stabilized stack into downstream capability. We begin at the operator level and validate the two kernel-optimization strategies against the same DeepSeek-V4-Pro training workload whose bottlenecks motivated them (Section~\ref{sec:kernel_bottleneck}). Section~\ref{subsec:eval_aurakernel} evaluates AuraKernel on the heavy head of architecture-intrinsic kernels, where profiling-in-the-loop tuning reduces the per-invocation duration of the dominant sparse-attention and normalization kernels. Section~\ref{subsec:eval_operator_chain_fusion} evaluates the proactive operator-chain fusion on the fragmented, memory-bound tail, where collapsing framework-emitted eager chains into fused AscendC kernels removes redundant launches, intermediate tensors, and global-memory traffic. Because both are measured on the profiled training workload rather than on isolated microbenchmarks, the reported speedups attach directly to the device-time regimes identified in Section~\ref{sec:bottleneck}.

We then evaluate the post-training pipeline on Ascend 910C using
DeepSeek-V4-Flash as the experimental platform. We organize the experiments from infrastructure optimization to OR-specific model adaptation. We first verify that the training stack is stable on Ascend NPUs, retaining only the larger SFT stability run as the representative systems evidence. We then analyze the SFT scale and cleaning experiments, followed by the matched CPT$\rightarrow$SFT
transfer experiment. 
The final end-to-end result section is left as the summary comparison across Base, CPT, SFT, and CPT+SFT.

For clarity, the SFT labels in this section refer to two related but distinct points in the training trajectory. SFT-10K is the uncleaned 10K run used for scale analysis (Table~\ref{tab:sft_scale}) and as the pre-cleaning baseline in Table~\ref{tab:sft_cleaning}. Clean-CoT is the cleaned, CoT-enhanced checkpoint; the SFT row in the transfer and end-to-end tables (Tables~\ref{tab:cpt_sft_transfer} and~\ref{tab:main_results}) uses this matched-condition cleaned SFT checkpoint, so its 0-shot Pass@1 values align with the Clean-CoT row.

\subsection{Efficient Kernels via AuraKernel}
\label{subsec:eval_aurakernel}

\begin{figure}[h]
\centering
\includegraphics[width=\textwidth]{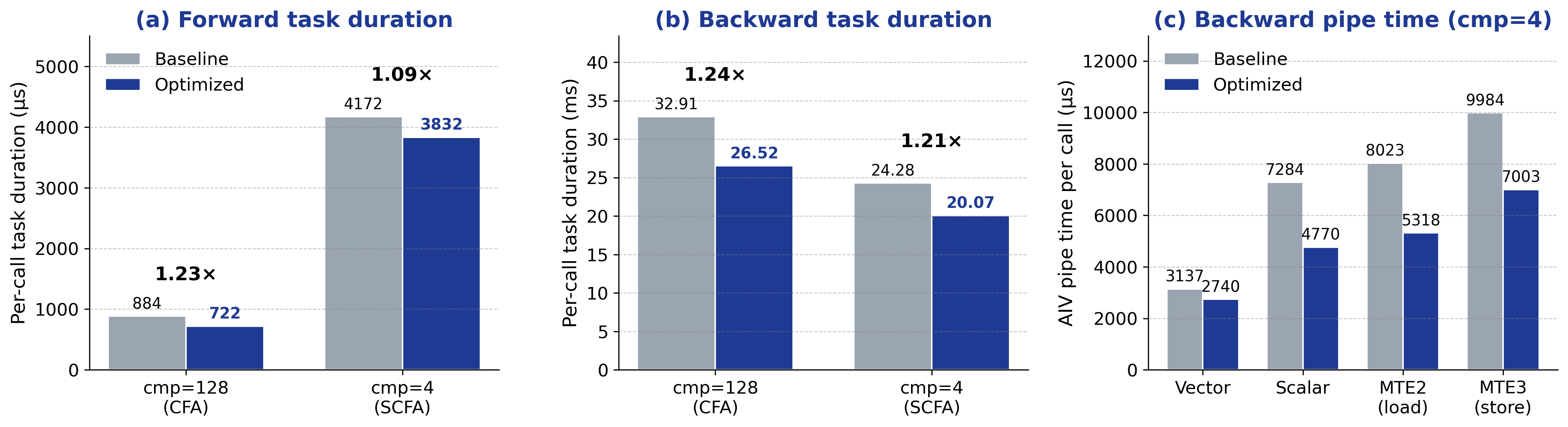}
\caption{AuraKernel optimization of the shared-key-value sparse-attention kernels on Ascend 910C. (a)~Per-call forward task duration ($1.23\times$ at compression ratio $128$, $1.09\times$ at compression ratio $4$). (b)~Per-call backward task duration of \texttt{SparseAttnSharedkvGrad} ($1.24\times$ and $1.21\times$). (c)~AIV pipeline-time decomposition of the backward kernel at compression ratio $4$: the reductions concentrate in the memory-movement pipes (MTE2 load and MTE3 store) rather than in vector compute, confirming that the gain comes from cutting global-memory traffic.}
\label{fig:sparse_series}
\end{figure}

\begin{figure}[h]
\centering
\includegraphics[width=\textwidth]{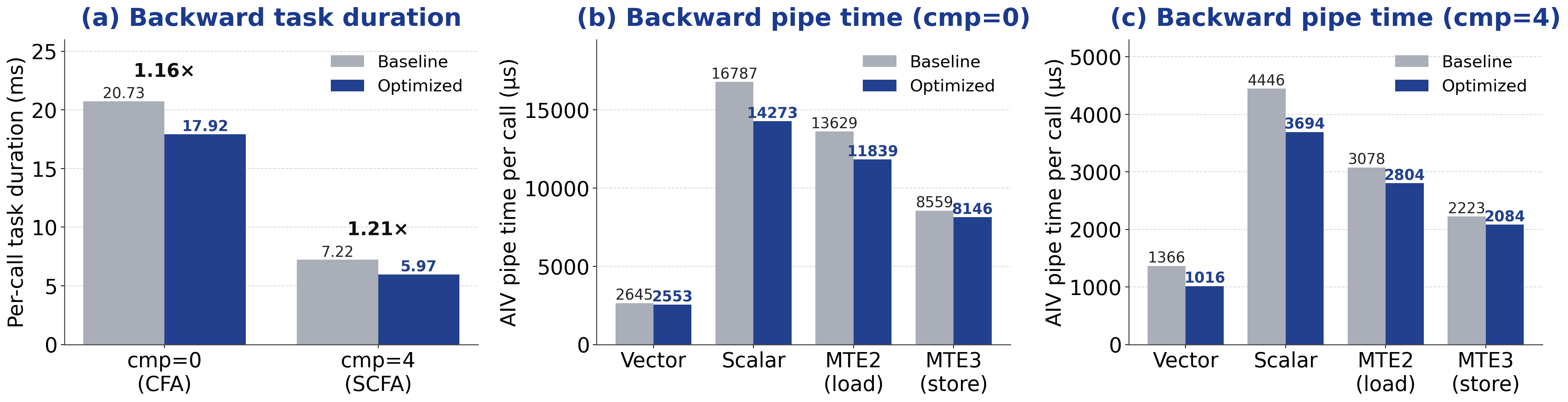}
\caption{AuraKernel optimization of the sparse lightning-indexer gradient on Ascend 910C. (a)~Per-call backward task duration speedup, $1.16\times$ on the full CFA path (cmp.\ ratio $0$) and $1.21\times$ on the SCFA path (cmp.\ ratio $4$). (b,c)~AIV pipeline-time decomposition before and after optimization at the two ratios: the kernel is scalar- and load-bound, and the reductions concentrate in the scalar and MTE2-load pipes rather than in vector compute, consistent with cutting address-computation and global-memory-load traffic.}
\label{fig:lightning_series}
\end{figure}

\paragraph{Optimization targets.}
AuraKernel is applied along the two axes identified in Section~\ref{sec:kernel_bottleneck}: the dominant architecture-intrinsic kernels and the reduction-heavy framework operators that dominate the long tail. Table~\ref{tab:kernel_results} reports representative speedups on both axes, while Figure~\ref{fig:sparse_series} and Figure~\ref{fig:lightning_series} visualize the task-duration and pipeline-metric changes for sparse attention and the lightning indexer, respectively. To keep the numbers hardware-agnostic, we describe the gains through speedups and pipeline ratios rather than absolute latencies and analyze each axis in turn below.
\begin{table}[h]
\centering
\caption{Representative operators optimized by AuraKernel on Ascend 910C, reported as single-operator task duration. The sparse-attention forward and backward and the sparse lightning-indexer gradient are per-invocation durations measured in the end-to-end training context at two compression ratios (for the indexer, cmp.\ ratio $0$ is the full CFA path with RoPE and softmax, cmp.\ ratio $4$ is the SCFA path); the RMS-normalization kernels are measured at the operator level ($tp{=}2$, input $[B{\cdot}S{=}8192,\,H{=}64,\,D{=}512]$, BF16). Speedup is baseline/optimized.}
\label{tab:kernel_results}
\resizebox{\textwidth}{!}{%
\begin{tabular}{llccc}
\toprule
Operator & Context & Baseline (ms) & Optimized (ms) & Speedup \\
\midrule
\texttt{SparseAttnSharedkv}             & cmp.\ ratio $128$ & $0.88$ & $0.72$  & $1.23\times$ \\
\texttt{SparseAttnSharedkv}             & cmp.\ ratio $4$   & $4.17$ & $3.83$  & $1.09\times$ \\
\texttt{SparseAttnSharedkvGrad}         & cmp.\ ratio $128$ & $32.9$ & $26.5$  & $1.24\times$ \\
\texttt{SparseAttnSharedkvGrad}         & cmp.\ ratio $4$   & $24.3$ & $20.1$  & $1.21\times$ \\
\texttt{SparseLightningIndexerGradKLLoss} & cmp.\ ratio $0$ (CFA)  & $20.7$ & $17.9$ & $1.16\times$ \\
\texttt{SparseLightningIndexerGradKLLoss} & cmp.\ ratio $4$ (SCFA) & $7.22$ & $5.97$ & $1.21\times$ \\
\texttt{RMSNormWithoutWeightFwd}        & operator, $tp{=}2$ & $20.9$ & $1.76$  & $11.90\times$ \\
\texttt{RMSNormWithoutWeightBwd}        & operator, $tp{=}2$ & $27.7$ & $8.09$  & $3.42\times$ \\
\bottomrule
\end{tabular}
}
\end{table}

\begin{figure}[h]
\centering
\includegraphics[width=\textwidth]{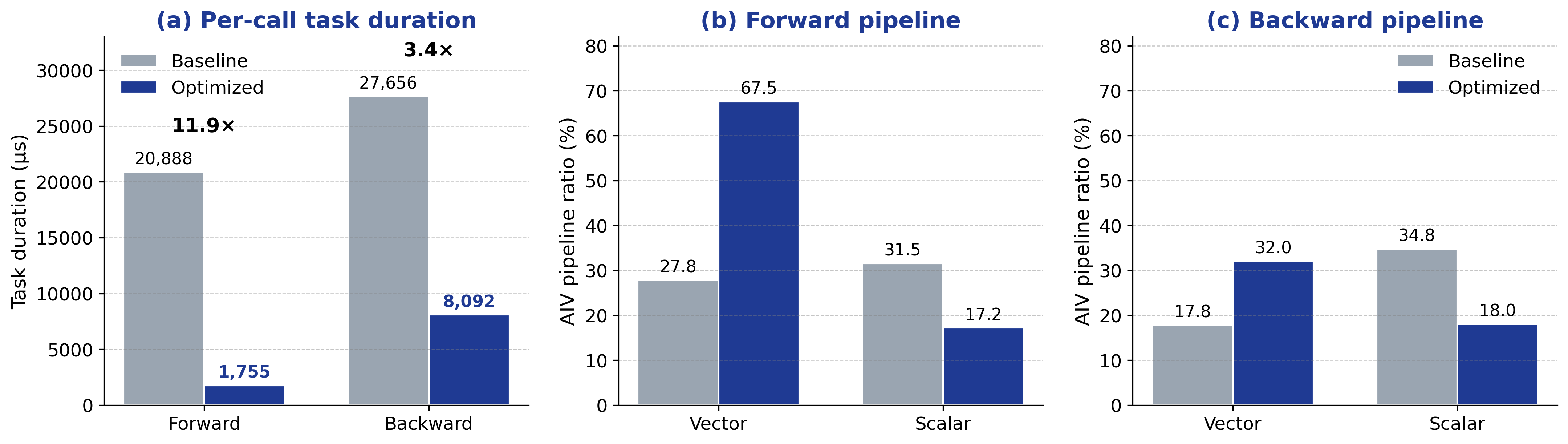}
\caption{AuraKernel optimization of the weight-free RMS-normalization kernels ($tp{=}2$). (a)~Per-call task duration speedup: forward $11.9\times$ and backward $3.4\times$. (b,c)~Duration-weighted AIV pipeline ratios before and after optimization: the vector-compute ratio rises (forward $27.8\%\rightarrow67.5\%$, backward $17.8\%\rightarrow32.0\%$) while the scalar ratio falls (forward $31.5\%\rightarrow17.2\%$, backward $34.8\%\rightarrow18.0\%$), showing that 2D head blocking, multi-row processing, and explicit FP32 accumulation turn a scalar/address-bound reduction loop into a vector-compute-dominated kernel.}
\label{fig:rms_series}
\end{figure}

\paragraph{Top hot-spot operators.}
The heavy head is formed by the architecture-intrinsic kernels of the DeepSeek-V4 hierarchical sparse attention, which Section~\ref{sec:kernel_bottleneck} identified as the largest single consumers of device time, with the shared-key-value sparse-attention backward alone at $\approx 16.33\%$ of per-step compute. Their metric-vector fingerprint is not compute-bound but movement- and scalar-bound: for \texttt{SparseAttnSharedkvGrad} the AIV load/store ratios (MTE2$+$MTE3) reach $\approx 0.559$ against a MAC ratio of only $\approx 0.10$, and the sparse lightning-indexer gradient is scalar- and load-bound (AIV scalar $\approx 0.81$, MTE2 load $\approx 0.66$ on the CFA path), so the duration floor is set by global-memory (GM) traffic and address computation rather than arithmetic. A top-$k$ locality analysis on real routing indices explains where this traffic originates: adjacent queries select almost identical compressed key-value blocks (adjacent-query Jaccard $\approx 0.998$ at $K{=}1024$), yet the baseline processes one query per task and re-gathers and re-scatters the same shared blocks once per query. AuraKernel maps this to two hardware-grounded rewrites: (i)~it batches consecutive queries per task and accesses the compressed side densely over the causal range instead of per-query sparse gather, cutting compressed-side GM traffic by roughly the batch factor; and (ii)~once batching shifts the floor onto the cube engine, it stores the heaviest FP32 cube$\rightarrow$vector hand-offs in BF16 to narrow the remaining GM round-trips, under an exact top-$k$ membership mask and gradient-magnitude and loss-stability gates that keep the rewrite numerically faithful. The gain lands on the modeled bottleneck: the sparse-attention backward improves by $1.24\times$ and $1.21\times$ at compression ratios $128$ and $4$, its forward by $1.23\times$ and $1.09\times$, and the lightning-indexer gradient by $1.16\times$ on the CFA path and $1.21\times$ on the SCFA path (Table~\ref{tab:kernel_results}, Figures~\ref{fig:sparse_series} and~\ref{fig:lightning_series}); in every case the per-call reduction concentrates in the MTE2/MTE3 load-store pipes, and additionally the scalar pipe for the indexer, rather than in vector compute, confirming that the improvement comes from removing GM traffic and address overhead rather than from raising raw compute utilization.

\paragraph{Triton-optimized operators.}
The long tail is carried by reduction-heavy framework operators expressed as Triton-Ascend kernels, which Section~\ref{sec:kernel_bottleneck} flagged as launch-fragmented and memory- and scalar-bound rather than compute-bound. Profiling localizes the cause to structural inefficiency inside these kernels: the weight-free RMS-normalization forward runs a serial per-head loop that leaves the vector engine idle behind a high scalar ratio, its backward emits poor scalar code because the reductions are not explicitly promoted to FP32, and several MHC operators accumulate through atomic-add or precede the main kernel with a standalone zero-init launch. Because these are Triton operators, AuraKernel restructures them within the Triton programming model, adjusting only host-side tiling and pipeline parameters while preserving the kernel mathematics: 2D head blocking and multi-row processing to replace the serial head loop, explicit FP32 accumulation and head-dimension unrolling, three-phase load-compute-store software pipelining to overlap the MTE2, vector, and MTE3 pipes, and elimination of atomic-add and standalone zero-init launches. A single campaign drives all $14$ Triton-Ascend operators (the RMS-normalization pair and the DeepSeek-V4 MHC family) to $100\%$ completion at a $2.06\times$ average and an $11.90\times$ best speedup, the backward RMS-normalization improving by $3.42\times$ (Table~\ref{tab:kernel_results}, Figure~\ref{fig:rms_series}). The gain shows up as a pipeline shift: the AIV vector-compute ratio rises sharply while the scalar ratio falls (RMS-normalization forward $27.8\%\rightarrow67.5\%$ vector and $31.5\%\rightarrow17.2\%$ scalar; backward $17.8\%\rightarrow32.0\%$ vector and $34.8\%\rightarrow\approx18\%$ scalar), turning scalar- and address-bound reduction loops into vector-compute-dominated kernels. The largest wins concentrate on reduction-class operators with structural inefficiency, whereas operators already at the bandwidth limit gain only a few percent.

\subsection{Performance Improvements from Kernel Fusion}
\label{subsec:eval_operator_chain_fusion}

\begin{table}[!ht]
\centering
\small
\caption{Operator-chain fusion results from the pre-optimization (fusion off) and post-optimization (fusion on) profiling traces, measured at global batch size $1024$. Operator counts are trace records attributed to the target module. ``Records'' gives the module-level count before and after fusion and its net change $\Delta$. ``Fused calls'' is the count of the inserted AscendC kernel, cross-checked against the post-optimization trace by matching operator name and type. ``Module time'' is the summed device task duration of the module before and after fusion, under the same submodule attribution. ``Per call'' divides each module time by the number of semantic invocations, so it reports how heavy one invocation of the eager chain is against its fused replacement. The speedup is identical whether read at the module or per-call level, because both share the same invocation count.}
\label{tab:operator_chain_fusion_results}
\resizebox{\textwidth}{!}{%
\begin{tabular}{@{}lccccc@{}}
\toprule
Optimization target & Fused kernel & \makecell{Records\\($\Delta$)} & \makecell{Fused\\calls} & \makecell{Module time\\(before\,$\rightarrow$\,after)} & \makecell{Per call\\($\mu$s)} \\
\midrule
mHC pre-projection &
\texttt{WindHcPreBmmForward} &
\makecell{$14336\rightarrow8064$\\$(-6272)$} &
$896$ &
\makecell{$1.287\rightarrow0.425$\,s\\$3.03\times$} &
\makecell{$1437\rightarrow474$} \\
mHC pre-projection backward &
\texttt{WindHcPreBmmBackward} &
\makecell{$21504\rightarrow17920$\\$(-3584)$} &
$448$ &
\makecell{$1.913\rightarrow1.019$\,s\\$1.88\times$} &
\makecell{$4270\rightarrow2274$} \\
mHC post-projection &
\texttt{WindMhcPostPart} &
\makecell{$7168\rightarrow3584$\\$(-3584)$} &
$896$ &
\makecell{$1.710\rightarrow0.624$\,s\\$2.74\times$} &
\makecell{$1908\rightarrow697$} \\
Limited SwiGLU forward &
\texttt{SwiGluLimitV2} &
\makecell{$5376\rightarrow896$\\$(-4480)$} &
$896$ &
\makecell{$1.480\rightarrow0.167$\,s\\$8.86\times$} &
\makecell{$1652\rightarrow186$} \\
Limited SwiGLU backward &
\texttt{SwiGluLimitBackwardV2} &
\makecell{$5824\rightarrow448$\\$(-5376)$} &
$448$ &
\makecell{$0.805\rightarrow0.246$\,s\\$3.27\times$} &
\makecell{$1797\rightarrow549$} \\
RoPE forward sites &
\texttt{WindScaleRope} &
\makecell{$14208\rightarrow2176$\\$(-12032)$} &
$2176$ &
\makecell{$1.421\rightarrow0.611$\,s\\$2.33\times$} &
\makecell{$653\rightarrow281$} \\
RoPE backward sites &
\texttt{WindScaleRopeGrad} &
\makecell{$15744\rightarrow2176$\\$(-13568)$} &
$1088$ &
\makecell{$1.599\rightarrow0.451$\,s\\$3.54\times$} &
\makecell{$1470\rightarrow415$} \\
\bottomrule
\end{tabular}}
\end{table}

We next evaluate the operator fusion and rewrite path in Section~\ref{sec:kernel_fusion}. The measurement unit is different from the AuraKernel results above. Whereas AuraKernel improves kernels that already appear as large execution units in the trace, this path changes the unit of execution itself. The baseline is therefore the complete eager chain emitted by Python, torch, and autograd for a model-level operation, or the generic Triton kernel that a stage was previously lowered into, not a single primitive operator selected from that chain. This distinction is important for interpreting the result: a fused or rewritten implementation may preserve the same mathematical work while removing the launches, intermediate tensors, dtype round trips, and scatter-style writebacks that surrounded it in the original trace, and while relocating the remaining arithmetic onto a substrate with explicit on-chip residency and pipeline scheduling. We present the targets in the order of Section~\ref{sec:kernel_fusion}: the mHC suite first, then limited SwiGLU and unified RoPE.

\begin{figure}[htbp]
\centering
\includegraphics[width=0.92\textwidth]{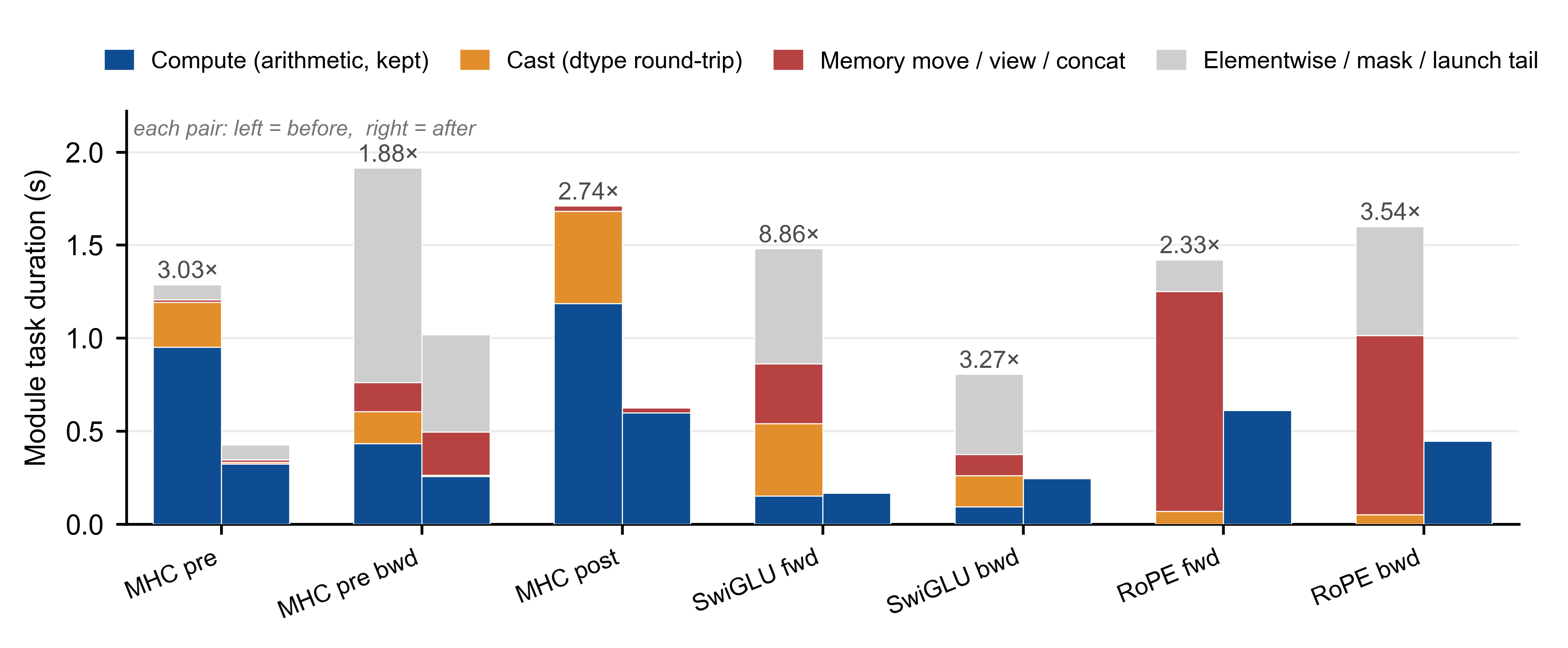}
\caption{Module task duration before (left bar) and after (right bar) operator fusion and rewrite for the seven targets, decomposed by operator category. Category durations are re-aggregated from the per-operator trace times of the pre- and post-optimization profiles. Each bar pair represents kernel-module durations~(Table~\ref{tab:operator_chain_fusion_results}), with the module speedup annotated above. Across targets, the baseline implementations are dominated by memory movement, dtype casts, and small elementwise/mask fragments, whereas the proposed fusion retain achieves significant efficiency improvements.}
\label{fig:fusion_time_breakdown}
\end{figure}

The mHC results exercise the two entries of this path together, so Table~\ref{tab:operator_chain_fusion_results} reports each stage separately. Pre-projection absorbs the RMS-normalization fragments and the original pre-BMM kernel; its backward absorbs the pre-BMM backward path, RMS backward fragments, casts, and elementwise reconstruction; and post-projection replaces the add/post-BMM1/cast portion. The three stages reach $3.03\times$, $1.88\times$, and $2.74\times$ at the module level (Table~\ref{tab:operator_chain_fusion_results}). This attribution is corroborated independently of the record counts: summing the inserted mHC fused kernels directly from the post-optimization trace reproduces the same stage totals, with \texttt{WindMhcPostPart} the single largest contributor at $0.313$\,s across all matched calls. The post-projection backward is left on its original path in this profile and is therefore not reported as a fused target, and folding it into the same boundary is left to future work.

Four of these kernels---the weight-free RMS normalization forward and backward and the pre-BMM contraction forward and backward---reach the AscendC substrate through the Triton rewrite rather than through eager-chain fusion, and comparing the original Triton kernel against its AscendC replacement at the single-operator level isolates what the substrate change alone contributes. Table~\ref{tab:triton_rewrite} reports this comparison. The RMS-normalization pair, measured at an identical BF16 shape, drops from $2307$\,\textmu s to $309$\,\textmu s forward ($7.48\times$) and from $3318$\,\textmu s to $439$\,\textmu s backward ($7.55\times$). Much of this comes from the tiling change, as the generic Triton kernel launches across $4096$ blocks whereas the AscendC kernel processes multiple rows per core across $48$ blocks. These ratios measure the Triton-to-AscendC rewrite against the original Triton kernel, and are therefore distinct from the AuraKernel figures in Section~\ref{subsec:eval_aurakernel} ($11.90\times$/$3.42\times$), which tune the resulting AscendC kernel further at a different shape. The pre-BMM contraction, whose rewrite also folds in the mixed-precision contract (FLOAT inputs become BF16), improves by $2.10\times$ forward and $1.89\times$ backward. The AIV pipeline decomposition confirms the affinity mechanism of Section~\ref{sec:kernel_fusion} rather than a mere clock reduction: on the pre-BMM forward the vector-compute ratio rises from $0.17$ to $0.86$ while the MTE2 load ratio falls from $0.59$ to $0.35$, so the \texttt{Brcb} broadcast and UB-resident contraction turn a movement-bound generic kernel into a vector-compute-dominated one. The same shift, in weaker form, appears on the backward kernels (pre-BMM backward vector ratio $0.14\rightarrow0.47$, RMS backward $0.28\rightarrow0.36$). Beyond these single-operator gains, the rewrite carries a stage-wide precision benefit that a generic Triton lowering left on the table: making the mixed-precision contract explicit removes a full-stage FP32 up-cast across the mHC pre-projection calls. Measured on the post-optimization trace, the \texttt{Cast} kernels of the stage fall from $241$\,ms to $8$\,ms in the forward path and from $172$\,ms to $6$\,ms in the backward path---about $400$\,ms of cast traffic removed across the combined stage, whose pre-optimization cost is $1.29$\,s forward plus $1.91$\,s backward---with a further $157$\,ms recovered from a backward \texttt{Mul} that the promotion had widened to FP32.

\begin{table}[h]
\centering
\caption{Triton-to-AscendC operator rewrite for the four mHC pre-projection kernels, measured as single-operator task duration at the profiled training shape. The RMS-normalization pair is compared at an identical BF16 input, while the pre-BMM pair additionally narrows its I/O dtype from FLOAT to BF16 as part of the rewrite. ``Vec.\ ratio'' is the duration-weighted AIV vector-compute ratio, whose rise on the AscendC side indicates a shift from movement-bound to compute-bound execution. Speedup is Triton/AscendC.}
\label{tab:triton_rewrite}
\begin{tabular}{lccccc}
\toprule
\multirow{2}{*}{Kernel} & \multicolumn{2}{c}{Task duration (\textmu s)} & \multirow{2}{*}{Speedup} & \multicolumn{2}{c}{Vec.\ ratio} \\
\cmidrule(lr){2-3}\cmidrule(lr){5-6}
 & Triton & AscendC & & Triton & AscendC \\
\midrule
RMS-norm forward     & $2307$ & $309$ & $7.48\times$ & $0.30$ & $0.32$ \\
RMS-norm backward    & $3318$ & $439$ & $7.55\times$ & $0.28$ & $0.36$ \\
Pre-BMM forward      & $160$  & $76$  & $2.10\times$ & $0.17$ & $0.86$ \\
Pre-BMM backward     & $510$  & $270$ & $1.89\times$ & $0.14$ & $0.47$ \\
\bottomrule
\end{tabular}
\end{table}

The remaining targets isolate the boundary-fusion effect, where autograd-visible tensor boundaries disappear. Limited SwiGLU forward collapses its eager clamp-and-gate chain into a single \texttt{SwiGluLimitV2} call, cutting the module from $1.480$\,s to $0.167$\,s ($8.86\times$), because the eager forward materialized clamp constants and a concatenation that the fused kernel keeps on chip. The backward path is more launch-dense still---fills, boolean materialization, selections, casts, \texttt{SwiGluGrad}, and concatenation---and there the fused \texttt{SwiGluLimitBackwardV2} operator removes almost the entire chain (Figure~\ref{fig:fusion_trace_schematic}). In the MoE path the same kernel also absorbs the routing-weight gradient, so the eager reduction that formed it (a full-size $\ell\cdot h$ product of $61$\,ms plus a \texttt{ReduceSum} of $20$\,ms per step, about $82$\,ms together) disappears into the fused pass at no added kernel cost. Counting the routing-gradient chain as part of the backward, the module falls from $805$\,ms to $246$\,ms ($3.27\times$). RoPE retains a larger optimized residual because the rewritten path still contains shape and indexing support operators, but the count accounting in Table~\ref{tab:operator_chain_fusion_results} exposes the same mechanism, yielding $2.33\times$ (forward) and $3.54\times$ (backward). The backward gain is the larger of the two because the fused kernel removes the eager slice-assignment scatter chain at the graph level. These module-level figures aggregate the query, key-value, and output sites.

These results close the loop with the bottleneck anatomy in Section~\ref{sec:kernel_bottleneck}. The gains are not explained by faster arithmetic alone, as Figure~\ref{fig:fusion_time_breakdown} makes explicit: they come from changing which tensors are materialized, which boundaries are visible to autograd, which dtype crosses global memory, and---for the Triton-originated stages---which substrate the arithmetic runs on. This is why the strongest effects appear on chains dominated by movement, conversion, and writeback rather than on compute-dense cube kernels, why the resource-utilization shift on the rewritten kernels turns scalar- and movement-bound loops into vector-compute-dominated ones, and why the appropriate baseline is the emitted trace slice, stage subgraph, or generic Triton kernel rather than an isolated primitive operator.

\subsection{Long-term Training Stability with Ascend SuperPOD}
\label{subsec:npu_training_stability}

Stable training is a prerequisite for meaningful downstream evaluation. If the training process is unstable, subsequent benchmark differences become difficult to interpret. In this work, we implement stable SFT training of DeepSeek-V4-Flash in a Ascend-based environment with 8 nodes and 128 Ascend 910C NPUs in total.

Using the 50K self-distilled dataset as an example, we complete a relatively full-scale SFT verification run on DeepSeek-V4-Flash. The training configuration uses \texttt{GBS=128}, \texttt{MBS=1}, and \texttt{SEQ\_LEN=8192}. The model is trained for \texttt{800} steps, and checkpoints are saved at \texttt{200}, \texttt{400}, \texttt{600}, and \texttt{800} steps. The training log shows that the job successfully reaches \texttt{iteration 800/800}, without skipped iterations or NaN iterations. At the final iteration, the \texttt{lm loss} converges to approximately \texttt{0.079}, the \texttt{mtp\_1 loss} is approximately \texttt{0.093}, and the gradient norm remains within a normal range. The overall training curve decreases continuously and then gradually becomes stable.

\begin{figure}[t]
\centering
\includegraphics[width=\linewidth]{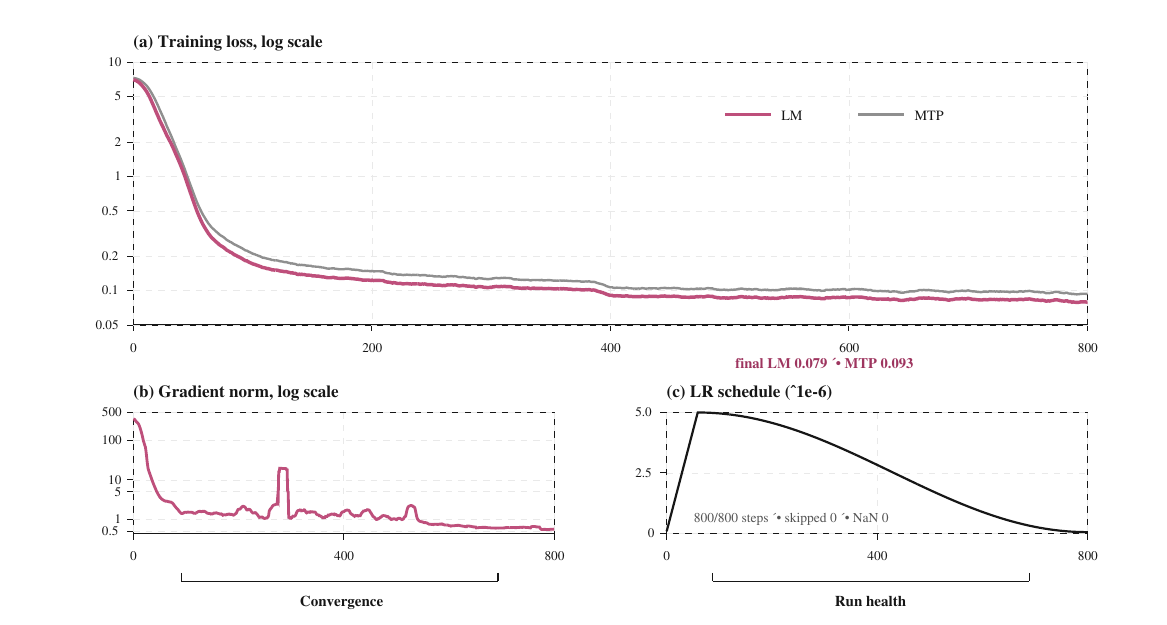}
\caption{Training stability of the 50K self-distilled SFT run on Ascend 910C NPUs. The model is trained for 800 steps with \texttt{GBS=128}, \texttt{MBS=1}, and \texttt{SEQ\_LEN=8192}, and the training process reaches the final iteration without skipped or NaN iterations.}
\label{fig:sft_50k_stability}
\end{figure}

As shown in Figure~\ref{fig:sft_50k_stability}, these observations indicate that the current data preprocessing pipeline, MindSpeed-LLM~\citep{mindspeed-llm} training script, DeepSeek-V4-Flash base model, and 8-node 910C training environment form a reproducible and sustainable full-parameter SFT workflow. This provides a stable foundation for subsequent experiments on different data scales, data mixtures, and task-oriented data-distillation strategies.

\subsection{SFT Scaling and Cleaning Analysis}
\label{subsec:sft_training_cleaning}

The first systematic experiments on 3K, 10K, and 50K self-distilled data show that SFT brings two opposite effects. On the one hand, SFT helps the model learn code-protocol tasks such as B4O-Feasible. Using DeepSeek-V4-Flash as the baseline, we find that B4O-Feasible improves from 60.47 to 65.07 with SFT-10K, and B4O-ORGEval improves from 34.26 to 47.21. These gains indicate that SFT makes the model more familiar with code-only output, \texttt{data.json} loading, Gurobi model construction, LP writing paths, and other benchmark-specific protocols, with the strongest relative improvement concentrated on structural-equivalence evaluation.

On the other hand, SFT also weakens part of the model's original ability on natural-language modeling tasks. The baseline obtains 84.08 on NL4OPT, while SFT-10K obtains 81.66. Similarly, the baseline obtains 63.33 on OptiBench, while SFT-10K obtains 62.67. Further scaling to 50K does not solve this issue; instead, OptiBench decreases to 58.68. Since the 3K, 10K, and 50K settings follow almost the same construction design, the degradation of 50K is difficult to explain as a data-mixture change. A more plausible explanation is that continuous iterative distillation introduces lower-quality samples and amplifies the model's own noise.

\begin{table}[t]
\centering
\small
\caption{Systematic comparison of SFT data scales on OR modeling benchmarks.}
\begin{tabular}{lcccc}
\toprule
Model & NL4OPT & OptiBench & B4O-Feasible & B4O-ORGEval \\
\midrule
DeepSeek-V4-Flash & 84.08 & 63.33 & 60.47 & 34.26 \\
SFT-3K  & 80.62 & 60.66 & 64.51 & 43.65 \\
SFT-10K & 81.66 & 62.67 & 65.07 & 47.21 \\
SFT-50K & 82.01 & 58.68 & 64.97 & 45.94 \\
\bottomrule
\end{tabular}

\label{tab:sft_scale}
\end{table}

Table~\ref{tab:sft_scale} supports two observations. First, supervised fine-tuning provides a clear benefit for protocol-oriented OR modeling tasks: all SFT variants improve B4O-Feasible and, more prominently, B4O-ORGEval relative to the base model, indicating that the self-distilled data contains effective supervision for solver-facing output formats and benchmark-specific conventions. Second, the gains are not monotonic with data scale. If additional self-distilled samples were uniformly beneficial, SFT-50K would be expected to dominate SFT-10K, and the uncleaned SFT-10K checkpoint would not fall below the base model on NL4OPT and OptiBench. The observed regressions suggest that scaling the same self-distillation pipeline also increases the weight of weakly verified or biased samples. Such samples can reinforce useful modeling patterns, but they can also amplify systematic errors inherited from the base model. Therefore, the 50K result should be interpreted less as evidence against SFT scale itself and more as evidence that scale must be coupled with stronger verification and data-quality control. This motivates returning to the 10K setting for fine-grained cleaning rather than further expanding the unfiltered pipeline to 100K.

The SFT results after cleaning validate the effectiveness of the contract-aware cleaning and CoT enhancement framework introduced in Section~\ref{subsec:sft_cot_cleaning}. Compared with the uncleaned SFT-10K, Clean-CoT improves NL4OPT by 5.27 percentage points, OptiBench by 1.50 percentage points, B4O-Feasible by 0.86 percentage points, and B4O-ORGEval by 1.52 percentage points. This result suggests that the main issue of the 10K data is not insufficient scale, but semantic noise that is strong enough to damage natural-language modeling ability. Once such noise is systematically filtered and repaired, SFT no longer harms NL4OPT and OptiBench, but instead outperforms both the baseline and the original SFT-10K while further improving both B4O metrics.

\begin{table}[t]
\centering
\small
\caption{Effect of data cleaning and CoT enhancement on OR modeling benchmarks.}
\begin{tabular}{lcccc}
\toprule
Model & NL4OPT & OptiBench & B4O-Feasible & B4O-ORGEval \\
\midrule
DeepSeek-V4-Flash & 84.08 & 63.33 & 60.47 & 34.26 \\
SFT-10K & 81.66 & 62.67 & 65.07 & 47.21 \\
SFT-Clean-only & 84.23 & 63.08 & 64.31 & 48.28 \\
SFT-Clean-CoT & \textbf{86.93} & \textbf{64.17} & \textbf{65.93} & \textbf{48.73} \\
\bottomrule
\end{tabular}

\label{tab:sft_cleaning}
\end{table}

However, the gains of Clean-CoT are not uniformly distributed across all tasks. It is most effective on NL4OPT and OptiBench because these targets allow \texttt{<think>}, \texttt{<model>}, and \texttt{<python>} style answers. The structured modeling checklist can directly enter the training response and help the model learn explicit judgments about variable domains, objective direction, and constraint direction. By contrast, B4O-Feasible and B4O-ORGEval are code-only tasks, where explanatory chain-of-thought cannot appear in the output. Therefore, the advantage of Clean-CoT is less directly expressed in the response format. In the updated results, Clean-CoT also outperforms clean-only on B4O-Feasible (65.93 vs.\ 64.31) and B4O-ORGEval (48.73 vs.\ 48.28), but the margins are smaller than on NL4OPT. This suggests that CoT-oriented enhancement is beneficial for code-only tasks, yet these tasks still require more specialized training on schema reading, index alignment, variable-family design, \texttt{model.write()} paths, and LP-structure consistency.

Error analysis further reveals the boundary of cleaning-based enhancement. On NL4OPT, Clean-CoT reduces \texttt{wrong\_objective\_or\_model} errors from 52 cases before cleaning to 31 cases. On OptiBench, the same error type decreases from 154 to 128 cases, indicating that modeling checklists indeed reduce coarse errors at the objective and constraint levels. However, \texttt{nonlinear\_or\_bad\_gurobi\_form} errors on OptiBench do not decrease; instead, they increase from 49 to 56 cases. This shows that ordinary cleaning and short chain-of-thought enhancement can improve the question of what model should be built, but are insufficient to stably solve the question of how to correctly express nonlinear terms, ratios, averages, and quadratic terms in Gurobi. This observation motivates the subsequent design of nonlinear repair and specialized nonlinear/Gurobi synthetic data, rather than forcing all issues into a single Clean-CoT path.

Therefore, the main conclusion of this SFT experiment is that chain-of-thought-oriented data enhancement is effective, but it must be constrained, contract-aware, and gated by reviewer and validation checks. Its value is not to make the model generate longer reasoning, but to inject the most error-prone optimization-modeling checkpoints into the training distribution in a stable form.

\subsection{Effect of CPT Initialization under Matched SFT Conditions}
\label{subsec:cpt_sft_transfer}

Following the evaluation protocol in Section~\ref{subsubsec:cpt_checkpoint_selection}, we evaluate whether the CPT checkpoint improves downstream SFT performance under matched SFT conditions. Both the direct SFT route and the CPT-initialized SFT route use the same 10K chain-of-thought OR modeling samples, 220 SFT iterations, peak learning rate \(5\times10^{-6}\) with cosine decay, global batch size 128, and sequence length 8192. Therefore, performance differences between the two routes can be attributed to CPT initialization rather than different SFT data or optimization settings.

We report results on four OR benchmarks. NL4OPT and OptiBench evaluate natural-language optimization modeling, B4O-Feasible evaluates whether the generated Gurobi program executes successfully and produces a feasible solution, and B4O-ORGEval evaluates whether the generated optimization model is structurally equivalent to the reference model, independent from implementations. Table~\ref{tab:cpt_sft_transfer} 
shows that SFT alone is already a strong baseline. Compared with the base model, direct SFT improves NL4OPT from 84.08\% to 86.93\%, OptiBench from 63.33\% to 64.17\%, B4O-Feasible from 60.47\% to 65.93\%, and B4O-ORGEval from 34.26\% to 48.73\%. This confirms that the 10K CoT OR modeling samples provide effective task-level supervision for activating optimization modeling capabilities.

\begin{table}[t]
\centering
\small
\caption{CPT-to-SFT transfer results on four OR modeling benchmarks.}
\begin{tabular}{lcccc}
\toprule
Model & NL4OPT & OptiBench & B4O-Feasible & B4O-ORGEval \\
\midrule
DeepSeek-V4-Flash
    & 84.08 & 63.33 & 60.47 & 34.26 \\
SFT-Clean-CoT
    & 86.93 & 64.17 & 65.93 & 48.73 \\
CPT+SFT-Clean-CoT
    & \textbf{89.52} & \textbf{67.12} & \textbf{71.22} & \textbf{59.39} \\
\bottomrule
\end{tabular}

\label{tab:cpt_sft_transfer}
\end{table}

CPT initialization provides further gains on top of this SFT baseline across all four benchmarks: +2.59 pp on NL4OPT, +2.95 pp on OptiBench, +5.29 pp on B4O-Feasible, and +10.66 pp on B4O-ORGEval. The gain is present on both executability and structure but is largest on B4O-ORGEval, indicating that CPT supplies OR-domain modeling priors that improve solver-facing feasibility and, most strongly, structural equivalence---capabilities that matched-condition SFT alone does not fully activate.

Since B4O-ORGEval emphasizes mathematical equivalence between the generated and reference optimization models, it is more directly aligned with the domain knowledge and modeling patterns introduced during CPT.

Overall, these results support the value of domain-adaptive CPT as a preparation stage for OR-oriented SFT. While direct SFT already improves task performance, CPT further strengthens the structurally demanding capabilities required for optimization model equivalence. This motivates scaling CPT with broader OR task coverage, higher-quality solver-verified documents, and retention-aware data mixtures in future experiments.

\subsection{Performance of the Proposed Post-Training Pipeline}

\textbf{Evaluation Setup.} We evaluate on four OR benchmarks---NL4OPT~\citep{ramamonjison2023nl4opt} and OptiBench~\citep{yang2025optibench} for natural-language optimization modeling and problem understanding, B4O-Feasible for executability, and B4O-ORGEval for structural equivalence~\citep{orgeval}---alongside general benchmarks MMLU~\citep{hendryckstest2021}, MMLU-Pro~\citep{wang2024mmlu}, CMMLU~\citep{hendryckstest2021}, HumanEval~\citep{chen2021codex}, GSM8K~\citep{cobbe2021gsm8k}, and MATH~\citep{hendrycks2021measuringmathematicalproblemsolving} (Table~\ref{tab:ours_vs_base_general}). For decoding, we use greedy for Pass@1 and sample at $T=0.8$ with top-$p=0.95$ for Pass@4, applying unbiased estimation~\citep{zhao2023unbiased}. Correctness for B4O-Feasible requires runtime execution and contract adherence; B4O-ORGEval uses the ORGEval WL reward on the normalized formulation graph to measure semantic structural match. All evaluations share the same prompt template and decoding budget for fair comparison.

While SFT improves extraction tasks, structural equivalence, and zero-shot B4O-Feasible performance, it does not consistently enhance executability across all decoding settings; the 5-shot B4O-Feasible result remains below the base model. The full CPT+SFT pipeline consistently improves 0-shot Pass@1 across the OR benchmarks and yields its strongest gains on B4O-ORGEval (structural equivalence) and B4O-Feasible (executability), but it should not be interpreted as dominating every decoding regime. This indicates that CPT supplies domain knowledge critical for both deep structural reasoning and the generation of executable code, while SFT aligns outputs with format and protocol requirements.

\begin{table*}[t!]
\centering
\caption{Performance of Base, SFT, and CPT+SFT models on operations research benchmarks. The SFT model and the SFT stage of CPT+SFT both use the Clean‑CoT dataset. All numbers are percentages. The best result in each column is in \textbf{bold}, and the second best is \underline{underlined}. Gains are absolute differences (percentage points) compared to the Base model. }
\label{tab:main_results}
\small
\setlength{\tabcolsep}{12pt}
\resizebox{\textwidth}{!}{
\begin{tabular}{llcccccc}
\toprule
\multirow{2}{*}{Dataset} & \multirow{2}{*}{Model} & \multicolumn{2}{c}{0-shot Pass@1} & \multicolumn{2}{c}{0-shot Pass@4} & \multicolumn{2}{c}{5-shot Pass@1} \\
\cmidrule(lr){3-4} \cmidrule(lr){5-6} \cmidrule(lr){7-8}
                        &                       & Value & Gain & Value & Gain & Value & Gain \\
\midrule
\multirow{3}{*}{NL4OPT} & \hspace{1em} \texttt{Base}      & 84.08           & ---   & 91.35           & ---   & 88.24           & ---   \\
                        & \hspace{1em} \texttt{SFT}       & \underline{86.93}& +2.85 & \underline{92.39}& +1.04 & \underline{88.97}& +0.73 \\
                        & \hspace{1em} \texttt{CPT+SFT}   & \textbf{89.52}  & \textcolor{slai}{+5.44} & \textbf{93.15}  & \textcolor{slai}{+1.80} & \textbf{90.66}  & \textcolor{slai}{+2.42} \\
\cmidrule(lr){1-8}
\multirow{3}{*}{OptiBench} & \hspace{1em} \texttt{Base}    & 63.33           & ---   & \underline{72.50}& ---   & \underline{66.00}& ---   \\
                            & \hspace{1em} \texttt{SFT}     & \underline{64.17}& +0.84 & 71.50           & -1.00 & 65.67           & -0.33 \\
                            & \hspace{1em} \texttt{CPT+SFT} & \textbf{67.12}  & \textcolor{slai}{+3.79} & \textbf{72.83}  & \textcolor{slai}{+0.33} & \textbf{66.89}  & \textcolor{slai}{+0.89} \\
\cmidrule(lr){1-8}
\multirow{3}{*}{B4O-Feasible} & \hspace{1em} \texttt{Base}   & 60.47           & ---   & 66.86           & ---   & \textbf{78.20}  & ---   \\
                              & \hspace{1em} \texttt{SFT}    & \underline{65.93}& +5.46 & \underline{68.69}& +1.83 & 73.55           & -4.65 \\
                              & \hspace{1em} \texttt{CPT+SFT}& \textbf{71.22}  & \textcolor{slai}{+10.75} & \textbf{82.41}  & \textcolor{slai}{+15.55}& \underline{74.78}& -3.42 \\
\cmidrule(lr){1-8}
\multirow{3}{*}{B4O-ORGEval} & \hspace{1em} \texttt{Base}    & 34.26           & ---   & 49.49           & ---   & 71.57           & ---   \\
                             & \hspace{1em} \texttt{SFT}     & \underline{48.73}& +14.47& \underline{56.35}& +6.86 & \underline{82.99}& +11.42\\
                             & \hspace{1em} \texttt{CPT+SFT} & \textbf{59.39}  & \textcolor{slai}{+25.13} & \textbf{67.28}  & \textcolor{slai}{+17.79} & \textbf{84.52}  & \textcolor{slai}{+12.95} \\
\bottomrule
\end{tabular}}
\end{table*}

To evaluate the complementary roles of knowledge injection and instruction tuning, we conduct end-to-end experiments comparing Base, SFT, and CPT+SFT under identical SFT conditions. Our central hypothesis is that CPT supplies domain-specific structural priors, while SFT grounds execution protocols and format compliance. The results in Table~\ref{tab:main_results} confirm that the full pipeline (CPT+SFT) consistently outperforms both Base and SFT across all OR benchmarks in 0-shot Pass@1. The gain is largest on B4O-ORGEval (e.g., +10.66 pp in 0-shot Pass@1 and +7.71 pp in the three-setting mean over SFT) and is also substantial on B4O-Feasible (+5.29 pp in 0-shot Pass@1 and +6.75 pp in the three-setting mean over SFT). SFT alone improves B4O-Feasible in zero-shot settings but drops below Base in 5-shot, indicating that instruction tuning by itself is insufficient to stabilize executability across prompting regimes.

\begin{figure}[t!]
    \centering
    \includegraphics[width=0.9\textwidth]{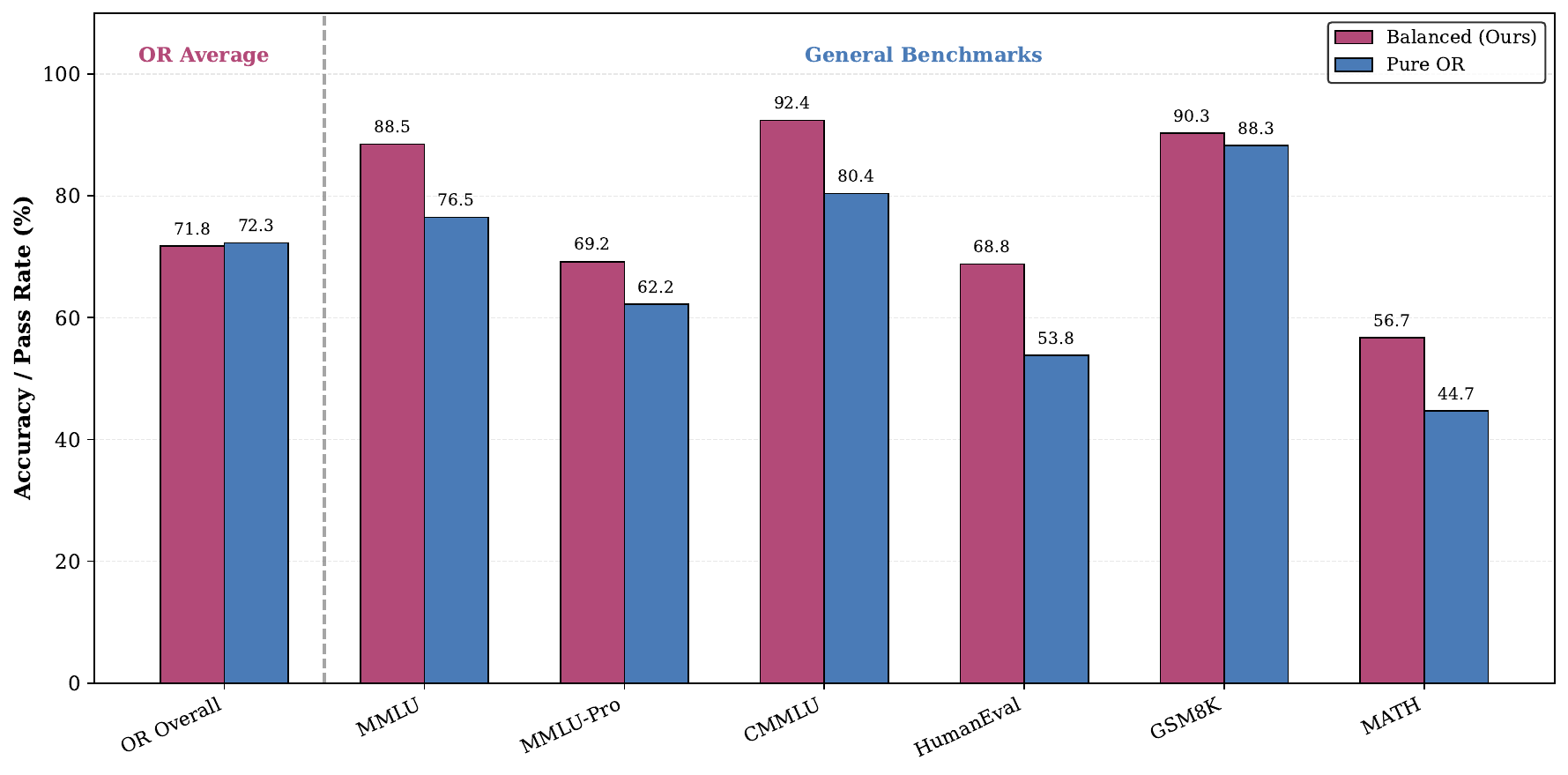}
    \caption{
        Ablation on data mixture under the end‑to‑end (CPT$\to$SFT) pipeline.
    }
    \label{fig:ablation}
\end{figure}

\begin{table*}[t]
\centering
\caption{Performance comparison between Ours and Base on general benchmarks.}
\label{tab:ours_vs_base_general}
\small
\setlength{\tabcolsep}{22pt}
\resizebox{\textwidth}{!}{
\begin{tabular}{llccc}
\toprule
\multirow{2}{*}{Category} & \multirow{2}{*}{Benchmark} & \multirow{2}{*}{\# Shots} & \multicolumn{2}{c}{Model}\\
\cmidrule(lr){4-5}
& & & \textbf{\modelname (Ours)} & Base \\
\midrule
             & MMLU~\citep{hendryckstest2021} & 5-shot & 88.5 (+0.9) & 87.6 \\
World Knowledge & MMLU-Pro~\citep{wang2024mmlu} & 5-shot & 69.2 (-1.8) & 71.0 \\
             & CMMLU~\citep{hendryckstest2021} & 5-shot & 92.4 (+0.3) & 92.1 \\
\midrule
             & HumanEval~\citep{chen2021codex} & 0-shot & 68.8 (-0.6) & 69.4 \\
Code \& Math & GSM8K~\citep{cobbe2021gsm8k} & 8-shot & 90.3 (+0.5) & 89.8 \\
             & MATH~\citep{hendrycks2021measuringmathematicalproblemsolving} & 4-shot & 56.7 (-1.7) & 58.4 \\
\bottomrule
\end{tabular}}
\end{table*}

\begin{table}[t!]
\centering
\caption{Performance of various models on operations research benchmarks. The best result in each column is in \textbf{bold}, and the second best is \underline{underlined}.}
\label{tab:models_or}
\small
\setlength{\tabcolsep}{5pt}
\resizebox{\textwidth}{!}
{
\begin{tabular}{lccccc}
\toprule
\multirow{2}{*}{Model} & \multicolumn{4}{c}{Operations Research Benchmarks} & \multirow{2}{*}{Overall}\\
\cmidrule(lr){2-5}
& NL4OPT & OptiBench & B4O-Feasible & B4O-ORGEval\\
\midrule
GLM-5.2~\citep{glm5team2026glm5vibecodingagentic}               & 78.55 & 55.33 & 61.05 & 31.47 & 56.60 \\
DeepSeek-V4-Flash~\citep{deepseekai2026deepseekv4highlyefficientmilliontoken}     & 84.08 & 63.33 & 60.47 & 34.26 & 60.54 \\
MiMo-V2.5-Pro~\citep{mimo2026v25pro}         & 80.97 & 56.83 & 70.35 & 41.88 & 62.51 \\
Kimi-K2.6~\citep{kimiteam2026kimik2openagentic}             & 78.89 & 59.83 & \underline{72.67} & \underline{46.95} & 64.59 \\
Gemini-3-Flash~\citep{google_gemini_flash}        & 79.93 & \textbf{68.83} & 71.80 & 46.70 & 66.82 \\
GPT-5.4-Mini~\citep{openai_gpt_5_4_mini}         & \underline{87.89} & 66.17 & \textbf{73.84} & 43.40 & \underline{67.83} \\
\textbf{\modelname~(Ours)}                  & \textbf{89.52} & \underline{67.12} & 71.22 & \textbf{59.39} & \textbf{71.81} \\
\bottomrule
\end{tabular}
}
\end{table}

Disaggregating by task type reveals the distinct contributions of each stage. For structural equivalence tasks (B4O-ORGEval), SFT alone improves moderately over the base model, but the additional CPT prefix yields a substantial leap. This suggests that optimization modeling judgments---which require understanding mathematical intent independent of surface form---benefit from dense exposure to OR corpora during CPT, an exposure that SFT cannot fully replicate due to its instruction-focused, smaller-scale format. For code-execution and feasibility tasks (B4O-Feasible), SFT improves 0-shot Pass@1 and Pass@4 but remains below Base in 5-shot, yielding only a modest three-setting mean gain (69.39 vs.\ 68.51), whereas CPT+SFT reaches 76.14. This suggests that domain-specific pretraining is crucial for generating executable optimization code, beyond what format and protocol alignment alone can provide. On general benchmarks (Table~\ref{tab:ours_vs_base_general}), CPT+SFT achieves performance comparable to Base, with slight fluctuations in both directions (e.g., +0.9 on MMLU~\citep{hendryckstest2021}, -1.7 on MATH~\citep{hendrycks2021measuringmathematicalproblemsolving}), and no signs of catastrophic forgetting. This confirms that the staged data mixture (\textasciitilde 75\% OR + retained general corpora) preserves broad reasoning capabilities. As shown in Table~\ref{tab:models_or}, our CPT+SFT model attains the highest overall score across all four OR benchmarks, outperforming strong baselines such as GPT-5.4-Mini~\citep{openai_gpt_5_4_mini} and Gemini-3-Flash~\citep{google_gemini_flash} by a substantial margin.

To dissect the sources of CPT's effectiveness, we ablate the CPT data mixture under the same end-to-end pipeline (Figure~\ref{fig:ablation}). We compare two configurations: Balanced (our mixture of OR and general data) and Pure OR (only OR data). As shown, Pure OR causes a drastic drop (10–15 points) on general benchmarks while yielding no improvement on OR Overall compared to Balanced---in fact, it slightly underperforms Balanced. This indicates that exclusive domain pretraining leads to overfitting that harms general capabilities without benefiting domain performance. In contrast, Balanced achieves strong results on both OR and general benchmarks, demonstrating that interleaving domain-specific and general data during CPT is essential to inject structural priors while retaining broad reasoning abilities and guarding against catastrophic forgetting.

In summary, the end-to-end experiments show a complementary pattern between
SFT and CPT. SFT establishes supervised task alignment and protocol-oriented
output behavior, while CPT supplies domain-structural priors that drive the
largest additional gains, most clearly on B4O-ORGEval and also on
B4O-Feasible. Together, the two stages form a complementary pipeline for
improving OR-domain performance while preserving general capability.

\subsection{Scale-Up Validation on DeepSeek-V4-Pro}
\label{subsec:pro_scaleup}

The preceding DeepSeek-V4-Flash experiments characterize the effects of SFT
data scaling and cleaning, CPT initialization, and retention-aware data-mixture
design. We next evaluate whether the resulting OR-oriented post-training
workflow remains effective when applied to DeepSeek-V4-Pro, the
trillion-parameter model used in our Ascend system study.

The Pro SFT stage follows the same Clean-CoT data construction and supervised
fine-tuning methodology used for DeepSeek-V4-Flash. We compare the original
DeepSeek-V4-Pro model, an SFT-only checkpoint, and the complete
SLAI T-Rex-Pro configuration obtained through CPT followed by SFT. The Pro
CPT configuration is not separately ablated in this report; consequently, the
results in this subsection are used to evaluate the end-to-end scalability of
the complete workflow, while the isolated effect of CPT initialization is
examined through the matched-condition Flash experiments in
Section~\ref{subsec:cpt_sft_transfer}.

\begin{table*}[t]
\centering
\small
\caption{
Scale-up validation of SLAI T-Rex on DeepSeek-V4-Pro across OR benchmarks
under zero-shot Pass@4 evaluation. Overall is the unweighted mean of NL4OPT,
OptiBench, B4O-Feasible, and B4O-ORGEval. Overall Gain denotes the absolute
improvement over the original DeepSeek-V4-Pro model. All numbers are
percentages, and the best result in each benchmark column is in
\textbf{bold}.
}
\label{tab:pro_or_scaleup}
\setlength{\tabcolsep}{7pt}
\resizebox{\textwidth}{!}{
\begin{tabular}{lcccccc}
\toprule
Model
& NL4OPT
& OptiBench
& B4O-Feasible
& B4O-ORGEval
& Overall
& Overall Gain \\
\midrule
DeepSeek-V4-Pro
& 86.51
& 70.00
& 80.23
& 43.91
& 70.16
& --- \\

DeepSeek-V4-Pro + SFT
& 89.97
& 69.50
& \textbf{86.04}
& 61.43
& 76.74
& +6.58 \\

\textbf{SLAI T-Rex-Pro}
& \textbf{92.04}
& \textbf{70.17}
& 85.17
& \textbf{61.93}
& \textbf{77.33}
& \textbf{+7.17} \\
\bottomrule
\end{tabular}}
\end{table*}

As shown in Table~\ref{tab:pro_or_scaleup}, OR-oriented post-training
substantially improves DeepSeek-V4-Pro.
Applying the Clean-CoT SFT workflow raises the average OR score from 70.16\%
to 76.74\%. In particular, NL4OPT improves from 86.51\% to 89.97\%,
B4O-Feasible from 80.23\% to 86.04\%, and B4O-ORGEval from 43.91\% to
61.43\%, while OptiBench changes slightly from 70.00\% to 69.50\%. These
results show that the SFT methodology developed and analyzed on
DeepSeek-V4-Flash remains effective when applied to the substantially larger
Pro model, with the largest gain appearing on B4O-ORGEval.

The complete SLAI T-Rex-Pro configuration achieves the highest average OR
score of 77.33\%, corresponding to a 7.17-percentage-point improvement over
the original model. It improves NL4OPT from 86.51\% to 92.04\%,
OptiBench from 70.00\% to 70.17\%, B4O-Feasible from 80.23\% to
85.17\%, and B4O-ORGEval from 43.91\% to 61.93\%. The largest improvement
is again observed on B4O-ORGEval, indicating that SLAI T-Rex-Pro
substantially strengthens the model's ability to produce optimization
formulations that preserve the mathematical structure of the target problem.

Compared with SFT alone, SLAI T-Rex-Pro further improves NL4OPT from
89.97\% to 92.04\%, OptiBench from 69.50\% to 70.17\%, and B4O-ORGEval
from 61.43\% to 61.93\%, while obtaining a slightly lower B4O-Feasible score
of 85.17\% versus 86.04\%. Overall, the complete configuration achieves the
stronger aggregate OR result, increasing the average score from 76.74\% to
77.33\%.

\begin{table}[t]
\centering
\small
\caption{
Complementary general-capability retention check for the DeepSeek-V4-Pro
scale-up experiment. All numbers are percentages, and the best result in each
column is in \textbf{bold}.
}
\label{tab:pro_general_retention}
\setlength{\tabcolsep}{4pt}
\begin{tabular}{lccccc}
\toprule
Model
& AIME 2024
& AIME 2025
& CMMLU
& HumanEval
& LiveCodeBench \\
\midrule
DeepSeek-V4-Pro
& 83.33
& 70.00
& \textbf{93.88}
& \textbf{100.00}
& 55.00 \\

DeepSeek-V4-Pro + SFT
& 83.33
& 76.67
& 93.75
& \textbf{100.00}
& \textbf{72.00} \\

SLAI T-Rex-Pro
& \textbf{90.00}
& \textbf{86.67}
& 93.12
& \textbf{100.00}
& 67.00 \\
\bottomrule
\end{tabular}
\end{table}

Table~\ref{tab:pro_general_retention} provides a complementary check of
general-capability retention. SLAI T-Rex-Pro improves AIME 2024 from
83.33\% to 90.00\%, AIME 2025 from 70.00\% to 86.67\%, and
LiveCodeBench from 55.00\% to 67.00\%. HumanEval remains unchanged, while
CMMLU decreases slightly from 93.88\% to 93.12\%. The SFT-only checkpoint
obtains the highest LiveCodeBench result, whereas SLAI T-Rex-Pro performs best
on the two AIME benchmarks. Overall, the substantial OR improvements are not
accompanied by broad degradation on the evaluated general tasks.

The DeepSeek-V4-Pro experiment provides scale-up validation of SLAI T-Rex at
the trillion-parameter level. The complete workflow improves the average OR
score by 7.17 percentage points, with its largest gain on B4O-ORGEval, while
maintaining broadly stable performance on the evaluated general benchmarks.
These results show that the OR-oriented post-training workflow developed on
DeepSeek-V4-Flash remains effective when deployed on the optimized Ascend
SuperPOD training stack.

\section{Conclusion, Limitations, and Future Directions}
\label{sec:conclusion_future}

This technical report presents a first-stage practice for full-parameter
post-training of the DeepSeek-V4 model family on Ascend NPU SuperPOD. We refer
to the overall framework as \textbf{\modelname}, which couples full-stack
Ascend SuperPOD training optimization with solver-grounded CPT--SFT
specialization for OR.

At the system level, we demonstrate one of the first full-stack optimization
efforts for trillion-parameter MoE LLM training on a non-GPU computing
platform. The proposed end-to-end optimization strategy spans system-level
training orchestration and bottleneck-level kernel optimization, achieving up
to 34.22\% MFU. These results validate both the training efficiency and
practical feasibility of large-scale post-training on Ascend NPU SuperPOD.

For post-training tasks, we establish an OR-oriented CPT--SFT workflow on
DeepSeek-V4-Flash. Under identical SFT settings, direct SFT improves
B4O-Feasible from 60.47\% to 65.93\% and B4O-ORGEval from 34.26\% to
48.73\%. With CPT initialization, the corresponding scores further increase
to 71.22\% and 59.39\%, respectively. The full benchmark results show the
same overall trend across NL4OPT and OptiBench. These results suggest that
direct SFT provides supervised task alignment, while CPT initialization
introduces additional OR-domain modeling knowledge that improves both
executable optimization-program generation and formulation correctness. The
largest additional improvement is observed on B4O-ORGEval, highlighting the
value of domain-adaptive pre-training for complex OR modeling.

We further apply the OR-oriented workflow to DeepSeek-V4-Pro as a
trillion-parameter scale-up validation. The
SFT-only checkpoint increases the average OR score from 70.16\% to 76.74\%,
while the complete SLAI T-Rex-Pro configuration further reaches 77.33\%.
These results show that the post-training workflow developed and analyzed on
DeepSeek-V4-Flash remains effective when applied to the larger Pro model on
the optimized Ascend SuperPOD training stack.

The SFT experiments further highlight the importance of data quality. Scaling
self-distilled data from 10K to 50K does not lead to monotonic improvements
and may even degrade performance on natural-language modeling benchmarks. In
contrast, contract-aware cleaning and concise CoT-oriented enhancement improve
the 10K data more reliably, especially on NL4OPT, OptiBench, and ORGEval.
This indicates that OR-domain SFT benefits more from targeted repair of
modeling errors, strict output-contract enforcement, compact reasoning
checklists, and solver-grounded verification than from simply increasing the
volume of synthetic training data.

\paragraph{Reproducibility and artifact release.}
To support full reproducibility and facilitate follow-up research, we release
the key artifacts underlying the proposed optimization and training workflow,
including prompt templates, format-contract specifications, curriculum
scheduling configurations, and monitoring callbacks. The AI-assisted data and
evaluation tools used in our pipeline, including Cleaner, Reviewer, and
Diagnostic Resolver, are implemented as configurable prompt- and rule-driven
modules. Their templates and configuration files are also released, allowing
future users to inspect, reproduce, adapt, and extend the data-cleaning,
sample-review, diagnostic-resolution, and monitoring procedures introduced in
this work.

In addition, we plan to separately release the optimized AscendC kernels
developed through our kernel-optimization pipeline. This release will provide
concrete kernel-level implementations and optimization examples, enabling
future work to reproduce the reported operator-level improvements and further
adapt the optimized kernels to related Ascend-based training workloads.

\paragraph{Future directions.}
Building on the proposed high-performance training infrastructure for Ascend
SuperPOD, future work will further improve the efficiency of end-to-end
full-parameter training while extending the post-training workflow toward
agentic reinforcement learning~(AgenticRL). In particular, we will investigate
operations-research-based mathematical modeling for training parallelism
optimization, aiming to automate the exploration of multi-dimensional parallel
configurations under memory, communication, and orchestration constraints. In
parallel, we will continue scaling OR-oriented continued pre-training~(CPT)
and supervised fine-tuning~(SFT) to broader solver-verified corpora, extend
evaluation to more diverse OR problem families and general-capability
benchmarks, and strengthen structural verification for nonlinear expressions,
ratio constraints, quadratic terms, and Gurobi-specific modeling patterns.

Overall, this report demonstrates that Ascend NPU SuperPOD can support stable
full-parameter post-training of trillion-scale models with MoE and hybrid
attention architectures. Furthermore, the proposed CPT--SFT workflow produces
measurable gains in OR-oriented mathematical modeling across both
DeepSeek-V4-Flash and DeepSeek-V4-Pro. The key finding is that effective
vertical-domain adaptation depends on the alignment among domain knowledge
acquisition, solver-grounded data construction, supervised data quality,
system stability, and task-specific evaluation.


\newpage
\phantomsection
\addcontentsline{toc}{section}{References}
\bibliographystyle{slai}
\bibliography{custom}

\newpage

\appendix
\phantomsection
\section*{Appendix}
\addcontentsline{toc}{section}{Appendix}
\setcounter{section}{0}
\setcounter{subsection}{0}

\section{Author List}

\paragraph*{Core Contributors} Dongfang Li, Xiaodong Luo, Ruoyu Sun, Xuhui Chen, Linyuan Qiu, Jian Meng, Zhengxuan Lu, Yiting Wang, Yucheng Xie, Tao Guo, Tianxiang Fang, Jing Li, Sihang Chen, Shihao Hong, Chang Liu, Weihua Dai, Zirong Zeng, Ziwei Zhu, Zhuohan Wang, Zhengjun Yue, Igor Vasilyev, Min Liu, Weijian Sun,  Xin Chen 
\paragraph*{Contributors} Yingmeng Gao, Jinhua Zhou, Taolue Chen, Chenwei Wu, Dong Zhang, Wenlong Jin, Jinmin Xiang, Barkova Maria, Ushakov Anton, Xianfei Jin, Tian Ding, Zhihang Lin, Qian Chen, Linxin Yang, Mingzhe Yang, Bingwei Zhang, Hongzhang Yang, Fangxue Zhang, Shijun Qin, Jie Yu, Cuihua Hu, Tolstykh Vasiliy, Nosov Ivan, Abdullin Amir, Zhicheng Zhou, Xin Zhang, Zhixiong Ning, Xutong Zhao, Junjie Huang, Jiajun Liu, Weiyan Kong, Zheng Zhang, Wenhan Luo, Lin Hu, Yangbo Guo, Li Zeng, Shihao Zhang 
\paragraph*{Project Leaders} Baotian Hu, Min Zhang, Haizhou Li, Zhiquan Luo

\section{Post-training Recipe Details} \label{app:post_training_recipes} This appendix summarizes the main recipe-level settings used in the post-training experiments, focusing on configurations that affect optimization, parallel execution, memory usage, and reproducibility.

\paragraph{CPT recipe.} Table~\ref{tab:cpt_training_recipe} reports the CPT recipe used for the Ascend-based DeepSeek-V4 continued-pretraining run, including the long-context setup, optimization configuration, parallelism strategy, and memory-saving mechanisms.

\paragraph{SFT recipe.}
Table~\ref{tab:sft_training_recipe} reports the main recipe-level configuration
used for supervised fine-tuning. In contrast to CPT, which optimizes
autoregressive language modeling over domain-adaptive corpora, SFT optimizes
instruction-response trajectories and masks the loss on prompt tokens so that
gradients are applied only to the target response. The reported items are
selected to match the capabilities discussed in Section~\ref{sec:sft}, including
instruction-following alignment, solver-oriented modeling, and
pipeline-optimized distributed execution on Ascend NPUs.


\paragraph{Agentic rollout configuration.}
The agentic rollout pipeline is evaluated in inference mode and does not update
model parameters in this work. Its configuration is therefore separated from
the CPT and SFT training recipes. If reported, the rollout configuration should
include only parameters that affect inference-time behavior and future
reinforcement-learning reproducibility, such as the number of sampled
formulation candidates, maximum self-review rounds, maximum code-repair rounds,
solver timeout, execution success criterion, and objective-matching tolerance.
These settings are rollout parameters rather than optimizer or training
hyperparameters.

% ...

%
%
%

\definecolor{ORGray}{RGB}{245,247,249}

\lstdefinestyle{orcptappendix}{
  basicstyle=\ttfamily\footnotesize,
  breaklines=true,
  columns=fullflexible,
  keepspaces=true,
  frame=single,
  rulecolor=\color{black!25},
  backgroundcolor=\color{ORGray},
  showstringspaces=false,
  xleftmargin=0.5em,
  xrightmargin=0.5em
}

\section{Solver-Verified OR-CPT Data Synthesis: Engine Design and Illustrative Cases}
\label{app:or-cpt-engine}

Section 3.2 introduced the synthesized component of the operations-research
(OR) corpus and its bidirectional construction pipeline.  This appendix focuses
on the method: how a parameterized optimization generator is converted into a
self-contained CPT document, how mathematical invariants are preserved across
language-model transformations, and how deterministic checks prevent plausible
but incorrect samples from entering the training corpus.

The central design principle is that language-model fluency is not used as a
proxy for mathematical correctness.  Language models perform two semantic
transformations---from a verified optimization instance to a business problem,
and from that business problem back to a mathematical model and executable
program.  Generator contracts, static checks, and independent solver execution
surround these transformations and determine whether the resulting document is
admissible for CPT.

\begin{table}[H]
\centering
\small
\caption{Main recipe-level configuration for the Ascend-based CPT run under
the DeepSeek-V4-Flash long-context setting. The table reports settings
that affect optimization, parallel execution, memory behavior, and
NPU-specific training support. Dataset composition, checkpoint paths, logging
options, profiling settings, and environment-specific launch details are
omitted.}
\label{tab:cpt_training_recipe}
\begin{tabular}{p{0.36\linewidth}p{0.52\linewidth}}
\toprule
Configuration item & Setting \\
\midrule
\multicolumn{2}{c}{\textit{Training Setup}} \\
\midrule
Training stage & Full-parameter continued pre-training \\
Sequence length & 65,536 tokens \\
Global batch size & 128 sequences \\
MTP & Enabled \\
\midrule
\multicolumn{2}{c}{\textit{Optimization}} \\
\midrule
Optimizer & AdamW \\
Peak learning rate & $1.0\times10^{-6}$ \\
Learning-rate schedule & Cosine decay \\
Warmup rate & 20\% \\
Minimum learning-rate factor & 0.01 \\
Optimizer epsilon & $1.0\times10^{-6}$ \\
\midrule
\multicolumn{2}{c}{\textit{Parallelism}} \\
\midrule
Tensor parallel degree & 1 \\
Pipeline parallel degree & 1 \\
Context parallel degree & 16 \\
Expert parallel degree & 32 \\
FSDP-style sharding & Enabled with reshard-after-forward \\
\midrule
\multicolumn{2}{c}{\textit{Memory and NPU Support}} \\
\midrule
Activation checkpointing & Full \\
Optimizer memory strategy & Swap optimizer \\
Model-specific NPU kernels & Adapted for attention and MoE computation \\
\bottomrule
\end{tabular}
\end{table}
\begin{table}[t]
\centering
\small
\caption{Main recipe-level configuration for the SFT run. The table reports settings that affect optimization behavior, reproducibility, and the instruction-tuning experiments discussed in Section~\ref{sec:sft}.}
\begin{tabular}{p{0.36\linewidth}p{0.52\linewidth}}
\toprule
Configuration item & Setting \\
\midrule
\multicolumn{2}{c}{\textit{Training Setup}} \\
\midrule
Training stage & Supervised fine-tuning\\
Sequence length & 8,192 tokens \\
Global batch size & 128 sequences \\
MTP & Enabled \\
\midrule
\multicolumn{2}{c}{\textit{Optimization}} \\
\midrule
Optimizer & AdamW \\
Peak learning rate & $5.0\times10^{-6}$ \\
Minimum learning rate & $5.0\times10^{-8}$ \\
Learning-rate schedule & Cosine decay \\
Weight decay & $1.0\times10^{-2}$ \\
Gradient clipping & Maximum norm of 1.0 \\
Initial loss scale & 65,536 \\
\midrule
\multicolumn{2}{c}{\textit{Parallelism}} \\
\midrule
Tensor parallel degree & 1 \\
Pipeline parallel degree & 4 \\
Context parallel degree & 1 \\
Context parallel algorithm & Ulysses CP \\
Expert parallel degree & 32 \\
Sequence parallel & Enabled \\
\midrule
\multicolumn{2}{c}{\textit{Memory and Execution}} \\
\midrule
Activation checkpointing & Full \\
Optimizer memory strategy & Swap optimizer \\
MTP memory-efficient logits & Enabled \\

\bottomrule
\end{tabular}
\label{tab:sft_training_recipe}
\end{table}
\subsection{Role in the post-training workflow}
\label{app:orcpt-role}

The OR-CPT Engine is an upstream data-construction system.  It is separate from
the Ascend 910C training stack: the engine establishes mathematical fidelity,
semantic coverage, and provenance, while the training stack tokenizes, packs,
samples, and optimizes the model on the resulting corpus.  The engine exports
plain self-contained documents for autoregressive continued pre-training rather
than instruction--response conversations.

This separation also clarifies the relationship between CPT and SFT.  CPT
documents expose the model to OR terminology, formulation patterns, solver APIs,
and verification habits.  SFT subsequently teaches the model to elicit that
knowledge under task instructions and output contracts.  The same OR taxonomy
can guide both stages, but the data representations and optimization objectives
remain distinct.

\subsection{End-to-end engine architecture}
\label{app:orcpt-architecture}

The engine follows a bidirectional verification path.  It first samples a
structured optimization instance from a parameterized generator, solves the
source model independently, and filters infeasible, degenerate, numerically
unstable, or duplicated seeds.  The verified seed is then backtranslated into a
self-contained business problem, where a natural-language quality filter checks
numeric coverage, unit consistency, business grounding, and solution leakage.
Finally, the engine reconstructs the model from the generated problem, applies
static and contract-based validation, executes the reconstructed model, compares
its status and objective against the independent reference, and renders accepted
pairs into CPT documents with lineage metadata.

For a seed $s$, let $M_s$ denote the source model and $z_s^\star$ its reference
objective.  Backtranslation yields a natural-language problem $q_s$, and forward
modeling yields an executable reconstruction $\widehat{M}_s$ with objective
$\widehat{z}_s$.  The core admission condition can be summarized as

\begin{equation}
  \operatorname{Accept}(s)=
  \mathbb{I}\!\left[
    Q_{\mathrm{seed}}(M_s)
    \land Q_{\mathrm{NL}}(q_s)
    \land Q_{\mathrm{contract}}(\widehat{M}_s)
    \land Q_{\mathrm{exec}}(\widehat{M}_s)
    \land Q_{\mathrm{obj}}(z_s^\star,\widehat{z}_s)
  \right].
  \label{eq:orcpt-acceptance}
\end{equation}

\begin{figure}[H]
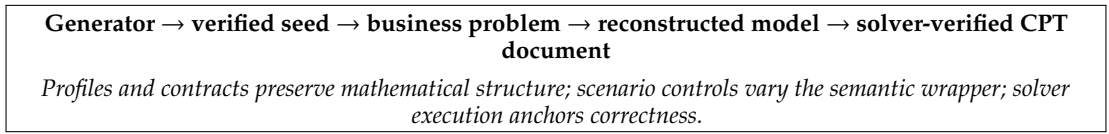

  \centering
  \fbox{%
    \begin{minipage}{0.94\linewidth}
      \centering\small
      \textbf{Generator}\ $\rightarrow$\
      \textbf{verified seed}\ $\rightarrow$\
      \textbf{business problem}\ $\rightarrow$\
      \textbf{reconstructed model}\ $\rightarrow$\
      \textbf{solver-verified CPT document}

      \vspace{0.5em}
      \textit{Profiles and contracts preserve mathematical structure; scenario
      controls vary the semantic wrapper; solver execution anchors correctness.}
    \end{minipage}}
  \caption{The trust path of a synthesized OR-CPT document.}
  \label{fig:orcpt-trust-path}
\end{figure}

The objective comparison uses absolute and relative tolerances.  It is a strong
executable consistency signal, but not a proof of symbolic equivalence: two
different formulations can share the same optimum on a particular numerical
instance. The static and contract-based checks therefore remain necessary even
when the objective agrees. Figure~\ref{fig:orcpt-trust-path} summarizes the
resulting end-to-end trust path from a generated seed to an accepted,
solver-verified CPT document.

\subsection{Mathematical seeds and domain contracts}
\label{app:orcpt-contracts}

\paragraph{Seed representation.}
A seed is a structured optimization instance, not merely a textual question.
It contains declared sets, coefficient tables, objective direction, variable
domains, bounds, constraint families, and executable or LP artifacts.  It also
carries semantic metadata such as task family, sub-family, modeling concepts,
difficulty axes, and formulation variant.  Deterministic random seeds support
regeneration when a generator must be audited or refined.

\paragraph{Generator profiles and family contracts.}
A generator profile describes the canonical mathematical signature of one
generator, including variable types, objective terms, core constraints, valid
indices, and forbidden reinterpretations.  A family contract describes
invariants shared across related generators, such as balance equations, linking
rules, conservation laws, and required variable families.  These layers are
grounded in the source facts of the seed---declared sets, coefficient tables,
bounds, units, and executable artifacts---so that the language model cannot
invent numbers or refer to nonexistent indices.  Seed-quality records further
capture feasibility, objective recomputation, solution activity, degeneracy, and
numerical stability, while lineage metadata records the generator, scenario
configuration, prompt revision, validation status, and rendering status outside
the training text.


\paragraph{Seed-quality filtering.}
Feasibility alone does not guarantee training value.  A model in which all
meaningful decisions are zero, every decision is fixed at a bound, or no
constraint participates in a genuine trade-off may be solvable but unhelpful.
The engine therefore checks solution activity, constraint activity, objective
recomputation, numerical precision, and duplicate signatures before a seed is
eligible for language generation.

\subsection{Prompt-controlled bidirectional synthesis}
\label{app:orcpt-prompts}

Backtranslation and forward modeling have different information boundaries and
are controlled by complementary fidelity requirements.  Backtranslation may
inspect the verified source facts, but it must preserve coefficients, bounds,
units, and indexed relations while hiding solver status, objective values,
variable solutions, LP artifacts, and code.  It may vary the industry setting,
stakeholder perspective, planning trigger, horizon, naming style, and units, but
it must not add or remove mathematical structure.  Forward modeling receives the
generated problem and the applicable structural contracts; it must reconstruct
the variables, objective sense, core constraints, index policy, explanation,
formulation, standalone Gurobi code, and self-check using only the declared data,
without hard-coding the reference answer or inferring missing coefficients from
the optimum.


The production prompts are versioned implementation artifacts.  Their essential
logic can be represented without reproducing their full text:

\begin{lstlisting}[style=orcptappendix,caption={Abstract prompt interface used by the bidirectional synthesis stage.}]
BACKTRANSLATE(verified_seed, profile, scenario):
    preserve source facts and mathematical structure
    hide reference solution and implementation artifacts
    return a complete business problem

FORWARD_MODEL(business_problem, profile, family_contract):
    reconstruct explanation, formulation, and executable code
    use only declared indices and coefficients
    return a structured self-check; do not hard-code the optimum
\end{lstlisting}

The self-check is treated as a claim made by the generated response, not as the
validator.  Static inspection and independent execution remain authoritative
because prose may mention a required constraint that the code omits or indexes
incorrectly.

\subsection{Controlled scenarios and CPT rendering}
\label{app:orcpt-scenarios}

The scenario catalog separates mathematical diversity from linguistic
diversity.  A base OR structure can be expressed through different industry
lenses, decision triggers, organization types, planning frames, entity-naming
styles, and unit conventions.  These controls change the surface narrative---for
example, whether the problem is framed as a healthcare allocation, logistics
planning, energy scheduling, manufacturing review, outage response, budget
revision, or service-network decision---while keeping the source coefficient
tables, variable roles, objective terms, constraints, and mathematical periods
invariant.


Negative guidance is equally important.  A transportation matrix must not be
turned into vehicle routing unless vehicles and sequencing exist.  A facility-
location task must retain its fixed opening decisions.  A lot-sizing task must
retain its period-by-period inventory balance.  These restrictions prevent a
fluent scenario from silently changing the model family.

After executable verification, the renderer converts the accepted pair into a
self-contained CPT document.  A full document can contain: (i) the business
problem and data; (ii) a business-to-model explanation; (iii) a mathematical
formulation; (iv) standalone solver code; (v) an interpreted solution; and
(vi) feasibility and objective-recalculation checks.  Style and section order
may vary, but mandatory mathematical and validation content does not.

\subsection{Illustrative case I: capacitated facility location}
\label{app:orcpt-case-cflp}

Suppose a regional service operator considers three candidate sites serving
three demand zones.  Opening a site incurs a fixed cost; service can be allocated
only from opened sites and cannot exceed site capacity.  The illustrative data
are shown in Table~\ref{tab:orcpt-example-data}.

\begin{table}[H]
  \centering
  \caption{Manually constructed facility-location example.}
  \label{tab:orcpt-example-data}
  \small
  \begin{tabular}{@{}lrrrrr@{}}
    \toprule
    Site & Fixed cost & Capacity & Cost to Z1 & Cost to Z2 & Cost to Z3 \\
    \midrule
    A & 120 & 100 & 4 & 7 & 6 \\
    B &  95 &  65 & 6 & 3 & 5 \\
    C &  80 &  60 & 8 & 5 & 2 \\
    \midrule
    Demand & -- & -- & 40 & 35 & 25 \\
    \bottomrule
  \end{tabular}
\end{table}

Let $y_i\in\{0,1\}$ indicate whether site $i$ is opened and let
$x_{ij}\geq0$ denote units served from site $i$ to zone $j$.  A faithful model
is

\begin{align}
  \min_{x,y}\quad
    & \sum_{i\in I} f_i y_i
      + \sum_{i\in I}\sum_{j\in J} c_{ij}x_{ij},
      \label{eq:example-objective}\\
  \text{s.t.}\quad
    & \sum_{i\in I}x_{ij}=d_j,
      && \forall j\in J,\\
    & \sum_{j\in J}x_{ij}\leq u_i y_i,
      && \forall i\in I,\\
    & y_i\in\{0,1\},\quad x_{ij}\geq0.
\end{align}

The optimal illustrative plan opens B and C.  Site B serves 30 units of Z1 and
all 35 units of Z2; site C serves the remaining 10 units of Z1 and all 25 units
of Z3.  Site B is exactly at its capacity of 65.  The objective is

\begin{equation}
  95+80+6(30)+8(10)+3(35)+2(25)=590.
  \label{eq:example-recalculation}
\end{equation}

This example contains a genuine fixed-charge trade-off.  Site C has the lowest
fixed cost and the cheapest service to Z3, while B is cheapest for Z2.  Neither
can serve total demand alone under the stated capacities.  The model must jointly
choose sites and allocate service rather than select the cheapest coefficient in
each demand column.

The same mathematics could be rendered as clinic siting, regional repair depots,
or computing-service hubs.  The entities and units may change, but the three
invariants---fixed activation, demand satisfaction, and capacity linking---must
remain present.  Forward reconstruction is accepted only if it recovers these
invariants and reproduces the reference objective within tolerance.

\subsection{Illustrative case II: executable but structurally wrong models}
\label{app:orcpt-case-failure}

This example also shows why execution success is insufficient.
Table~\ref{tab:orcpt-example-errors} contrasts the faithful formulation with
two executable but structurally incorrect reconstructions and identifies the
validation gates that detect each error.

\begin{table}[H]
  \centering
  \caption{Illustrative structural errors and the gates that detect them.}
  \label{tab:orcpt-example-errors}
  \small
  \begin{tabularx}{\linewidth}{@{}p{0.22\linewidth}p{0.27\linewidth}p{0.19\linewidth}X@{}}
    \toprule
    Reconstruction & Structural defect & Illustrative optimum & Detection \\
    \midrule
    Faithful model
      & Fixed costs and capacity linking retained
      & 590
      & Accepted after contract and objective checks. \\
    Fixed costs omitted
      & Sites can be opened without paying activation cost
      & 315
      & Contract reports a missing objective term; objective comparison fails. \\
    Capacity constraints omitted
      & Site B serves all demand despite capacity 65
      & 565
      & Contract reports a missing constraint; objective comparison fails. \\
    \bottomrule
  \end{tabularx}
\end{table}

All three programs can be syntactically valid, executable, and solved to an
\texttt{OPTIMAL} status.  That status certifies only the optimum of the program
that was actually constructed.  It does not certify faithfulness to the
intended business problem.  The example therefore motivates three distinct
checks: static presence of required terms, family-level structural contracts,
and independent objective comparison.

Repair follows the same rule.  When a problem statement lacks a coefficient or
bound, the safe action is regeneration, rejection, or recovery from an
authoritative source table.  A repair must not invent a value merely to make the
model bounded or to approach the reference objective.

\section{CPT--SFT--Deployment--Evaluation Provenance}
\label{app:provenance}

Post-training large-scale MoE models involves multiple stages—CPT, SFT, weight conversion, serving, and evaluation—each producing artifacts dependent on upstream outputs. We introduce a lightweight provenance system that links all stages via per-stage manifests and a local file-based registry, to avoid opacity between benchmark scores and training configurations, checkpoints, or deployed weights, thereby ensuring clear attribution of improvements, reproducibility, and auditability of evaluation claims. The system serves three functions: (i) it enforces a verifiable chain from training data to scores, tracing each result to its exact checkpoint; (ii) it enables root-cause analysis of regressions by tracing deployed models back through conversion and training to the upstream CPT run; (iii) it provides programmatic support for automated tracking, dashboard monitoring, and artifact management, without database dependencies or disruption to existing workflows.

\subsection{Design principles}

\begin{enumerate}
  \item \textbf{Model identity is decoupled from the service endpoint.}
  A served model name, IP address, and port identify the inference entry point,
  not the model version. The authoritative model identity is carried by the
  artifact manifest at the weight directory, together with the upstream training
  lineage recorded at each stage. This separation allows multiple model versions
  to be served under the same endpoint name while preserving unambiguous
  identification of which weights are in use.

  \item \textbf{Every stage records its ground truth at the point of execution.}
  Each stage writes the actual paths, key environment variables, and upstream
  relationships into a structured manifest immediately upon completion. This
  manifest, rather than any later reconstruction, serves as the authoritative
  record for that stage. The point-of-execution recording eliminates the
  fragility of post-hoc documentation and ensures that environmental details
  (e.g., CANN version, parallelism configuration, tokenizer path) are captured
  before they can be altered or forgotten.

  \item \textbf{Historical experiments can be incorporated post-hoc.}
  Registration tooling accepts existing configuration snapshots, checkpoint
  directories, and converted artifacts, allowing runs completed before the
  provenance system was deployed to be retroactively linked into the provenance
  graph. This ensures that valuable legacy experiments are not excluded from
  cross-run comparison and trend analysis.
\end{enumerate}

\subsection{Provenance chain and registry layout}

The end-to-end chain spans five stages: CPT training, SFT training, HF artifact
conversion, model deployment, and benchmark evaluation. Each stage produces a
manifest file at its output location and registers a copy in a centralized JSON
registry. The registry is organized by stage type (training runs, artifacts,
deployments, evaluation runs), with a lightweight index file maintaining
aggregate counts. In addition to the centralized entries, manifests are always
co-located with the artifacts they describe---the HF artifact manifest resides
in the weight directory, the deployment manifest resides in the serving log
directory, and the evaluation provenance resides in the benchmark summary
directory. This dual placement ensures that the provenance record travels with
the artifact even if the centralized registry is unavailable, and that any
downstream consumer of the artifact (e.g., a separate evaluation pipeline) can
discover its lineage without consulting the central registry.

\subsection{Stage-level linkage mechanism}

\paragraph{CPT $\rightarrow$ SFT.}
Both CPT and SFT runs produce a configuration snapshot and lineage manifest
through a shared registration interface. The parent--child relationship is
established by matching the SFT checkpoint load path against the CPT checkpoint
save path, recording \texttt{linked} status with \texttt{path\_match} confidence
on success, or \texttt{missing} otherwise. This link is essential for
CPT-to-SFT transfer analysis: without it, a reported transfer gain cannot be
traced to the specific CPT checkpoint that provided the initialization.

\paragraph{SFT $\rightarrow$ HF artifact.}
Upon weight format conversion to HuggingFace-compatible shards, the converter
writes an artifact manifest recording the artifact identifier, source training
run, checkpoint directory and iteration, HF output directory, and upstream CPT
lineage, with a copy registered centrally. This manifest is the authoritative
record connecting deployable weights to their training provenance.

\paragraph{HF artifact $\rightarrow$ model deployment.}
The serving launch procedure reads the artifact manifest from the weight
directory and generates a deployment manifest capturing the deployment
identifier, artifact and training run references, model path, served model name,
and inference endpoint URL, with copies in both the log directory and the
central registry. This artifact-to-endpoint binding enables the dashboard to
track which model version is served at each endpoint.

\paragraph{Model deployment $\rightarrow$ eval run.}
Evaluation accepts an optional deployment manifest reference via an environment
variable. When provided, the harness extends the benchmark summary with full
provenance fields spanning evaluation, deployment, artifact, SFT, and CPT.
Without it, evaluation proceeds normally but scores cannot be automatically
traced to specific weights or associated with the corresponding dashboard
experiment.

\subsection{Confidence levels}

Each provenance link carries a confidence annotation, summarized in
Table~\ref{tab:provenance_confidence}, that distinguishes
manifest-based exact linkage from path-based or heuristic matching.

\begin{table}[t]
\centering
\small
\caption{Provenance confidence levels assigned to each link in the chain.}
\label{tab:provenance_confidence}
\begin{tabular}{lp{0.65\linewidth}}
\toprule
Level & Condition \\
\midrule
\texttt{exact\_manifest}
  & The downstream stage directly reads an explicit manifest produced by the
  upstream stage (e.g., the deployment script reads the artifact manifest
  from the HF weight directory). \\
\texttt{path\_match}
  & The link is established by matching path fields across independently
  produced configuration records (e.g., the SFT checkpoint load path matches
  the CPT checkpoint save path). \\
\texttt{inferred\_from\_log}
  & Reserved for future use, where training or deployment logs may provide
  auxiliary evidence for a link. \\
\texttt{missing}
  & Insufficient information exists to establish the link; the downstream
  stage operates without a confirmed upstream connection. \\
\bottomrule
\end{tabular}
\end{table}

\subsection{Dashboard monitoring and retroactive registration}

The provenance registry directly feeds a monitoring dashboard that visualizes the
full experiment graph, each row corresponds to a training run, and
columns display the run identifier, job type, training iterations, key metrics
(loss, benchmark scores), and---critically---provenance status and confidence
annotations for each link in the chain. Runs with complete manifest coverage are
marked with \texttt{exact\_manifest} or \texttt{path\_match} confidence; legacy
runs that predate the provenance system are displayed with a \texttt{missing}
annotation but are otherwise fully functional in the dashboard.

The dashboard is refreshed by a build script that scans the training archive
directory and the centralized registry, aggregating manifest contents into a
single JSON consumed by the frontend. Newly completed runs appear automatically
on the next rebuild; legacy runs without manifests can be registered
retroactively by pointing the registration tooling at their existing
configuration files, checkpoint directories, or artifact outputs. This design
ensures that the dashboard reflects both actively tracked experiments and
backfilled historical data within a unified view, making it the operational
interface through which experiment provenance is queried, audited, and
cross-referenced against benchmark claims.

\section{Training Schedule and Monitoring Design}
\label=/{sec:curriculum}

While the data refinement pipeline improves individual sample quality, the
macro-level organization of training data also requires careful design.
Our data-mixture analysis shows that overly concentrated OR-domain adaptation
can weaken general capabilities, motivating a training schedule that balances
OR, mathematics, code, and general-domain data. We therefore design a
predefined curriculum schedule together with a monitoring dashboard for
tracking training stability and capability retention. The schedule, stage
boundaries, and transition rules are specified before training, while the
dashboard serves as an observability tool for manual inspection.







\subsection{Static Curriculum Learning Design}

We organize the training data mixture into a pre-defined, three-stage schedule. The schedule uses a mixing vector \(\bm{\alpha}(t) = [\alpha_{\text{OR}}, \alpha_{\text{math}}, \alpha_{\text{code}}, \alpha_{\text{gen}}]\) that is varied across training steps \(t\) according to a fixed plan:

\begin{enumerate}
    \item \textbf{Foundation stage (first 30\% of steps):} Heavy emphasis on mathematical reasoning and clean general code (\(\alpha_{\text{math}}=0.35, \alpha_{\text{code}}=0.35, \alpha_{\text{gen}}=0.20, \alpha_{\text{OR}}=0.10\)). This phase reinforces precise token generation and algorithmic logic.

    \item \textbf{Domain ramp-up (30--70\%):} \(\alpha_{\text{OR}}\) is linearly increased to \(0.40\) while \(\alpha_{\text{code}}\) gradually reduces to \(0.15\). Within this stage, OR samples are introduced by difficulty format: first Data-in-Problem (DP), then Data-in-Table (DT), and finally Data-Problem-Separate (DPS). The transition to a more difficult format is gated by a fixed evaluation threshold on a held-out subset of the current format: the model must achieve more than \(85\%\) execution success before the next format is unlocked. This threshold is a hard rule, not an adaptive algorithm.

    \item \textbf{Contract hardening (70--100\%):} The mixture shifts back toward code-only and LP-writing tasks (\(\alpha_{\text{code}}\) raised to \(0.25\)) to reinforce strict output contracts, using validated samples from the data refinement pipeline.
\end{enumerate}

This schedule is a \textbf{default trajectory} whose parameters (proportions, thresholds, stage boundaries) were fixed based on preliminary runs and the data-mixture ablation in Section~\ref{subsec:sft_training_cleaning} (Figure~\ref{fig:ablation}). It is not modified dynamically during training. Curriculum-like scheduling strategies have been explored in other contexts~\citep{tzannetos2026curriculum,zhao2026rethinking,liang2025boosting}; our specific instantiation is one manually designed configuration.

\subsection{Monitoring Dashboard}

To observe training behavior and detect potential problems early, we implemented a monitoring setup that logs two groups of metrics at regular intervals (every 500 training steps). This dashboard is not a closed-loop controller; it is an observability layer for human developers.

\textbf{Training-health track.} Standard optimization signals: loss, gradient norm, NaN counts, and Model FLOPs Utilization (MFU). A spike in gradient norm or a NaN loss immediately signals a data or training issue requiring manual investigation.

\textbf{Capability-retention track.} A set of metrics that track both domain and general performance:
\begin{itemize}
    \item \textbf{OR capability:} validation perplexity on an OR-domain set (partitioned by format), and task-specific rewards (code\_reward, wl\_reward) on small proxy subsets of NL4OPT and OptiBench.
    \item \textbf{General code quality:} a \textit{syntax health score} computed on a held-out set of 200 pure-Python algorithmic problems. The score checks for non-ASCII tokens, indentation errors, import correctness, and basic static validity.
    \item \textbf{General language fluency:} perplexity on a fixed general-domain corpus (Wikipedia, books) to detect catastrophic forgetting.
\end{itemize}

These metrics are logged and visualized in WandB dashboards. During development, they were used for manual inspection: for example, a persistent decline in syntax health could prompt the developer to pause OR-centric training, inspect recent data, or adjust the mixture. The specific thresholds and the resulting decisions were made by human engineers, not by an automated state machine. The dashboard design is inspired by systematic analyses of training dynamics~\citep{qi2026evolm}, and its instrumentation code is included in the released artifacts.

\newpage
\section{Representative Original-Checkpoint Error Cases}
\label{app:original_checkpoint_error_cases}

\raggedbottom

Section~\ref{subsec:original_checkpoint_diagnosis} summarizes the aggregate
failure distribution of the original DeepSeek-V4-Flash checkpoint. This
appendix records the concrete examples used in that diagnosis. The cases are
traced back to the corresponding evaluation artifacts and are grouped by the
six failure types discussed in the Inference-Setting Shift analysis.

\definecolor{casePrimary}{HTML}{B34A78}
\definecolor{caseDark}{HTML}{7A2E50}
\definecolor{caseLight}{HTML}{F8EDF2}
\definecolor{caseMid}{HTML}{D99AB7}
\definecolor{caseText}{HTML}{2B2227}
\definecolor{caseMuted}{HTML}{7B6A72}
\definecolor{errRed}{HTML}{B21F3A}

\lstdefinestyle{pythoncase}{
  language=Python,
  basicstyle=\ttfamily\scriptsize,
  keywordstyle=\color{caseDark}\bfseries,
  commentstyle=\color{caseMuted},
  stringstyle=\color{casePrimary},
  breaklines=true,
  columns=fullflexible,
  keepspaces=true,
  showstringspaces=false,
  aboveskip=0.4em,
  belowskip=0.2em,
  escapeinside={(*@}{@*)}
}

\newtcolorbox{casebox}[1]{
  enhanced,
  colback=caseLight,
  colframe=caseDark,
  coltitle=white,
  colbacktitle=casePrimary,
  title={#1},
  fonttitle=\bfseries,
  sharp corners,
  boxrule=0.9pt,
  boxsep=1mm,
  left=1mm,
  right=1mm,
  top=0.8mm,
  bottom=0.8mm,
  before skip=0pt,
  after skip=0pt
}

\newenvironment{casepage}[1]{%
  \clearpage
  \begin{casebox}{#1}
}{%
  \end{casebox}
}

\newenvironment{casepagestart}[1]{%
  \begin{casebox}{#1}
}{%
  \end{casebox}
}

\vspace{1em}
\begin{casepagestart}{Case Study: Canonical LP-Graph Mismatch under ORGEval}

\begin{tabularx}{\linewidth}{@{}p{0.22\linewidth}X@{}}
\toprule
Benchmark & B4O ORGEval \\
Sample ID & \texttt{BENCH4OPT\_0}; related: \texttt{BENCH4OPT\_1}, \texttt{BENCH4OPT\_12} \\
Evaluation & \texttt{code\_reward=1.0}, \texttt{wl\_reward=0.0}; \texttt{var\_num\_check=True}, \texttt{cons\_num\_check=False} \\
Failure Type & Executable LP with non-reference constraint family \\
\bottomrule
\end{tabularx}

\vspace{0.5em}
\textbf{Original Problem}

{\small
The problem is a pharmaceutical blending task. The model must choose
ingredient quantities to minimize cost while respecting ingredient
availability, medication dosage requirements, and active-ingredient purity
bounds. ORGEval does not only check whether the generated LP can be written;
it compares the variable family, constraint family, and their graph structure
against the reference LP.
\par}

\vspace{0.5em}
\textbf{Model Output}

\begin{lstlisting}[style=pythoncase]
x = model.addVars(I, lb=0.0, ub=GRB.INFINITY, name='x')
model.setObjective(gp.quicksum(costs[i] * x[i] for i in range(I)), GRB.MINIMIZE)

for i in range(I):
    model.addConstr(x[i] <= availability[i], name=f'availability_{i}')

# The model assumes dosage bounds apply to the total medication amount.
for m in range(M):
    total_med = gp.quicksum(requirements[m][i] * x[i] for i in range(I))
    model.addConstr(total_med >= min_dosage[m], name=f'min_dosage_{m}')
    model.addConstr(total_med <= max_dosage[m], name=f'max_dosage_{m}')

total_ingredients = gp.quicksum(x[i] for i in range(I))
for i in range(I):
    model.addConstr(x[i] >= min_purity[i] * total_ingredients,
                    name=f'min_purity_{i}')
    model.addConstr(x[i] <= max_purity[i] * total_ingredients,
                    name=f'max_purity_{i}')
\end{lstlisting}

\vspace{0.3em}
\textbf{Error Annotation.}
The program is executable and writes an LP file, but the generated constraint
family does not match the reference formulation. In this family of cases,
\texttt{BENCH4OPT\_1} first fails variable-count matching, and
\texttt{BENCH4OPT\_12} shows the next level of difficulty: even when variable
and constraint counts match, the WL graph can still be wrong. These examples
show why structural ORGEval failures motivate CPT for domain templates and SFT
for benchmark-facing canonical formulations.

\end{casepagestart}

\begin{casepage}{Case Study: Output-Contract and Schema Failures}

\begin{tabularx}{\linewidth}{@{}p{0.22\linewidth}X@{}}
\toprule
Benchmark & B4O ORGEval \\
Sample ID & \texttt{BENCH4OPT\_4}; \texttt{BENCH4OPT\_176} \\
Evaluation & \texttt{BENCH4OPT\_4}: syntax failure from markdown fence; \texttt{BENCH4OPT\_176}: \texttt{TypeError} from list indexing with a string shift label \\
Failure Type & Code-only protocol violation and JSON/list schema mismatch \\
\bottomrule
\end{tabularx}

\vspace{0.5em}
\textbf{Original Problems}

{\small
\texttt{BENCH4OPT\_4} is a capital-budgeting problem: select investment
projects to maximize NPV under budget, risk, and dependency constraints.
\texttt{BENCH4OPT\_176} is a hospital nurse-scheduling problem: assign nurses
to shifts while satisfying shift coverage, skill requirements, and maximum
working hours.
\par}

\vspace{0.5em}
\textbf{Model Output}

\begin{lstlisting}[style=pythoncase]
# BENCH4OPT_4: code-only target, but the generated response starts with:
(*@\textbf{\textcolor{errRed}{```python}}@*)
import json
import gurobipy as gp
from gurobipy import GRB

# BENCH4OPT_176: shifts are string labels, while shift_requirements is a list.
for j in shifts:
    model.addConstr(
        gp.quicksum(x[i, j] for i in nurses) >=
        (*@\textbf{\textcolor{errRed}{\texttt{shift\_requirements[j]}}}@*),
        name=f"shift_coverage_{j}"
    )
\end{lstlisting}

\vspace{0.3em}
\textbf{Error Annotation.}
The first error blocks evaluation before the model is even built: the executor
sees the markdown fence token as Python syntax. The second error reaches
model construction but fails when a string shift name is used as a list index.
Both examples primarily motivate SFT: the model already has useful OR
knowledge, but Pass@1 behavior must obey code-only contracts and schema-safe
data-access templates.

\end{casepage}

\begin{casepage}{Case Study: Discrete Semantics Error in NL4OPT}

\begin{tabularx}{\linewidth}{@{}p{0.22\linewidth}X@{}}
\toprule
Benchmark & NL4OPT \\
Sample ID & \texttt{NL4OPT\_15} \\
Evaluation & Ground truth: 28250; Prediction: 28125 \\
Failure Type & Continuous relaxation of integer workforce variables \\
\bottomrule
\end{tabularx}

\vspace{0.5em}
\textbf{Original Problem}

{\small
A store employs senior citizens who earn \$500 per week and young adults who earn \$750 per week.
The store must keep the weekly wage bill below \$30000.
On any day, the store requires at least 50 workers, of whom at least 10 must be young adults.
To ensure the store runs smoothly, the number of young adults should be at least a third the number of senior citizens.
Formulate a LP to minimize the wage bill.
\par}

\vspace{0.5em}
\textbf{Model Output}

\begin{lstlisting}[style=pythoncase]
import gurobipy as gp
from gurobipy import GRB

m = gp.Model("MinimizeWageBill")

x = m.addVar(vtype=(*@\textbf{\textcolor{errRed}{GRB.CONTINUOUS}}@*), name="seniors")
y = m.addVar(vtype=(*@\textbf{\textcolor{errRed}{GRB.CONTINUOUS}}@*), name="young_adults")

m.setObjective(500 * x + 750 * y, sense=GRB.MINIMIZE)

m.addConstr(x + y >= 50, "total_workers")
m.addConstr(y >= 10, "min_young")
m.addConstr(3 * y >= x, "young_senior_ratio")
m.addConstr(500 * x + 750 * y <= 30000, "wage_budget")

m.optimize()

if m.status == GRB.OPTIMAL:
    print(f"Optimal solution: x = {x.X:.2f}, y = {y.X:.2f}")
    print(f"Minimum wage bill = ${m.ObjVal:.2f}")
else:
    print("No optimal solution found.")
\end{lstlisting}

\vspace{0.3em}
\textbf{Error Annotation.}
The problem asks for numbers of workers, but the model uses continuous variables.
This relaxes the integer workforce decision and yields a fractional optimum.

\end{casepage}

\begin{casepage}{Case Study: Loss/Yield-Driven Conservation Error}

\begin{tabularx}{\linewidth}{@{}p{0.22\linewidth}X@{}}
\toprule
Benchmark & B4O Feasible \\
Sample ID & \texttt{BENCH4OPT\_2} \\
Evaluation & Pass@1 prediction: 265.5768398591736; ground truth: 531.1536797183472. 5-shot and Pass@16 recover the reference value. \\
Failure Type & Evaporation loss mixed into both cost and flow conservation \\
\bottomrule
\end{tabularx}

\vspace{0.5em}
\textbf{Original Problem}

{\small
The task is a water-distribution network-flow problem. Each edge has a
capacity, transportation cost, and evaporation rate. The model must satisfy
node demand while minimizing the total cost induced by flow and evaporation.
\par}

\vspace{0.5em}
\textbf{Model Output}

\begin{lstlisting}[style=pythoncase]
flow = {}
for i, (source, target, capacity, cost, evaporation_rate) in enumerate(edges):
    flow[(source, target)] = model.addVar(lb=0.0, ub=capacity,
                                         name=f"flow_{source}_{target}")

obj = gp.LinExpr()
for (source, target), var in flow.items():
    ...
    obj += (*@\textbf{\textcolor{errRed}{\texttt{(cost + evaporation\_rate) * var}}}@*)
model.setObjective(obj, GRB.MINIMIZE)

for node_idx, node in enumerate(nodes):
    inflow = gp.LinExpr()
    outflow = gp.LinExpr()
    evaporation_loss = gp.LinExpr()
    ...
    model.addConstr(
        inflow - outflow - (*@\textbf{\textcolor{errRed}{\texttt{evaporation\_loss}}}@*) == demand[node_idx],
        name=f"flow_conservation_{node}"
    )
\end{lstlisting}

\vspace{0.3em}
\textbf{Error Annotation.}
The code is syntactically valid and reaches an optimal solution, but it
implements the wrong physical system. Evaporation is treated as an additive
unit cost and also deducted from conservation, blurring whether loss affects
delivered quantity, incurred cost, or both. This is a domain-pattern failure:
network flow with loss, yield, shrinkage, and evaporation needs stable
conservation templates rather than generic LP scaffolding.

\end{casepage}

\begin{casepage}{Case Study: Ratio, Unit, and Total-Quantity Coupling Error}

\begin{tabularx}{\linewidth}{@{}p{0.22\linewidth}X@{}}
\toprule
Benchmark & OptiBench \\
Sample ID & \texttt{OPTIBENCH\_6} \\
Evaluation & Pass@1 prediction: 10000.0; 5-shot prediction: 28000.0; Pass@16 prediction: 10000.0; ground truth: 18000.0 \\
Failure Type & Investment-return units and total savings are coupled incorrectly \\
\bottomrule
\end{tabularx}

\vspace{0.5em}
\textbf{Original Problem}

{\small
A logistics company allocates 50 trucks among three goods. Each good has an
initial per-truck delivery cost. Investment in route-optimization software can
reduce per-truck cost, and total investment must not exceed the total savings
generated by the investment. The objective is to minimize total delivery cost
while satisfying minimum truck counts for Goods X and Y.
\par}

\vspace{0.5em}
\textbf{Model Output}

\begin{lstlisting}[style=pythoncase]
x = m.addVars(goods, vtype=GRB.INTEGER, name="trucks")
I = m.addVars(goods, vtype=GRB.CONTINUOUS, name="investment")

# cost per truck = c_i - (r_i/100)*I_i
obj = gp.QuadExpr()
for i in goods:
    obj += (c[i] - (r[i]/100.0) * I[i]) * x[i]
m.setObjective(obj, GRB.MINIMIZE)

m.addConstr(gp.quicksum(I[i] for i in goods) <= B, name="budget")

# Savings constraint: total investment <= total cost savings
savings_lhs = gp.quicksum(I[i] for i in goods)
savings_rhs = gp.quicksum(
    (*@\textbf{\textcolor{errRed}{(r[i]/100.0) * I[i] * x[i]}}@*) for i in goods
)
m.addConstr(savings_lhs <= savings_rhs, name="savings")
m.params.NonConvex = 2
\end{lstlisting}

\vspace{0.3em}
\textbf{Error Annotation.}
The model recognizes that investment and truck allocation interact, but it does
not correctly reconcile dollars, hundreds-of-dollars, per-truck reduction, and
total savings. In 5-shot, the error changes shape: the model moves toward a
linear-looking savings template but can drop the scale variable. This case
therefore exposes a deeper semantic issue than API usage: ratio, ROI,
average-cost, and efficiency statements require explicit tracking of numerator,
denominator, unit, and total quantity.

\end{casepage}

\begin{casepage}{Case Study: Solver-Compatible Nonlinear Reformulation Failure}

\begin{tabularx}{\linewidth}{@{}p{0.22\linewidth}X@{}}
\toprule
Benchmark & OptiBench \\
Sample ID & \texttt{OPTIBENCH\_13} \\
Evaluation & Pass@1: optimization failed with ``Objective must be linear or quadratic''; 5-shot prediction: 0.125; Pass@16 prediction: 0.125; ground truth: 0.35355235026576626 \\
Failure Type & Fourth-order objective passed directly to Gurobi \\
\bottomrule
\end{tabularx}

\vspace{0.5em}
\textbf{Original Problem}

{\small
A person is trapped near a swamp boundary described by \(y \ge x^2\) with
\(x \ge 0\). The task is to find the closest feasible rescue point to
\((3/4,0)\), equivalently the shortest rope length to a point on or above the
parabolic boundary.
\par}

\vspace{0.5em}
\textbf{Model Output}

\begin{lstlisting}[style=pythoncase]
m = gp.Model("SwampRescue")
x = m.addVar(lb=0.0, vtype=GRB.CONTINUOUS, name="x")

# Objective: minimize (x - 0.75)^2 + (x^2)^2
obj = (x - 0.75)*(x - 0.75) + (*@\textbf{\textcolor{errRed}{x*x*x*x}}@*)
m.setObjective(obj, GRB.MINIMIZE)

m.optimize()
\end{lstlisting}

\vspace{0.3em}
\textbf{Error Annotation.}
The mathematical expression is meaningful, but it is not a linear or quadratic
objective. Later inference settings avoid the initial build error yet still
return the wrong distance, so the missing capability is not merely a Gurobi API
habit. The model needs solver-aware reformulation knowledge: when to introduce
auxiliary variables, when nonconvex quadratic or general constraints are
legitimate, and when analytic simplification is required before coding the
optimization model.

\end{casepage}

\end{document}